\documentclass[a4paper,fleqn]{cas-sc}

\usepackage[numbers]{natbib}
\usepackage{graphicx}%
\usepackage{multirow}%
\usepackage{amsmath,amssymb,amsfonts}%
\usepackage{amsthm}%
\usepackage{mathrsfs}%
\usepackage[title]{appendix}%
\usepackage{xcolor}%
\usepackage{textcomp}%
\usepackage{manyfoot}%
\usepackage{booktabs}%
\usepackage{algorithm}%
\usepackage{algorithmicx}%
\usepackage{algpseudocode}%
\usepackage{listings}%
\usepackage{hyperref}%
\usepackage{booktabs}%
\usepackage{adjustbox}%
\usepackage{tabularx}%
\usepackage{diagbox}%
\usepackage{caption}%
\usepackage{subcaption, soul}%

\usepackage{acronym}
\newacro{gan}[GAN]{generative adversarial network}
\newacro{vae}[VAE]{variational autoencoder}
\newacro{cnn}[CNN]{convolutional neural network}
\newacro{dire}[DIRE]{diffusion reconstruction error}
\newacro{mlp}[MLP]{multi-layer perceptron}
\newacro{clip}[CLIP]{contrastive language image pre-training}
\newacro{vlm}[VLM]{vision-language model}
\newacro{vit}[ViT]{vision transformer}
\newacro{nlp}[NLP]{natural language processing}
\newacro{cv}[CV]{computer vision}
\newacro{blip}[BLIP]{bootstrapping language image pre-training}
\newacro{vqa}[VQA]{visual question answering}
\newacro{lora}[LORA]{low-rank adaptation}
\newacro{peft}[PEFT]{parameter-efficient fine-tuning}
\newacro{gpt}[GPT]{generative pre-trained transformer}
\newacro{q-former}[Q-Former]{querying transformer}
\newacro{sedid}[SeDID]{stepwise error for diffusion-generated image detection}
\newacro{sd}[SD]{stable diffusion}
\newacro{lsun}[LSUN]{large-scale scene understanding}
\newacro{fc}[FC]{fully-connected}
\newacro{deit}[DeiT]{data-efficient image transformers}
\newacro{ldm}[LDM]{latent diffusion model}
\newacro{adm}[ADM]{ablated diffusion model}
\newacro{ddpm}[DDPM]{denoising diffusion probabilistic models}
\newacro{iddpm}[IDDPM]{improved denoising diffusion probabilistic models}
\newacro{pndm}[PNDM]{pseudo numerical methods for diffusion models on manifolds}
\newacro{srm}[SRM]{spatial rich model}
\newacro{lasted}[LASTED]{language-guided synthesis detection}
\newacro{rf}[RF]{random forest}
\newacro{dm}[DM]{diffusion model}
\newacro{ddim}[DDIM]{denoising diffusion implicit models}
\newacro{multilid}[multiLID]{multi local intrinsic dimensionality}
\newacro{ifdl}[IFDL]{image forgery detection and localization} 
\newacro{svm}[SVM]{support vector machine} 
\newacro{ai}[AI]{artificial intelligence} 

\newacro{amsff}[AMSFF]{attention-based multi-scale feature fusion} 
\newacro{psm}[PSM]{patch selection module}
\newacro{llm}[LLM]{large language model}
\newacro{clip}[CLIP]{Contrastive Language-Image Pretraining}
\newacro{c2p}[C2P]{category common prompt}
\newacro{sid}[SID]{synthetic image detection}
\newacro{fosid}[FOSID]{Fact-checked Online Synthetic Image Dataset}
\newacro{rasid}[RASID]{retrieval-assisted synthetic image detection}
\newacro{gff}[GFF]{guided and fused frozen CLIP-ViT}
\newacro{fuseformer}[FuseFormer]{multi-stage fusion module}
\newacro{dfgm}[DFGM]{deepfake-specific feature guidance module}
\newacro{ffaa}[FFAA]{face forgery analysis assistant}
\newacro{mllm}[MLLM]{multimodal large language model}
\newacro{mids}[MIDS]{Multi-answer Intelligent Decision System}
\newacro{ssm}[SSM]{state-space model}
\newacro{einfft}[EinFFT]{Einstein FFT}
\newacro{fft}[FFT]{Fast Fourier Transform}
\newacro{vim}[ViM]{Vision Mamba}
\newacro{ms2d}[MS2D]{Multi-Scale 2D}
\newacro{convffn}[ConvFFN]{onvolutional Feed-Forward
Network}
\newacro{cnn}[CNN]{convolutional neural network}
\newacro{dvae}[dVAE]{discrete variational autoencoder}
\newacro{coglm}[CogLM]{cross-modal general language model}
\newacro{qformer}[Q-Former]{querying transformer}
\newacro{sdgs}[SDGS]{Synthetic Data Generation System}
\newacro{sota}[SOTA]{state-of-art}
\newacro{mse}[MSE]{mean squared error}
\newacro{ssim}[SSIM]{structural similarity index measure}
\newacro{psnr}[PSNR]{peak signal-to-noise ratio}
\newacro{vssd}[VSSD]{visual state space duality}
\newacro{moe}[MoE]{Mixture of Experts}

\def\tsc#1{\csdef{#1}{\textsc{\lowercase{#1}}\xspace}}
\tsc{WGM}
\tsc{QE}
\begin{document}
\let\WriteBookmarks\relax
\def\floatpagepagefraction{1}
\def\textpagefraction{.001}

\shorttitle{}    

\shortauthors{}  

\title [mode = title]{Can Visual Mamba Improve AI-Generated Image Detection? An In-Depth Investigation}  
% On the Effectiveness of Vision Mamba in Detecting AI-Generated Images



% \author[uphf]{Mamadou Keita}
% \author[tii]{Wassim Hamidouche}
% \author[uphf]{Hessen Bougueffa Eutamene}
% \author[sorbonne]{Abdenour Hadid}
% \author[uphf]{Abdelmalik Taleb-Ahmed}
% \affiliation[uphf]{
%     organization={Institut d’Electronique de Microélectronique et de Nanotechnologie (IEMN), UMR 8520, Université Polytechnique Hauts de France},
%     city={Valenciennes},
%     postcode={59313}, 
%     % state={},
%     country={France}
% }
% \affiliation[tii]{
%     organization={Technology Innovation Institute},
%     postcode={P.O.Box: 9639},
%     city={Masdar City}, 
%     % state={},
%     country={Abu Dhabi, UAE}
% }
% \affiliation[sorbonne]{
%     organization={Sorbonne Center for Artificial Intelligence, Sorbonne University},
%     % postcode={},
%     % city={, 
%     % state={},
%     country={Abu Dhabi, UAE}
% }
\author[1]{Mamadou Keita}[orcid=0009-0009-7618-9253]

\cormark[1]


\ead{mamadou.keita@uphf.fr}



\affiliation[1]{organization={Laboratory of IEMN, CNRS, Centrale Lille, UMR 8520, Univ. Polytechnique Hauts-de-France},
            city={Valenciennes},
%          citysep={}, % Uncomment if no comma needed between city and postcode
            postcode={59300}, 
            % state={Nord Pas de Calais},
            country={France}}

\author[2]{Wassim Hamidouche}[orcid=0000-0002-0143-1756]

% Footnote of the second author

% Email id of the second author
\ead{whamidouche@gmail.com}


% \credit{ownership}

\affiliation[2]{organization={Khalifa University},
            addressline={}, 
            city={Abu Dhabi},
%          citysep={}, % Uncomment if no comma needed between city and postcode
            country={UAE}}
            
\author[1]{Hessen Bougueffa Eutamene}[orcid=0009-0009-0556-9996]

% Footnote of the second author
% \fnmark[3]

% Email id of the second author
\ead{Hessen.BougueffaEutamene@uphf.fr}

\author[1]{Abdelmalik Taleb-Ahmed}[orcid=0000-0001-7218-3799]

% Footnote of the second author
% \fnmark[4]

% Email id of the second author
\ead{abdelmalik.taleb-ahmed@uphf.fr}

\author[3]{Xianxun Zhu}[orcid=0000-0003-3958-7040]

% Footnote of the second author
% \fnmark[5]

% Email id of the second author
\ead{zhuxianxun@shu.edu.cn}



\affiliation[3]{organization={School of Communication and Information Engineering, Shanghai University},
            city={Shanghai},
%          citysep={}, % Uncomment if no comma needed between city and postcode
            country={China}}

\author[4]{Abdenour Hadid}[orcid=0000-0001-9092-735X]

% Footnote of the second author
% \fnmark[5]

% Email id of the second author
\ead{abdenour.hadid@ieee.org}



\affiliation[4]{organization={Sorbonne Center for Artificial Intelligence, Sorbonne University Abu Dhabi},
            %addressline={}, 
            city={Abu Dhabi},
%          citysep={}, % Uncomment if no comma needed between city and postcode
            country={UAE}}

\cortext[1]{Corresponding author}



\begin{abstract}
In recent years, computer vision has witnessed remarkable progress, fueled by the development of innovative architectures such as Convolutional Neural Networks (CNNs), Generative Adversarial Networks (GANs), diffusion-based architectures, Vision Transformers (ViTs), and, more recently, Vision-Language Models (VLMs). This progress has undeniably contributed to creating increasingly realistic and diverse visual content. However, such advancements in image generation also raise concerns about potential misuse in areas such as misinformation, identity theft, and protecting privacy and security. In parallel, Mamba-based architectures have emerged as a versatile tools for a range of image analysis tasks, including classification, segmentation, medical imaging, object detection, and image restoration, in this rapidly evolving field. However, their potential for identifying AI-generated images remains relatively unexplored compared to established techniques. This study provides a systematic evaluation and comparative analysis of Vision Mamba models for AI-generated image detection. We benchmark multiple Vision Mamba variants against representative CNNs, ViTs, and VLM-based detectors across diverse datasets and synthetic image sources, focusing on key metrics such as accuracy, efficiency, and generalizability across diverse image types and generative models. Through this comprehensive analysis, we aim to elucidate Vision Mamba’s strengths and limitations relative to established methodologies in terms of applicability, accuracy, and efficiency in detecting AI-generated images. Overall, our findings highlight both the promise and current limitations of Vision Mamba as a component in systems designed to distinguish authentic from AI-generated visual content. This research is crucial for enhancing detection capabilities in an age where distinguishing between authentic and AI-generated content is increasingly challenging.
\end{abstract}


\begin{highlights}
\item  The paper deals with a timely topic which is the detection of AI generated images.
\item  Vision Mamba has emerged as a versatile tool for various image analysis tasks but its potential for identifying AI-generated images remains relatively unexplored.
\item The paper thoroughly investigates the performance of Vision Mamba models in detecting AI-generated images.
\item Extensive experiments are reported with comparison with State-of-the-art including methods based on CNNs, attention mechanisms (Transformers), and VLMs.
\item The experiments indicate that Vision Mamba models perform well while exhibiting difficulties in generalizing across diverse data distributions.
\item The results underscore the superior generalization ability of VLMs over existing baselines and state-of-art (SOTA) methods.
\end{highlights}

% Keywords

\maketitle

% Main text
\section{Introduction}

The emergence of advanced image generation models, fueled by deep learning, has profoundly transformed computer vision. Notably, models like \acp{gan}~\cite{goodfellow2014generative} and diffusion-based architectures~\cite{sohl2015deep} have achieved the remarkable feat of generating photorealistic images that closely resemble real-world visuals~\cite{karras2021alias,ramesh2021zero,rombach2022high,saharia2022photorealistic}. While this advancement has ushered in new opportunities in entertainment, art, and content creation, it also presents substantial challenges concerning trust, security, and authenticity. In particular, \ac{ai}-generated images can be exploited for malicious purposes, ranging from disseminating fake news and deceptive media to creating identities and violating privacy~\cite{paik2023affective}. As the risk of such misuse grows, the imperative for reliable methods to identify and counter the effects of \ac{ai}-generated content has become paramount. 

\begin{figure*}[!ht]
    \centering
    \includegraphics[width=.9\linewidth]{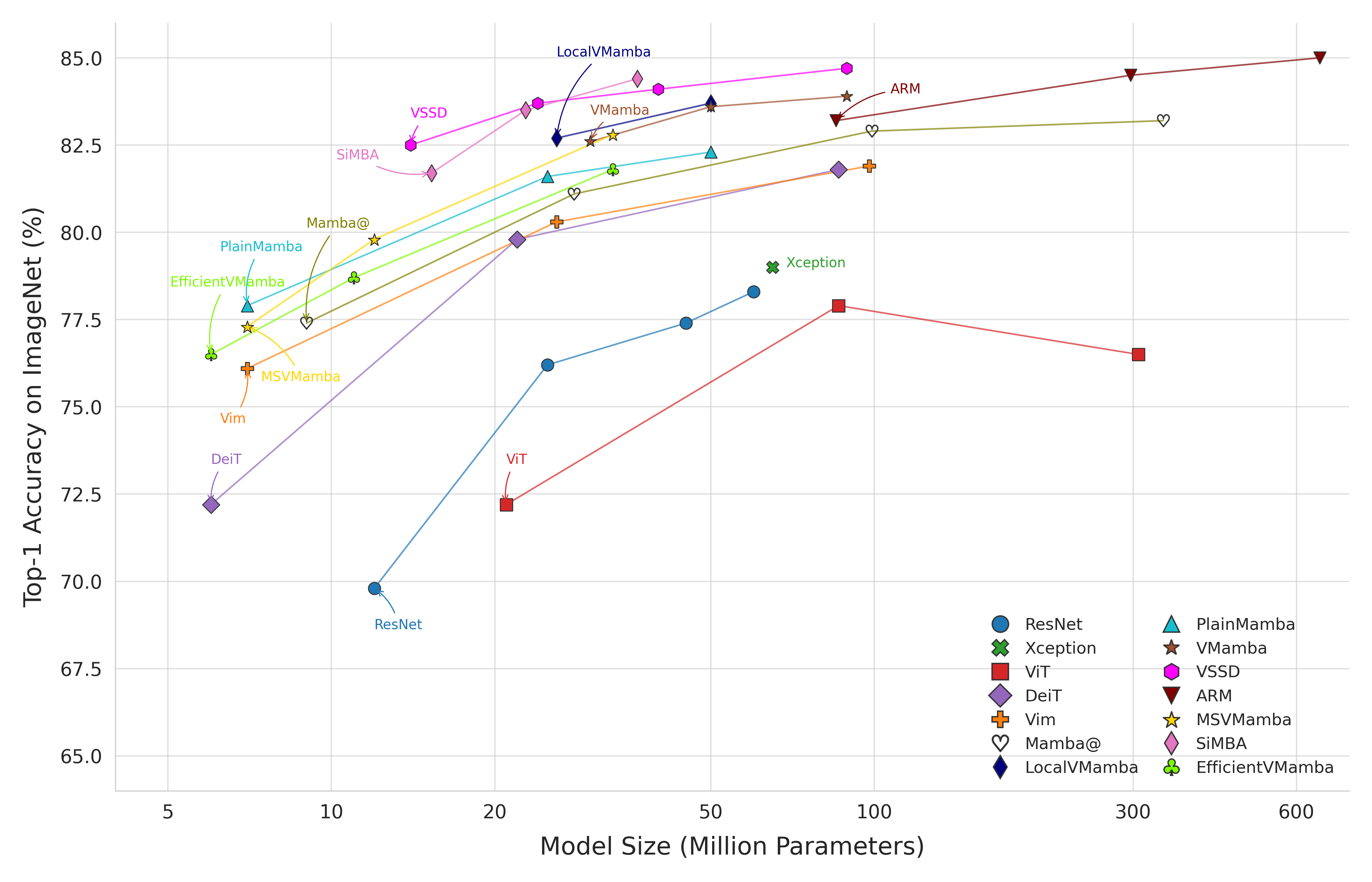}
    % \caption{Comparison of visual backbone architectures (ResNet, ViT, DeiT, Mamba-based models) in terms of model size (million parameters) and ImageNet top-1 accuracy. Different markers/colors represent model families, with dashed lines connecting variants (tiny, small, base, large) to show scalability. The plot highlights trade-offs between size and accuracy, illustrating efficiency and performance trends across architectures.}
    \caption{Comparison of backbone architectures.Each marker corresponds to a specific model family (ResNet, DeiT, VSSD, ...), and straight lines connect variants belonging to the same architectural family (tiny, small, base, large). The x-axis shows the number of parameters on a logarithmic scale, while the y-axis shows the ImageNet-1K top-1 accuracy reported for each model family in the original papers. The plot highlights trade-offs between size and accuracy, illustrating efficiency and performance trends across architectures.}
    \label{fig:VisualBackbone}
\end{figure*}

\noindent In recent years, computer vision has witnessed remarkable progress, fueled by the development of innovative architectures such as \acp{cnn}, \acp{vit}, and, more recently, \acp{vlm}, as illustrated in Figure~\ref{fig:VisualBackbone}. Figure~\ref{fig:VisualBackbone} provides an overview of the parameter–accuracy trade-off across several representative backbones. The x-axis reports the number of parameters on a logarithmic scale, while the y-axis indicates ImageNet-1K top-1 accuracy as reported in the original papers. Each marker corresponds to a model family (ResNet, DeiT, VSSD, …), and straight lines connect variants within the same architecture (e.g., tiny, small, base, large). This visualization highlights the efficiency–performance trade-offs characterizing different model families, revealing how design choices impact parameter budgets and achievable accuracy. These trends motivate our investigation of Vision Mamba architectures for synthetic-image detection, particularly in resource-constrained scenarios.

\noindent These models have led to breakthroughs in various tasks, from basic image classification to more complex challenges like detecting \ac{ai}-generated images~\cite{keita2025bi,chang2023antifakeprompt,wang2020cnn,tan2025c2p,xu2024fd,xu2023tall}. While \acp{cnn}, with their hierarchical feature extraction capabilities, long dominated the field, attention-based models like \acp{vit} offer a powerful alternative for capturing global relationships within visual data. Furthermore, the emergence of \acp{vlm} marks an exciting development, integrating visual perception with natural language understanding to enable more sophisticated multimodal tasks. The rise of \ac{ai}-generated images presents a significant challenge for image analysis and authentication. In response, a diverse set of detection techniques has emerged, encompassing \acp{cnn}, attention-based transformers, and \acp{vlm}. Each technique offers unique accuracy, interpretability, and scalability advantages but also has inherent limitations. In this rapidly evolving field, Mamba~\cite{gu2023mamba} has emerged as a versatile tool for various image analysis tasks, including classification, segmentation, medical imaging, object detection, and image restoration~\cite{ma2024u, zhang2024vim, liu2024vmamba, sun2021mamba, dong2024fusion}. However, its potential for identifying \ac{ai}-generated images remains relatively unexplored compared to established techniques. This paper aims to bridge this gap by rigorously evaluating Vision Mamba's efficacy in detecting \ac{ai}-generated images. We benchmark the performance of several Vision Mamba models against \ac{cnn}-based methods, Transformers, and \acp{vlm}, focusing on key metrics such as accuracy, efficiency, and generalizability across diverse image types and generative models. Through this comprehensive analysis, we aim to elucidate Vision Mamba's strengths and limitations relative to established methodologies. This research is crucial for enhancing detection capabilities in an age where distinguishing between authentic and \ac{ai}-generated content is increasingly challenging.

The remainder of this paper is structured as follows. Section~\ref{sec:relatedWork} concisely reviews related work on \acs{ai} image generation and detection techniques. Section~\ref{sec:mamba} describes the Mamba architecture and its adaptation to the vision domain. Section~\ref{sec:conVisMamba} describes the Vision Mamba models evaluated in our experimentation. Section~\ref{sec:empiricalStudy} presents an empirical analysis of Vision Mamba architectures for \ac{ai}-generated image detection. Finally, Section~\ref{sec:conclusion} concludes the paper with a summary of our findings and a discussion of future research directions.

\section{Related Works}
\label{sec:relatedWork}
\noindent Recent years have witnessed significant advancements in computer vision, fueled by innovative architectures like \acp{cnn}, \acp{vit}, diffusion, and \acp{vlm}. These models have propelled progress in generating and detecting synthetic images, leading to a surge of sophisticated techniques in both domains. This section provides a comprehensive review of these state-of-the-art methods, encompassing the creation and identification of synthetic images.

\subsection{Synthetic Image Generation}

Deep learning models for synthetic image generation have been around for some time. A pioneering method named \ac{gan} was introduced by Goodfellow {\it et al.}~\cite{goodfellow2014generative}, a neural network architecture for unconditional synthetic image generation. Then, innovative work followed, focusing on improving the learning process of \acs{gan}s, improving the quality and diversity of generated images, and conditional image synthesis. In recent years, text-image models have attracted interest following the introduction of diffusion models. Most recent diffusion-based image synthesis models, including Glide~\cite{nichol2021glide}, LDM~\cite{rombach2022high}, ADM~\cite{dhariwal2021diffusion}, DALL-E 3~\cite{dalle3_openai},  Midjourney v5~\cite{midjourney_v5},  Firefly~\cite{adobe_firefly}, Imagen~\cite{saharia2022photorealistic}, SDXL~\cite{podell2023sdxl}, SGXL~\cite{sauer2022stylegan} have demonstrated their ability to produce high-quality images. Diffusion models have also shown the ability to generate images in a wider range of categories and scenes than \acs{gan}s.

\noindent The following is an overview of recent advances in models for generating synthetic images. Ramesh {\it  et al.}~\cite{ramesh2021zero} introduced a transformer-based model to model text and image tokens as a unified data stream. They address memory constraints by compressing images into discrete tokens using a two-stage process involving a \ac{dvae}. Additionally, they enhance sample quality with contrastive reranking and optimize large-scale training with mixed-precision techniques and PowerSGD~\cite{vogels2019powersgd}. Drawing on transformer-based models, Saharia {\it  et al.}~\cite{saharia2022photorealistic} advanced the field with Imagen, a text-to-image diffusion model that combines large transformer language models with diffusion techniques to achieve high photorealism and accurate text-image alignment. The core innovation lies in using \acp{llm} like T5 to encode text, which significantly enhances image fidelity and alignment compared to scaling the diffusion model. Using dynamic thresholding and noise conditioning in Imagen improves image quality while maintaining alignment. Despite its advances, the model raises ethical concerns about dataset biases and societal impacts. Ding {\it et al.}~\cite{ding2022cogview2} introduced CogView2, a hierarchical transformer-based model, to address the slow generation and complexity issues in high-resolution text-to-image tasks. The approach involves pretraining a 6B-parameter transformer with a \ac{coglm}, which unifies autoregressive generation with bidirectional context-aware mask prediction. CogView2 achieves competitive performance with DALL-E-2 and supports interactive text-guided image editing. The method integrates hierarchical transformers and local parallel autoregressive generation for efficient high-resolution image synthesis, improving generation speed and image quality while maintaining flexibility in text-guided editing. Further pushing efficiency of training diffusion models, Rombach {\it et al.}~\cite{rombach2022high} introduced \acp{ldm}, which reduce computational demands by operating in the latent space of pretrained autoencoders rather than directly in pixel space. This approach maintains high visual fidelity by leveraging a perceptual autoencoder to compress images into a lower-dimensional latent space. \acp{ldm} incorporate cross-attention layers to support various conditioning inputs such as text or bounding boxes, enabling high-resolution synthesis and improved performance on tasks like image inpainting, text-to-image synthesis, and super-resolution. The method achieves state-of-the-art results with reduced computational costs compared to traditional pixel-based diffusion models. Nichol {\it et al.}~\cite{nichol2021glide} explored diffusion models for text-conditional image synthesis, comparing two guidance strategies: \acs{clip} guidance and classifier-free guidance. The authors also demonstrated that using classifier-free guidance, their 3.5 billion parameters model can be finetuned for image inpainting, allowing effective text-driven image editing. The study found that classifier-free guidance is preferred by human evaluators for both photorealism and caption similarity, often producing more photorealistic results than DALL-E, even with \acs{clip}. However, the model occasionally struggles with highly unusual prompts and has slower sampling times compared to \ac{gan} methods, which impacts its suitability for real-time applications. Towards a more personalized generation, Shi {\it et al.}~\cite{shi2024instantbooth} proposed InstantBooth, a personalized text-guided image generation approach that eliminates the need for test-time finetuning. The method uses a concept encoder to capture global embeddings and integrates adapter layers into a pre-trained diffusion model, preserving identity details while maintaining language coherence. InstantBooth generates images rapidly with a single forward pass, achieving results comparable to finetuning-based methods like Dreambooth~\cite{ruiz2023dreambooth}. In the context of artistic image generation, Xue {\it et al.}~\cite{xue2024raphael}  introduced RAPHAEL, a text-conditional image diffusion model that generates highly artistic images aligned with complex text prompts. The model employs stacked space-\ac{moe} and time-\ac{moe} layers to create billions of diffusion paths, each acting as a "painter" to map textual concepts to specific image regions. More recently, Dos {\it et al.}~\cite{dos2024synthetic} introduced \ac{sdgs}, a fully automated synthetic data generation system based on \ac{vae}. \Ac{sdgs} operates through three key functionalities: extracting data from multiple sources using the linked data (LD) paradigm, merging datasets to enhance information richness, and incorporating a feature engineering layer to optimize features for the \acs{vae} model. This architecture facilitates the creation of larger synthetic datasets by identifying relevant data sources, fusing datasets, and generating new data that resembles the constructed dataset.

\subsection{Detection of Synthetic Image}

\Ac{ai}-generated image detection has become an increasingly pressing challenge as synthetic image proliferates. Over recent years, several methods have been proposed to improve the detection of synthetic images for reliable detection as \ac{ai}-generated images evolve. One of the pivotal studies by Karageorgiou {\it et al.}~\cite{karageogiou2024evolution} examined the performance of state-of-the-art \ac{sid} methods in real-world scenarios, emphasizing the evolution of synthetic images as they circulate online. Their study highlights that existing detectors struggle to differentiate between real and synthetic images in the wild, particularly as these images undergo various post-processing operations over time. The authors introduce the \ac{fosid} to facilitate this evaluation, capturing synthetic images' temporal evolution and variability online. The study finds that most \acs{sid} methods are not well-calibrated for real-world conditions, with performance degrading as the time since the image's initial online appearance increases. To address this, they propose a \ac{rasid} approach, which maintains detection efficacy by leveraging near-duplicate images from earlier online instances, improving accuracy across several \acs{sid} methods. Building on detection challenges, Wang {\it et al.}~\cite{wang2023dire} demonstrate that genuine images exhibit high reconstruction errors when subjected to \ac{ddim} inversion. In contrast, counterfeit images generated from diffusion models show lower errors. However, \ac{dire} primarily focuses on the initial timestep \(x_0\), potentially overlooking valuable information from intermediate steps during the diffusion process. A significant drawback, however, is the inference time required, as the \ac{dire} framework necessitates invoking the \ac{adm} model at least 40 times to produce the \ac{dire} image. Cozzolino {\it et al.}~\cite{cozzolino2023raising} proposed a lightweight detection method for \ac{ai}-generated images using features extracted from the CLIP model. They demonstrated that a detector based on CLIP features, trained on a small dataset of images from a single generative model, generalizes well across diverse architectures, including DALL-E 3, Midjourney v5, and Firefly. This method avoids the need for extensive training on domain-specific data. Instead, it leverages paired real and fake images with the same textual description to train a simple \ac{svm} classifier. Their approach outperforms state-of-the-art methods, even under challenging post-processing conditions, demonstrating robustness and effective generalization.

\noindent Tan {\it et al.}~\cite{tan2025c2p} introduced C2P-\acs{clip}, an advanced approach designed to enhance \acs{clip}'s ability to detect \ac{ai}-generated images, particularly deepfakes. The method involves integrating a \ac{c2p} into the \acs{clip} model to embed category-specific concepts into the image encoder, thereby improving its generalization capabilities. The approach decodes \acs{clip} features into text, analyzing word frequencies to understand how \acs{clip} performs deepfake detection. By fine-tuning the image encoder with category-specific prompts through contrastive learning, C2P-\acs{clip} significantly improves detection performance across various unseen sources, achieving state-of-the-art results in generalizable deepfake detection. Through comprehensive experiments, the authors demonstrate that C2P-\acs{clip} significantly outperforms existing methods on several benchmark datasets, showing superior performance in detecting a wide range of \ac{ai}-generated manipulations. A further intriguing development comes from  Chang {\it et al.}~\cite{chang2023antifakeprompt}, who drew inspiration from the zero-shot capabilities of \acp{vlm} and proposed a method utilizing \acp{vlm} such as InstructBLIP. Their approach employs prompt tuning techniques to enhance deepfake detection accuracy on unseen data by framing the problem as a visual question-answering task. By fine-tuning soft prompts for InstructBLIP~\cite{instructblip}, the model can discern whether a query image is real or fake. In a novel shift,  Keita {\it et al.}~\cite{keita2025bi} introduces Bi-LORA, a novel approach for detecting synthetic images by reframing binary classification as an image captioning task. This method leverages \ac{vlm} combined with the \ac{lora} technique to enhance detection precision, particularly for diffusion-generated images. Bi-LORA significantly improves detection performance over traditional methods by generating a human-like label (textual description) that differentiates between real and synthetic images. The approach is memory-efficient, requiring fewer parameters to be tuned, making it a robust tool for identifying \ac{ai}-generated images across various datasets. Konstantinidou {\it et al.}~\cite{konstantinidou2024texturecrop} proposed TextureCrop, an innovative image pre-processing technique designed to enhance the accuracy of synthetic image detection by focusing on high-frequency texture components where generation artifacts are most prevalent. TextureCrop employs a sliding window approach to systematically analyze and crop texture-rich regions, filtering out areas with low texture variability. This method improves detection accuracy by selectively retaining critical image patches, resulting in significant performance gains across various detectors. The authors' extensive experiments demonstrate that TextureCrop outperforms traditional pre-processing methods, offering a more efficient and effective high-resolution synthetic image detection solution. Extending detection through image quality, Iliopoulou {\it et al.}~\cite{iliopoulou2024synthetic} proposed a novel method for detecting synthetic face images by leveraging deep learning-based image compression. Unlike traditional methods that rely on semantic features, this approach distinguishes between real and fake images based on the quality of their reconstruction post-compression. By utilizing a variational autoencoder (VAE) architecture, the method captures the response of face images to compression, with a focus on quality metrics such as \ac{mse}, \ac{ssim}, and \ac{psnr}. The technique is effective not only for \ac{gan}-generated images but also for those produced by diffusion models, offering a more generalized and computationally efficient alternative to existing detection methods. The results demonstrate high accuracy, with particular robustness against specific image manipulations, highlighting the method's potential for broader applications in synthetic image detection.

\noindent Following the trend towards utilizing pre-trained models, Chen {\it et al.}~\cite{chen2024guided}  introduced \ac{gff}, a novel approach for deepfake detection leveraging the pre-trained CLIP-ViT model. The \ac{gff} method enhances detection performance by integrating two key modules: the \ac{dfgm} and the \ac{fuseformer}. The \ac{dfgm} guides the frozen CLIP-ViT model to focus on deepfake-specific features, minimizing irrelevant information and maintaining generalization capabilities. \Ac{fuseformer} further improves detection by fusing low-level and high-level features extracted across different stages of the \ac{vit} encoder, ensuring comprehensive utilization of the extracted features. This approach achieves state-of-the-art performance by effectively harnessing the strengths of the pre-trained model with minimal additional training requirements. More recently, Huang {\it et al.}~\cite{huang2024ffaa} introduced OW-FFA-VQA, a novel task for face forgery analysis that moves beyond traditional binary classification by incorporating a \ac{vqa} framework. To support this task, they created the FFA-\ac{vqa} dataset using GPT-4, which includes diverse authentic and forged face images with detailed descriptions and forgery reasoning. They proposed \acs{ffaa}, a framework that combines a fine-tuned \ac{mllm} and a \ac{mids}. The \ac{mllm} is fine-tuned with hypothetical prompts to enhance analysis capabilities, while \ac{mids} selects the best-matching response, addressing fuzzy classification boundaries between real and forged faces. Extensive experiments demonstrate that \acs{ffaa} provides user-friendly, explainable results, significantly improving accuracy and robustness compared to existing methods.

\begin{figure*}[!htbp]
    \centering
    \begin{subfigure}[b]{0.3\textwidth}
        \includegraphics[width=\linewidth]{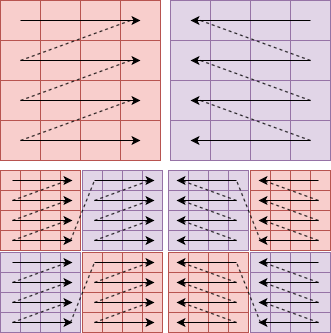}
        \caption{Local Scan}
    \end{subfigure}
    \begin{subfigure}[b]{0.3\textwidth}
        \includegraphics[width=\linewidth]{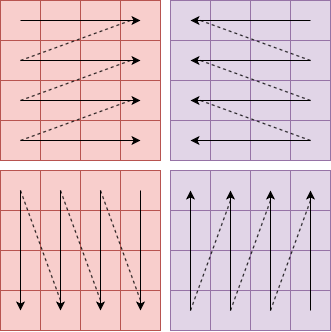}
        \caption{Cross Scan}
    \end{subfigure}
    \begin{subfigure}[b]{0.3\textwidth}
        \includegraphics[width=\linewidth]{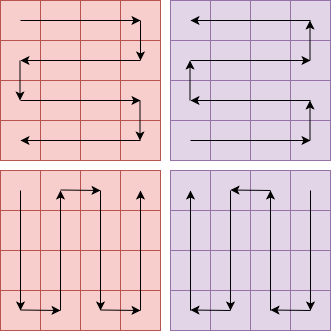}
        \caption{Continuous 2D Scan}
    \end{subfigure}
        \begin{subfigure}[b]{0.3\textwidth}
        \includegraphics[width=\linewidth]{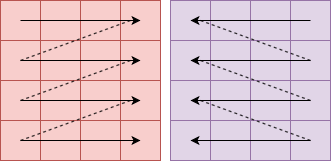}
        \caption{BiDirectional}
    \end{subfigure}
    \begin{subfigure}[b]{0.3\textwidth}
        \includegraphics[width=\linewidth]{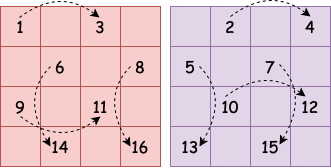}
        \caption{Efficient 2D Scan}
    \end{subfigure}
    \begin{subfigure}[b]{0.3\textwidth}
        \includegraphics[width=\linewidth]{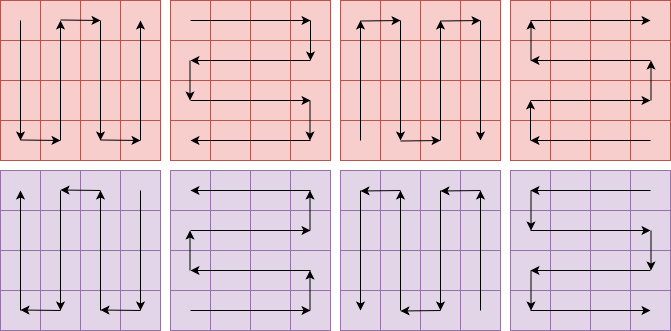}
        \caption{Zigzag Scanin}
    \end{subfigure}
    \begin{subfigure}[b]{0.3\textwidth}
        \includegraphics[width=\linewidth]{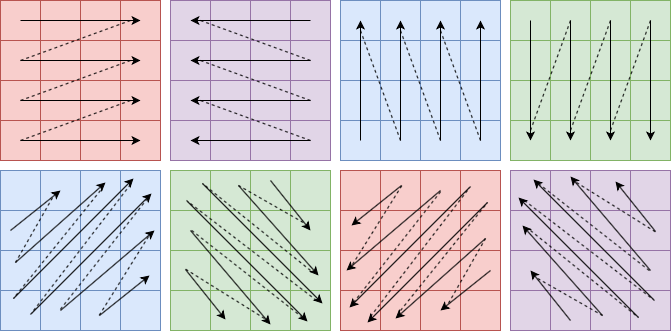}
        \caption{Omnidirectional Scan}
    \end{subfigure}
    \begin{subfigure}[b]{0.3\textwidth}
        \includegraphics[width=\linewidth]{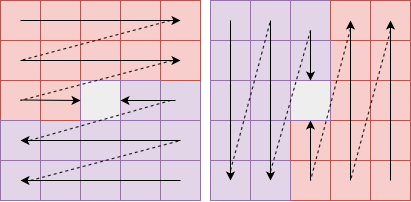}
        \caption{Selective Scan 2D}
    \end{subfigure}
    \caption{Illustration of various scanning strategies used in Mamba models to process visual inputs. Each scan strategy processes image sequences or spatial tokens in a distinct order. This balances computational efficiency, long-range dependency modeling, and fine-grained feature extraction. The scanning approaches shape the flow of information and the receptive field. They ultimately affect the model’s ability to understand and integrate visual content.}
    \label{fig:scan}
\end{figure*}

In brief, although existing work has made significant contributions to AI-generated image detection, none of it has explored the use of Visual Mamba as a backbone for this task. Yet Visual Mamba has proven its effectiveness in a number of fields, including object detection, segmentation and medical imaging. The lack of investigation into its application to AI-generated image detection is a major gap in the literature, justifying the need for this study.

\section{State Space Model Overview} 
\label{sec:mamba}

This section discusses the foundational principles and adaptations of the \ac{ssm}. We begin with an overview of the classical \acs{ssm} framework, which models dynamic systems through a latent state representation and linear ordinary differential equations. Next, we introduce the selective state space model, an extension designed to overcome the limitations of linear time-invariant assumptions by allowing greater flexibility for complex, non-stationary inputs. Finally, we explore the application of \acs{ssm}s in vision tasks, focusing on enhancing the Mamba block for efficient multi-dimensional data processing.

\subsection{State Space Model}
The \ac{ssm} is a mathematical framework used to describe the behavior of dynamic systems. It maps an input signal $x(t) \in \mathcal{R}$ to an output signal $y(t) \in \mathcal{R}$ through an implicit latent state vector $h(t) \in \mathcal{R}^{d \times 1}$. The dynamics of the system, parameterized by  ($\Delta, A, B, C$), can be formulated by the following linear ordinary differential equations.

\begin{equation}
    \begin{split}
        h'(t) &= Ah(t) + Bx(t)\\
        y(t) &= Ch(t) + Dx(t)
    \end{split}
\end{equation} where, $A \in \mathcal{R}^{d \times d}$ represents the system's internal transition matrix, $B \in \mathcal{R}^{d \times 1}$ is the input matrix that maps $x(t)$ to the hidden state, $C \in \mathcal{R}^{1 \times d}$ maps the hidden state $h(t)$ to the output, and $D \in \mathcal{R}$ directly maps the input to the output.\\\\
\noindent \textbf{Discretization of State-Space Model.}
To implement the \ac{ssm} in modern deep learning architectures, which operate in discrete time steps, the continuous model must be discretized. A common approach is the zero-order hold (ZOH) method, which discretizes the continuous parameters $A$ and $B$ using a timescale parameter $\Delta \in \mathcal{R}$. The discretized form of the parameters is given by:
\begin{equation}
    \begin{split}
        \bar{A} &= exp(\Delta A)\\
        \bar{B} &= (\Delta A)^{-1}(exp(\Delta A) - I) \cdot \Delta B  \approx \Delta B
    \end{split}
\end{equation}
With these transformations, the continuous system is converted into its discrete form:

\begin{equation}
    \begin{split}
        h_t &= \bar{A}h_{t-1} + \bar{B}x_t \\
        y_t &= Ch_t + Dx_t 
    \end{split}
    \label{eq:dssm}
\end{equation} where $h_t$ and $x_t$ represent the hidden state and input at the $t\textsuperscript{th}$ time step, respectively. This discrete formulation allows the \ac{ssm} to be used in various computational models, including neural networks.\\\\

\noindent \textbf{Computation.}
Furthermore, the iterative process defined by Equation~\eqref{eq:dssm} presents computational challenges. This recursive process can be reformulated to enhance efficiency as a global convolution operation (denoted by $*$), leveraging parallel computation, as the expression in Equation~\eqref{eq:cssm}.
\begin{equation}
    \begin{split}
       \bar{K} &= (C\bar{B},C\bar{AB},\cdots,\bar{A}^{L-1}\bar{B})\\
        y &= x*\bar{K}
    \end{split}
    \label{eq:cssm}
\end{equation}
where $L$ denotes the length of the input sequence $x$, and $\bar{K}$ represents the \ac{ssm} convolution kernel, or filter.

\noindent Given $\bar{K}$, the convolution operation in Equation~\eqref{eq:cssm} can be computed efficiently using the \ac{fft}.

\begin{figure*}[ht]
      \centering
        \includegraphics[width=.9\linewidth]{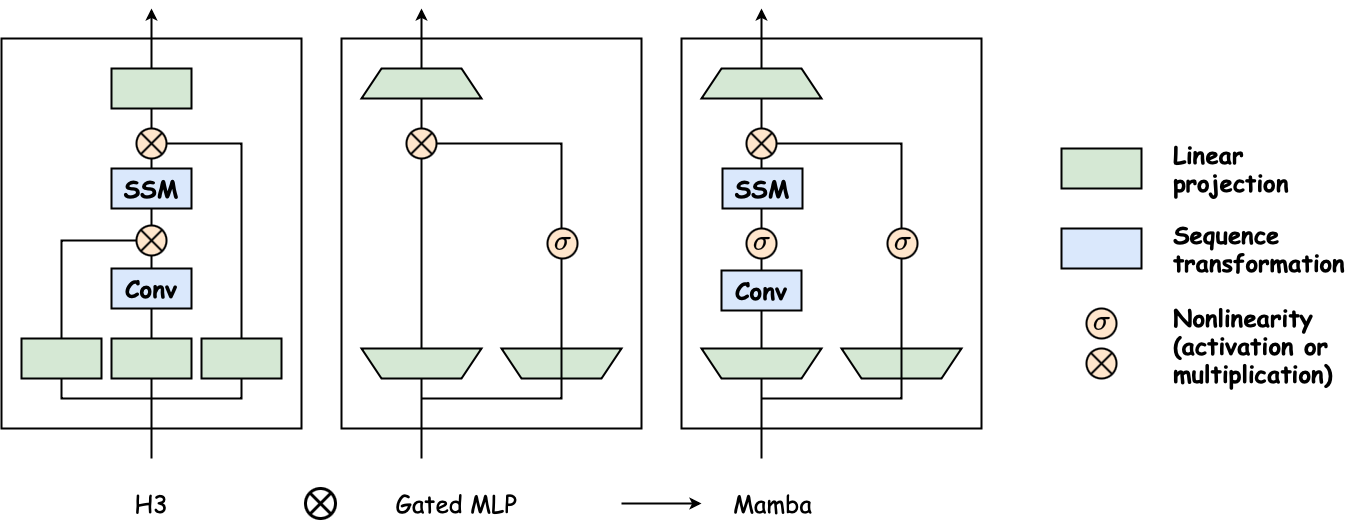}
        \caption{Architecture of Mamba block~\cite{gu2023mamba}}
      \label{fig:mamba}
\end{figure*}

\subsection{Selective State Space Model}

While the classical \acs{ssm} provides a robust framework for analyzing dynamic systems, it operates under the assumption of linear time-invariant (LTI) dynamics. This implies that the parameters $A$, $B$, and $C$ remain constant over time, limiting the model's flexibility when dealing with complex, non-stationary input signals.

\noindent To address this limitation, Gu {\it et al.}~\cite{gu2023mamba} introduced the selective \ac{ssm} (i.e., Mamba), illustrated in Figure~\ref{fig:mamba}, where key parameters are made input-dependent. Specifically, the matrices $B$ and $C$, as well as the timescale parameter $\Delta$, are functions of the input $x$.
 
\begin{equation}
    \begin{split}
        B &= s_B(x) \\
        C &= s_C(x) \\
        \Delta &=  	\tau_\Delta (\Delta + s_\Delta(x))  
    \end{split}
\end{equation} where $s_B(x)$ and $s_C(x)$ linearly project $x$ into a $N$-dimension space. This modification enhances the adaptability of the model to changing input dynamics. 

\noindent The selective \acs{ssm} can be expressed as:

\begin{equation}
    \begin{split}
        h_i &=  \bar{A}_ih_{i-1} + \bar{B}_ix_i\\
        y_i &= C_ih_i + Dx_i 
    \end{split}
\end{equation}

\noindent In this formulation, the discretized parameters $\bar{A}_i$ and $\bar{B}_i$  are now functions of the input $x_i$, enabling greater model flexibility. Unlike the classical \ac{ssm}, which relies on fixed parameters, the selective \ac{ssm} adapts its behavior dynamically based on the current input. \\\\

\noindent \textbf{Reformulation of the selective SSM.} To further enhance computational efficiency and simplify the model's structure, Mamba introduces several key modifications:
\begin{enumerate}
    \item \textbf{Diagonalization of $\bar{A}_i$:} The matrix $\bar{A}_i$ is assumed to be diagonal, meaning that each element of the hidden state is updated independently. Therefore, $\bar{A}_ih_{i-1}=\tilde{A}_i\odot h_{i-1}$, where $\tilde{A}_i = $ diag($\bar{A}_i$). 
    This results in the following update for $h_i$:
    \begin{equation}
        h_i = \tilde{A}_i \odot h_{i-1} + B_i(\Delta_i \odot x_i)
    \end{equation}
    Here, $\tilde{A}_i$ represents the diagonal elements of $A_i$, and $\odot$ denotes the element-wise (Hadamard) product.

    \item \textbf{Input scaling:} Given $\bar{B}_i = \Delta_i B_i$ and $\Delta_i \in \mathcal{R}$, we have  $\bar{B}_i x_i = \Delta_i B_i x_i = B_i (\Delta_i x_i) = B_i (\Delta_i \odot x_i)$. The term $B_i(\Delta_i \odot x_i)$ scales the input $x_i$ by the input-dependent timescale $\Delta_i$.

    \item \textbf{Element-wise input-output interaction:} $Dx_i =D \odot x_i$. The direct influence of the input $x_i$ on the output $y_i$ is retained as $D \odot x_i$, where $D$ is also applied element-wise.
\end{enumerate}
Thus, the selective \acs{ssm} is reformulated as:
\begin{equation}
    \begin{split}
         h_i &= \tilde{A}_i \odot h_{i-1} + B_i(\Delta_i \odot x_i)\\
        y_i &= C_ih_i + D \odot x_i
    \end{split}
\end{equation}
where $B_i$, $C_i$, and $\Delta_i$ are input-dependent and derived using projection matrices. This formulation allows the selective \acs{ssm} to handle complex data sequences in deep learning architectures, making it a versatile and powerful tool for modern computational tasks.\\\\
\noindent \textbf{Handling Input Sequences.}
To handle multi-dimensional input sequences $x \in \mathcal{R}^{N \times C}$, where $x_i \in \mathcal{R}^{1 \times C}$ represents a vector of inputs across multiple channels, Mamba extends the selective \acs{ssm} to operate independently over each channel. 

\begin{table*}[!htpb]
% \caption{Features and ImageNet-1K Classification Accuracy of Tiny Vision Mamba Models.}
\caption{Comparison of tiny Vision Mamba models, detailing their main architectural contributions, scanning strategies, and ImageNet-1K accuracy. By comparing these design choices along with their reported accuracies, this table highlights the rapidly evolving landscape of Mamba-based vision architectures and the influence of different scanning mechanisms on downstream recognition performance.}
\label{tab:MambaModels}
\renewcommand{\arraystretch}{1.2}
\begin{adjustbox}{width=\linewidth}
\begin{tabular}{l|c|l|l|c}
\toprule
Models & Year & Main contribution & Scan order & Accuracy \\ \midrule

Vim~\cite{zhu2024vision} & 2024 & Integrated forward and backward SSM paths. & BiDirectional & 76.1\% \\ \midrule

VMamba~\cite{liu2024vmamba} & 2024 & Introduced Cross-scan Module (CSM). & Cross-Scan & 82.6\% \\ \midrule

PlainMamba~\cite{yang2024plainmamba} & 2024 & \begin{tabular}[c]{@{}l@{}}Non-hierarchical architecture using zigzag scanning and \\ direction-aware updates for better feature fusion.\end{tabular} & Continuous 2D scan & $^\star$ \\ \midrule

Localmamba~\cite{huang2024localmamba} & 2024 & \begin{tabular}[c]{@{}l@{}}Divided images into local windows for directional SSM, while \\preserving global operations.\end{tabular} & Local scan & 82.7\% \\ \midrule

MambaVision~\cite{hatamizadeh2025mambavision} & 2024 & \begin{tabular}[c]{@{}l@{}}Hybrid Mamba-Transformer with redesigned Mamba block \\  and self-attention for long-range dependencies.\end{tabular} & Single direction & 82.30\% \\ \midrule

FractalMamba~\cite{tang2024scalable} & 2024 & \begin{tabular}[c]{@{}l@{}}Used fractal scanning curves to maintain spatial proximity,\\ improving pattern modeling and reducing redundancy.\end{tabular} & Fractal scanning & 82.7\% \\ \midrule

Mamba®~\cite{wang2024mamba} & 2024 & \begin{tabular}[c]{@{}l@{}}Incorporated register tokens throughout the token sequence \\ for improved final image predictions.\end{tabular} & BiDirectional & 77.4\% \\ \midrule

ARM~\cite{ren2024autoregressive} & 2024 & \begin{tabular}[c]{@{}l@{}}Enhanced pretraining using an autoregressive strategy treating\\  neighboring patches as prediction units.\end{tabular} & \begin{tabular}[c]{@{}l@{}}Uni-directional scan \\ Cross-scan\end{tabular} & $^\star$ \\ \midrule

Efficientvmamba~\cite{pei2024efficientvmamba} & 2024 & \begin{tabular}[c]{@{}l@{}}Efficient 2D Scanning (ES2D) technique with atrous sampling \\ to reduce computational cost and enhance feature fusion.\end{tabular} & Efficient 2D Scan & 76.5\% \\ \midrule

SiMBA~\cite{patro2024simba} & 2024 & \begin{tabular}[c]{@{}l@{}}Introduced \acs{einfft} for improved computational efficiency \\ and stability using Fourier Transform.\end{tabular} & Frequency-domain scan & $^\star$ \\ \midrule

Vim-F~\cite{zhang2024vim} & 2024 & \begin{tabular}[c]{@{}l@{}}Introduced frequency-domain component using FFT and \\  overlapping convolutions to capture spatial correlations.\end{tabular} & Frequency-domain scan & 76.0\% \\ \midrule

MSVMamba~\cite{shi2024multi} & 2024 & \begin{tabular}[c]{@{}l@{}}Introduced MS2D scanning with downsampling to improve \\ long-range dependency learning and reduce computational cost.\end{tabular} & Multi-Scale 2D scan & 82.8\% \\ \midrule

Mamba-ND~\cite{li2024mamba} & 2024 & \begin{tabular}[c]{@{}l@{}}Extended Mamba to multi-dimensional data using \\ alternating-directional orderings for improved performance.\end{tabular} & BiDirectional & $^\star$ \\ \midrule

VSSD~\cite{shi2025vssd} & 2024 & Introduced a non-causal variant of the state space duality model. & Multi-scan strategies & \bf 84.1\% \\ \bottomrule

\end{tabular}
\end{adjustbox}

\vspace{.1in}
 \begin{flushleft}
\scriptsize $^\star$ No Tiny Variant. 
\end{flushleft}

\end{table*}

\subsection{Mamba in General Vision Tasks}
\label{sec:sssm}
While the original Mamba block was designed for one-dimensional sequences, its direct application to vision-related tasks is limited due to the multi-dimensional nature of visual inputs such as images and videos. To extend Mamba's autoregressive formulation to these tasks, it is essential to enhance its scanning mechanism and architecture to process multi-dimensional data efficiently. Given the critical role of scanning strategies in visual tasks, we provide a comprehensive overview of existing 2D scanning mechanisms in Figure \ref{fig:scan}, highlighting their relevance and limitations.

\begin{figure*}[!htbp]
    \centering
    \begin{subfigure}{0.3\textwidth}
        \includegraphics[width=\textwidth]{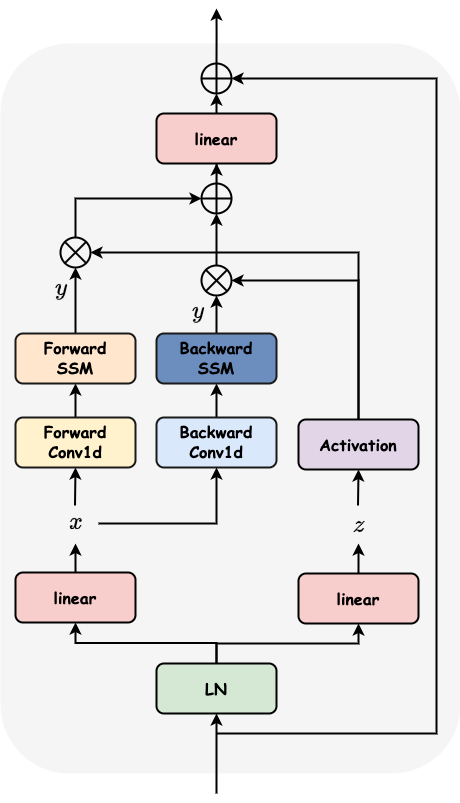}
        \caption{Vim block.}
    \end{subfigure}
    \begin{subfigure}{0.3\textwidth}
        \includegraphics[width=\textwidth]{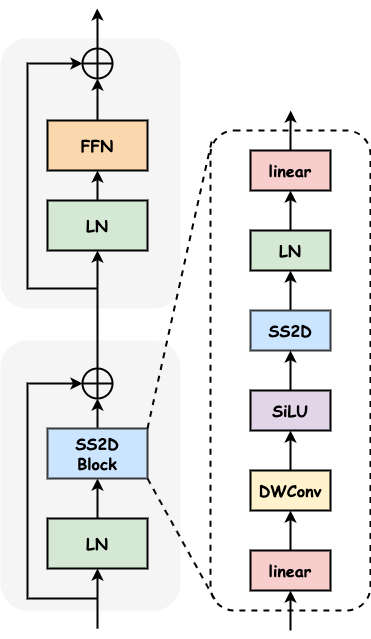}
        \caption{VSSD block.}
    \end{subfigure}
        \begin{subfigure}{0.3\textwidth}
        \includegraphics[width=\textwidth]{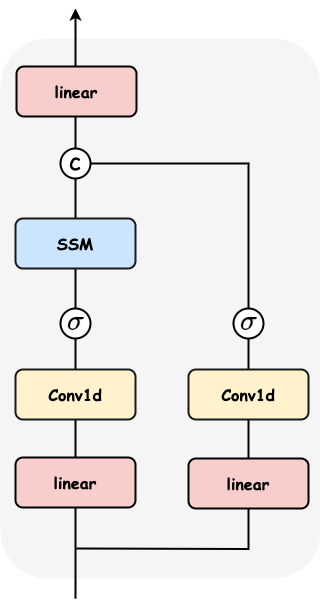}
        \caption{MambaVision mixer.}
    \end{subfigure}
    \begin{subfigure}{.8\textwidth}
        \includegraphics[width=\textwidth]{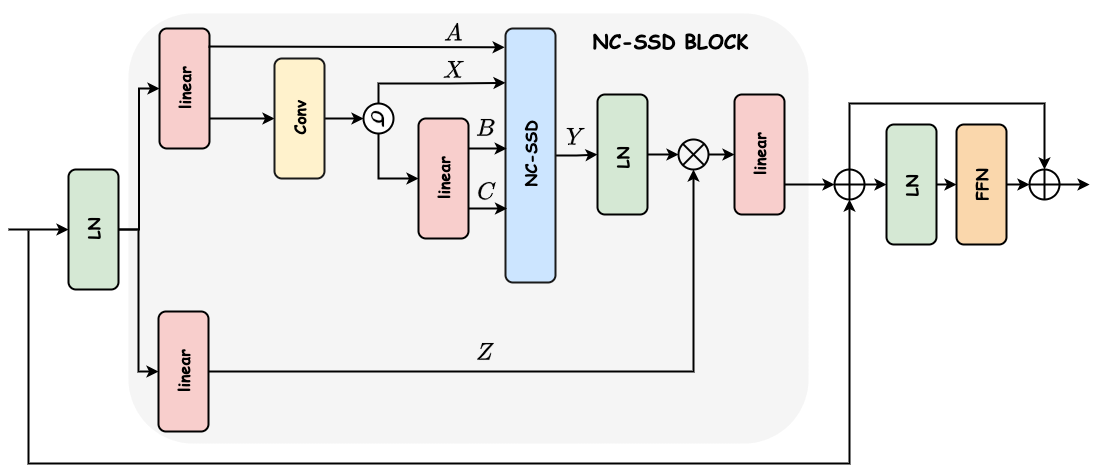}
        \caption{SSD block.}
    \end{subfigure}
    \caption{Illustration of representative Visual Mamba blocks, including Vim~\cite{zhu2024vision}, VSSD~\cite{liu2024vmamba}, SSD~\cite{shi2025vssd}, and the MambaVision mixer~\cite{hatamizadeh2025mambavision}. The figure delineates the architectural distinctions among these variants, showing how each design tailors the Mamba state-space paradigm to visual processing via dedicated token mixing, spatial scanning approaches, and feature refinement techniques. Together, these modules exemplify the breadth of ways in which Mamba-based architectures advance the selective state-space framework to enable efficient long-range modeling in computer vision.}
    % \caption{Visual Mamba blocks, including Vim~\cite{zhu2024vision}, VSSD~\cite{liu2024vmamba} SSD~\cite{shi2025vssd}, and MambaVision mixer~\cite{hatamizadeh2025mambavision}.}
    \label{fig:architecture}
\end{figure*}

\noindent Due to its promising potential for long sequence modeling (high-resolution image in our context), there has been significant interest from researchers in applying Mamba to various vision tasks~\cite{zhu2024vision,liu2024vmamba,hatamizadeh2025mambavision,yue2024medmamba,yang2024plainmamba,li2024mamba,huang2024localmamba,pei2024efficientvmamba}. For example, Vim~\cite{zhu2024vision}, in Figure~\ref{fig:architecture}a, introduces an architecture that integrates both forward and backward SSM paths within its blocks, enhancing the processing of image patch sequences, and employs a bidirectional scan strategy (Figure~\ref{fig:scan}(d)).

\noindent VMamba~\cite{liu2024vmamba}, in Figure~\ref{fig:architecture}(b), introduces a cross-scan module (CSM) to address the challenge of capturing direction-sensitive dependencies in 2D images by scanning patches in multiple directions: left to bottom right,  bottom right to top left, top right to bottom left, and bottom left to top right (Fig.\ref{fig:scan}(b)). This approach enhances Mamba's ability to integrate spatial information across an image, improving its performance in vision tasks. PlainMamba~\cite{yang2024plainmamba} introduces a non-hierarchical architecture that adapts Mamba's selective scanning for 2D images using a zigzag scanning (Fig.\ref{fig:scan}(f)) technique and direction-aware updates to maintain spatial continuity and enhance multi-level feature fusion. Mamba-ND~\cite{li2024mamba} extends Mamba to multi-dimensional data using alternating-directional orderings for improved performance. Localmamba~\cite{huang2024localmamba} addresses the disrupted spatial dependencies in Vim~\cite{zhu2024vision} and VMamba~\cite{liu2024vmamba} by dividing the input image into local windows for directional \acs{ssm} (Fig.\ref{fig:scan}(a)) while preserving global \acs{ssm} operations. It also introduces a spatial and channel attention module to enhance feature integration and reduce redundancy, along with a strategy to optimize scan directions for each layer, improving computational efficiency. 

\noindent MambaVision~\cite{hatamizadeh2025mambavision} introduces a hybrid Mamba-Transformer backbone tailored for vision tasks. It redesigns the Mamba block for efficient visual feature modeling and integrates self-attention blocks at the final layers to capture long-range dependencies. FractalMamba~\cite{tang2024scalable} improves \acs{ssm} performance by using fractal scanning curves for patch serialization. This approach maintains spatial proximity and adapts to different resolutions, reducing redundancy and enhancing the modeling of complex patterns. Mamba®~\cite{wang2024mamba} introduces a refinement to Vision Mamba by incorporating register tokens throughout the token sequence and concatenating them at the end to improve image representation for final predictions, as opposed to appending them only at one end as in {\it Darcet et al.}~\cite{darcet2023vision}. 

\noindent ARM~\cite{ren2024autoregressive} enhances Mamba's pretraining by using an autoregressive strategy that treats neighboring image patches as prediction units. This method significantly boosts visual performance and scalability. Efficientvmamba~\cite{pei2024efficientvmamba} introduces the Efficient 2D Scanning (ES2D) (Fig.\ref{fig:scan}(e)) technique, which uses atrous sampling to reduce computational costs by extracting global features. These features are processed alongside local features in an EVSS block, efficiently combining global and local information for improved visual state space modeling. SiMBA~\cite{patro2024simba} addresses instability in scaling Mamba by introducing \ac{einfft} for sequence modeling. \acs{einfft} uses Fourier Transform and Einstein Matrix Multiplication in the frequency domain to enhance computational efficiency and stability by ensuring all eigenvalues of the evolution matrix are negative real numbers. Vim-F~\cite{zhang2024vim} addresses the limitations of \acs{vim} by introducing a frequency-domain component through the \ac{fft}, enhancing global spatial understanding. It also removes the need for position embedding, using overlapping convolutions to capture spatial correlations between tokens better. MSVMamba~\cite{shi2024multi} introduces a \ac{ms2d} scanning strategy that downsamples the image in multiple directions to improve long-range dependency learning and reduce computational costs. Additionally, it incorporates a \ac{convffn} to enhance channel mixing and local feature extraction. In order to address the limitations of causal SSMs in vision tasks, VSSD~\cite{shi2025vssd} introduces a non-causal variant of the State Space Duality (SSD) model. VSSD discards token dependencies on previous tokens to achieve non-causality and integrates multi-scan strategies. As a result, performance and efficiency are enhanced. Table~\ref{tab:MambaModels} summarizes the models discussed above. 

\section{Considered Vision Mamba Models}
\label{sec:conVisMamba}

This section describes the three Vision Mamba models evaluated in this paper for \ac{ai}-generated image detection.

\subsection{Vim}

\begin{figure}[!htbp]
      \centering
        \includegraphics[width=.6\textwidth]{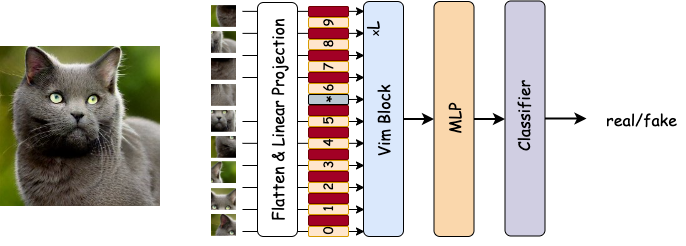}
        \caption{Refined Vim~\cite{zhu2024vision}  architecture for AI-generated image detection.}
      \label{fig:vim}
\end{figure}

Vim~\cite{zhu2024vision}  introduces the first pure \acs{ssm}-based model for vision tasks. The authors highlight two major challenges of applying \acs{ssm} to vision tasks: modeling uni-directionality and lack of location awareness. Vim incorporates bidirectional \acs{ssm} and positional embedding techniques to overcome these challenges. As depicted in Figure~\ref{fig:vim}, Vim first transforms a multi-dimensional image \(X \in \mathcal{R}^{H \times W \times C}\) into $N$ non-overlapping patches \(X_p \in \mathcal{R}^{J \times (P^2 \cdot C)}\), where \( N = \frac{HW}{P^2} \), ($H,W$) are the size of the input image, $C$ is the number of channels and $P$ is the size of image patch. Following the transformer's positional embedding approach, Vim linearly projects $X_p$ into a latent vector of size \( D \). It adds a positional embedding \( E_{pos} \in \mathcal{R}^{(J+1) \times D} \) to retain the spatial information and also uses a class token (CLS) to represent the entire patch sequence. Then, the resulting sequence token sequence is fed into $l$ layers of Vim block (Figure~\ref{fig:architecture}(a)), producing the output $O$. In contrast to the standard Mamba block, Vim employs a bidirectional \acs{ssm} block, where the inputs of the Vim block are processed from the forward and backward directions, respectively. The outputs of the forward and backward processes are computed through \acs{ssm}. Then, the outputs $y$ of both processes are selected by the gating signal $z$ and added together to obtain the output token sequence.

\noindent Bidirectional encoding in Vim's visual data processing poses challenges in terms of computational efficiency and understanding of the global context. While it improves information integration from multiple directions, it significantly increases the computational load, potentially slowing down training and inference. In addition, it remains challenging to achieve coherent global understanding, as some global context may be lost in the process. This problem is exacerbated in visual data, where the non-causal nature of the information means that applying techniques like Mamba directly to patches or flat images results in a limited receptive field, as relationships with unscanned patches cannot be effectively estimated.

\subsection{MambaVision}

\begin{figure*}[!htbp]
      \centering
        \includegraphics[width=\textwidth]{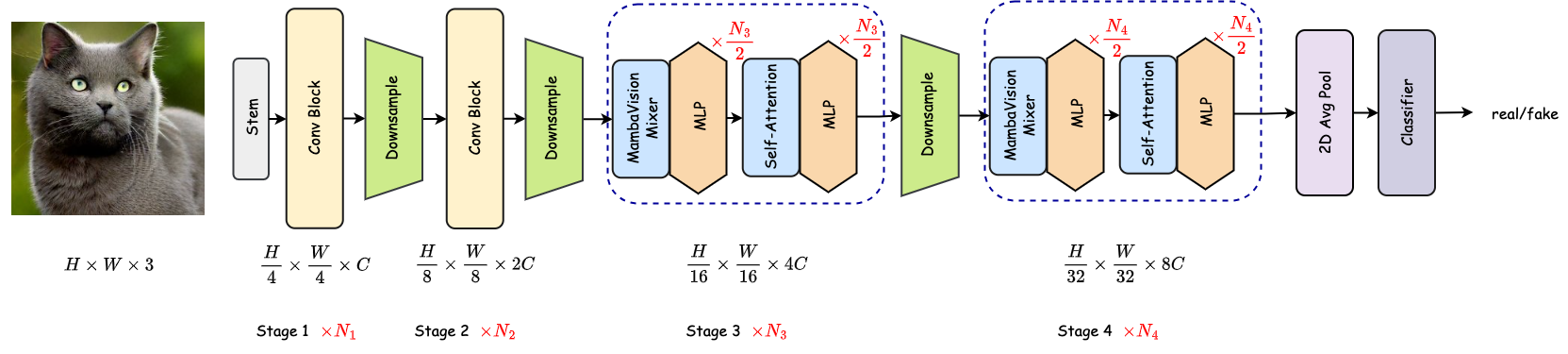}
        \caption{Refined MambaVision~\cite{hatamizadeh2025mambavision} architecture for AI-generated image detection.}
      \label{fig:mambaVision}
\end{figure*}

MambaVision~\cite{hatamizadeh2025mambavision} introduces a hybrid architecture optimized for vision tasks. The design combines \ac{cnn} layers for local feature extraction and a redesigned vision-friendly Mamba-based mixer integrated with Transformer blocks to capture global and spatial relationships. As depicted in Figure~\ref{fig:mambaVision}, the hierarchical structure of the model consists of $4$ different stages.

\noindent The first two stages leverage \acs{cnn} layers for high-resolution feature extraction. These stages use a residual block design that ensures efficient processing while preserving important spatial information. In between stages, a downsampler reduces the resolution of the feature maps to focus on progressively coarser details. The stem module, at the beginning of the approach, splits a given image of size $H \times W \times 3$ into overlapping patches with size $\frac{H}{4} \times \frac{W}{4} \times C $ and then projects them into a $C$ dimensional embedding space.

\noindent In stages $3$ and $4$, MambaVision introduces an innovative mixer block, as illustrated in Figure~\ref{fig:architecture}(c). In contrast to the original Mamba formulation, which is based on causal convolutions, MambaVision removes this constraint to align more closely with the non-sequential nature of image data. In addition, a symmetric branch has been incorporated into the mixer design. This branch complements the Mamba-based component's sequential modeling, focusing on spatial features through the implementation of additional convolutional operations. The outputs from both branches are combined, resulting in richer feature representations that account for sequential and spatial information.
The final layers in these stages incorporate Transformer-based self-attention mechanisms. These layers allow the model to capture long-range dependencies, ensuring a robust understanding of the global context. This is particularly beneficial for vision tasks, where the relationship between distant regions in an image can significantly influence the predictions.

\subsection{Visual State Space Duality (VSSD)}

\begin{figure*}[!htbp]
      \centering
        \includegraphics[width=\textwidth]{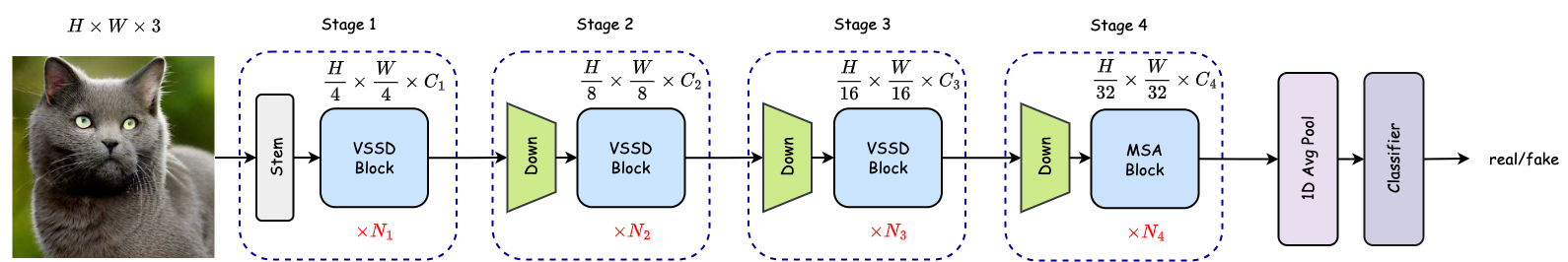}
        \caption{Refined VSSD~\cite{shi2025vssd} architecture for AI-generated image detection.}
      \label{fig:vssd-model}
\end{figure*}

VSSD~\cite{shi2025vssd} introduces a novel approach to applying state space models (SSMs) to vision tasks. An overview of the approach is illustrated in Figure~\ref{fig:vssd-model}. The authors identify two challenges in applying SSD/SSMs to non-causal image data: (1) causal constraints restrict information flow, and (2) flattening 2D feature maps into 1D sequences disrupts their intrinsic structural information. VSSD employs the  Non-Causal SSD (NC-SSD) to address these challenges and introduce other enhancements.

\noindent To handle vision tasks, the model first flattens a given 2D image through the stem module into a sequence of tokens. The NC-SSD integrates forward and backward scanning into a global hidden state $h$ shared across tokens, preserving the structural relationships of the 2D map. In addition, NC-SSD eliminates the causal mask and allows parallel computation, improving the efficiency of training and inference.

\noindent Similar to Mamba2's~\cite{dao2024transformers} SSD implementation, NC-SSD works linearly. However, it modifies the hidden state computation to remove causal dependencies. Instead of requiring token-wise repetition, all tokens share a common hidden state. The token sequence is fed into the VSSD block, as shown in Figure~\ref{fig:architecture}(d), which includes a depth-wise convolution (DWConv), a local perception unit (LPU), and a feed-forward network (FFN) for local feature extraction and channel-wise interaction. The final stage replaces NC-SSD with multi-head self-attention (MSA) for high-level feature processing. Compared to existing SSM-based models such as VMamba, VSSD achieves improved accuracy and efficiency.

\section{Empirical Study}
\label{sec:empiricalStudy}

This section presents the empirical results of our proposed synthetic image detection methodology. We first outline the experimental setup, followed by a detailed analysis of cross-generator evaluations.

\subsection{Experimental Setup}
\label{sec:expSetup}
\noindent \textbf{Dataset.} To assess the efficacy of Vision Mamba in detecting \ac{ai}-generated images, we employed three distinct datasets. These datasets were carefully curated to encompass a diverse range of real and synthetic images, generated through various techniques and spanning multiple image domains and attack scenarios. A detailed description of the composition and characteristics of each dataset is provided below. Table~\ref{tab:datasets} summarizes the key characteristics of the three datasets used in this study.

\begin{table*}[]
\caption{Key characteristics of the three datasets used in this study.}
\label{tab:datasets}
\begin{adjustbox}{width=\linewidth}
\begin{tabular}{lcc
>{\columncolor[HTML]{CCFFCC}}c 
>{\columncolor[HTML]{feb3b1}}c c}
\hline
Dataset       & \#Real     & \#Generated & Source of  Real Image & Generation Method                   & Year \\ \hline
BedRoom~\cite{ricker2022towards} &
  420,000 &
  530,000 &
  LSUN &
  \begin{tabular}[c]{@{}c@{}}LDM, ADM, DDPM, iDDPM, PNDM, \\SDv1.4, GLIDE  StyleGAN, ProGAN, \\Diff-ProjectedGAN,  ProjectedGAN, Diff-StyleGAN2\end{tabular} &
  2022 \\ 
AntifakePrompt~\cite{chang2023antifakeprompt}   &     125,870    &    136,500    & COCO, Flickr              & \begin{tabular}[c]{@{}c@{}}SD2, SDXL, IF, DALLE2, SGXL, ControlNet, \\LaMA, SD2IP, LTE, SD2SR,Deeper-Forensics, \\Adver, Backdoor, Data Poisoning\end{tabular} & 2023 \\ 

UniversalFakeDetect~\cite{ojha2023towards}   &   410,681       &    416,673    & LSUN, ImageNet, LAION              & \begin{tabular}[c]{@{}c@{}}ProGAN, CycleGAN, BigGAN, StyleGAN,\\ GauGAN, StarGAN, DeepFakes, SITD, SAN, \\CRN, IMLE, Guided, LDMs, GLIDEs, DALLE\end{tabular} & 2023 \\ 
% GenImage~\cite{zhu2024genimage}   &    1,331,167     &    1,350,000    & ImageNet             & \begin{tabular}[c]{@{}c@{}} Midjourney, SD V1.4, SD V1.5, ADM\\ GLIDE, Wukong, VQDM, BigGAN\end{tabular} & 2023 \\ 
  \hline
\end{tabular}
\end{adjustbox}
\end{table*}

\begin{enumerate}
    \item {\bf AntiFakePrompt} dataset, introduced by Chang {\it et al.}~\cite{chang2023antifakeprompt}, encompasses a wide range of domains. Real images are sourced from MS COCO (image captions) and Flickr (social media). Synthetic images include data generated by text-to-image models (SD2, SDXL, Imagen for Fashion, DALL-E 2, StyleGAN-XL), image stylization (ControlNet), image inpainting (LaMa, SD2 Inpainting), and super-resolution models (LTE, SD2-Inpainting). The dataset also covers deep forensic analysis and various attack scenarios, including adversarial attacks, backdoor attacks, and data poisoning. For training, we utilize 60,000 real images from MS COCO and 60,000 synthetic images, evenly split between SD2 and LaMa. During testing, 3,000 images from all subsets are used to ensure comprehensive evaluation across different domains, as detailed in Table~\ref{tab:antifakepromt}.

    \item {\bf Bedroom} dataset, introduced by Ricker {\it et al.}~\cite{ricker2022towards}, incorporates real images sourced from the LSUN Bedroom dataset~\cite{yu2015lsun}. We curated a collection of images generated by five distinct diffusion models, all trained on the LSUN Bedroom dataset. Four subsets of generated images (\acs{adm}~\cite{dhariwal2021diffusion}, \acs{ddpm}~\cite{ho2020denoising}, \acs{iddpm}~\cite{nichol2021improved}, and \acs{pndm}~\cite{liu2022pseudo}) are generated by unconditional diffusion models. The fifth subset, \acs{ldm}~\cite{rombach2022high}, is generated by a text-to-image diffusion model. To further evaluate generalization to text-to-image generation, we expanded the dataset by incorporating two additional models: \acp{sd}~\cite{rombach2022high} and GLIDE~\cite{nichol2021glide}. The text prompt used to generate these images was "A photo of a bedroom." In our study, all subsets consist of 42,000 generated images and their corresponding real samples from the LSUN Bedroom dataset. Each subset is divided into 40,000 images for training, 1,000 for validation, and 1,000 for testing. The same set of real images is used for testing across all subsets.

    \item \textbf{UniversalFakeDetect} dataset, introduced by Wang {\it et al.}~\cite{ojha2023towards}, incorporates real images from LSUN, ImageNet, and Laion, and uses ProGAN-generated images as the training set, comprising 20 subsets of generated images. For training, we adopt a 4-class setting (horse, chair, cat, car), as outlined in~\cite{tan2024rethinking,liu2024forgery,tan2025c2p}. The test set consists of 19 subsets generated by various generative models, including ProGAN~\cite{karras2017progressive}, StyleGAN~\cite{karras2019style}, BigGAN~\cite{brock2018large}, CycleGAN~\cite{zhu2017unpaired}, StarGAN~\cite{choi2018stargan}, GauGAN \cite{park2019gaugan}], Deepfake~\cite{rossler2019faceforensics}, CRN \cite{chen2017photographic}, IMLE~\cite{li2019diverse}, SAN~\cite{dai2019second}, SITD~\cite{chen2018learning}, Guided Diffusion~\cite{dhariwal2021diffusion}, \acs{ldm}~\cite{rombach2022high}, GLIDE~\cite{nichol2021glide}, and DALL-E~\cite{ramesh2021zero}.

\end{enumerate}

\noindent \textbf{Evaluation Protocol.} We follow the standard train/test splits as defined in the original datasets. Input images are resized to $224\times224$ pixels and normalized. Model performance is evaluated using accuracy (ACC), measuring the proportion of correct predictions. Higher ACC indicates better performance.
% \noindent {\textbf{Evaluation metrics.}} Following the convention of previous detection methods, we report the accuracy (ACC) in our study. Accuracy (ACC) measures the proportion of correct predictions, with higher scores indicating better performance. 

\noindent {\\ \textbf{Baselines.}} In this study, we fine-tuned several established models for binary classification. Specifically, we modified ResNet~\cite{he2016deep}, Xception~\cite{chollet2017xception}, ViT~\cite{dosovitskiy2020image}, and  DeiT~\cite{touvron2021training} by replacing their final \ac{fc} layers with a novel binary classification layer, initializing the model layers with pre-trained weights.  We trained from scratch Vim~\cite{zhu2024vision}, VSSD~\cite{shi2025vssd},  MambaVision~\cite{hatamizadeh2025mambavision}. Regarding AntiFakePrompt~\cite{chang2023antifakeprompt} and  Bi-LORA~\cite{keita2025bi,keita2024harnessing}, we tuned them on different datasets. For the remaining models, including Co-occurence~\cite{nataraj2019detecting}, Freq-spec~\cite{zhang2019detecting}, CNN-Spot~\cite{wang2020cnn}, FatchFor~\cite{chai2020makes}, UniFD~\cite{ojha2023towards}, LGrad~\cite{tan2023learning}, F3Net~\cite{qian2020thinking}, FreqNet~\cite{tan2024frequency}, NPR~\cite{tan2024rethinking}, Fatformer~\cite{liu2024forgery}, C2P-CLIP~\cite{tan2025c2p}, we utilized the results reported in the C2P-CLIP paper. \\\\

\noindent {\textbf{Implementation details.}} In our experiments, we leveraged the PyTorch deep learning framework on a Linux computer equipped with a 16 GB NVIDIA RTX A4500 GPU. Our study used baseline models obtained from their publicly available repositories, which we fine-tuned to align with our specific experimental setup. For the train/test split of the datasets, we followed the splitting protocols proposed in their original papers, as shown in Table~\ref{tab:antifakepromt}, and~\ref{tab:bedroom}.

\begin{table*}[]
\caption{AntifakePrompt Dataset Detail}
\label{tab:antifakepromt}
\begin{adjustbox}{width=\linewidth}
\begin{tabular}{@{}c|cccccccccccccccc|c@{}}
\toprule
\diagbox{Split}{Models} & MS COCO & Flickr & SD2 & SDXL & IF & DALLE-2 & SGXL & ControlNet & LaMa & SD2IP & LTE & SD2SR & \begin{tabular}[c]{@{}c@{}}Deeper-\\ Forensics\end{tabular} & Adver & Backdoor & \begin{tabular}[c]{@{}c@{}}Data\\ Poisoning\end{tabular} & \#Total \\ \midrule
\#Train & 116 870 & NA & 30 000 & NA & NA & NA & NA & NA & 30 000 & 30 000 & NA & NA & NA & NA & NA & NA & 206 870 \\
\#Val & 3 000 & NA & 1 500 & NA & NA & NA & NA & NA & 1 500 & 1 500 & NA & NA & NA & NA & NA & NA & 6 000 \\
\#Test & 3 000 & 3 000 & 3 000 & 3 000 & 3 000 & 3 000 & 3 000 & 3 000 & 3 000 & 3 000 & 3 000 & 3 000 & 3 000 & 3 000 & 3 000 & 3 000 & 48 000 \\ \bottomrule
\#Total & 122 870 & 3 000 & 34 500 & 3 000 & 3 000 & 3 000 & 3 000 & 3 000 & 34 500 & 34 500 & 3 000 & 3 000 & 3 000 & 3 000 & 3 000 & 3 000 & \bf260 870\\ \bottomrule
\end{tabular}
\end{adjustbox}
\end{table*}

\begin{table*}[]
\caption{Bedroom Dataset Detail}
\label{tab:bedroom}
\begin{adjustbox}{width=\linewidth}
\begin{tabular}{@{}c|ccccccccccccc|c@{}}
\toprule
\diagbox{Split}{Models} & LSUN Bed & LDM & ADM & DDPM & iDDPM & PNDM & SDv1.4 & Glide & ProGAN & StyleGAN & ProjectedGAN & \begin{tabular}[c]{@{}c@{}}Diff-\\ StyleGAN2\end{tabular}  & \begin{tabular}[c]{@{}c@{}}Diff-\\ ProjectedGAN\end{tabular} & \#Total \\ \midrule
\#Train & 400 000 & 40 000 & 40 000 & 40 000 & 40 000 & 40 000 & NA & NA & 40 000 & 40 000 & 40 000 & 40 000 & 40 000 &  800 000 \\
\#Val & 10 000 & 1 000 & 1 000 & 1 000 & 1 000 & 1 000 & NA & NA & 1 000 & 1 000 & 1 000 & 1 000 & 1 000 &  20 000 \\
\#Test & 10 000 & 10 000 & 10 000 & 10 000 & 10 000 & 10 000 & 10 000 & 10 000 & 10 000 & 10 000 & 10 000 & 10 000 & 10 000 &  130 000 \\ \bottomrule
\#Total & 420 000 & 51 000 & 51 000 & 51 000 & 51 000 & 51 000 & 10 000 & 10 000 & 51 000 & 51 000 & 51 000 & 51 000 & 51 000 & \bf 950 000 \\ \bottomrule
\end{tabular}
\end{adjustbox}
\end{table*}

\begin{figure*}
    \centering
    \begin{subfigure}[b]{0.21\textwidth}
        \includegraphics[width=.8\linewidth]{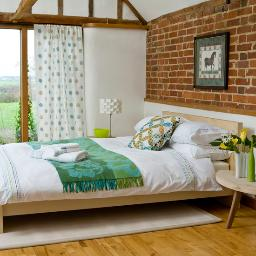}
        \caption{Real - $0|0|0|0$ }
    \end{subfigure}
%    \hspace{.1in}
 %   \vspace{.1in}
    \begin{subfigure}[b]{0.21\textwidth}
        \includegraphics[width=.8\linewidth]{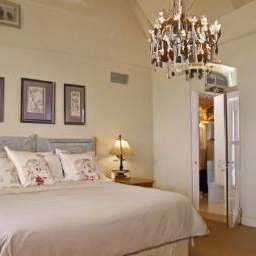}
        \caption{Fake (ADM) - $0|0|0|1$ }
    \end{subfigure}
   % \hspace{.1in}
   % \vspace{.1in}
    \begin{subfigure}[b]{0.21\textwidth}
        \includegraphics[width=.8\linewidth]{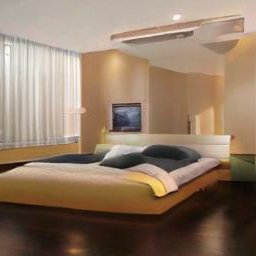}
        \caption{Fake (LDM) - $1|1|1|1$ }
    \end{subfigure}
   % \hspace{.1in}
   % \vspace{.1in}
    \begin{subfigure}[b]{0.21\textwidth}
        \includegraphics[width=.8\linewidth]{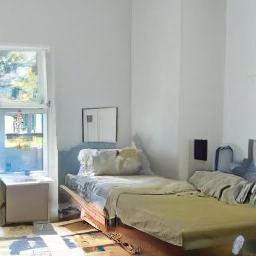}
        \caption{Fake (DDPM) - $0|0|0|1$ }
    \end{subfigure}
    \begin{subfigure}[b]{0.21\textwidth}
\includegraphics[width=.8\linewidth]{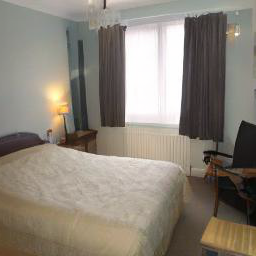}
        \caption{Fake (IDDPM) - $1|0|0|1$}
    \end{subfigure}
  %  \hspace{.1in}
  %  \vspace{.1in}
    \begin{subfigure}[b]{0.21\textwidth}
        \includegraphics[width=0.8\linewidth]{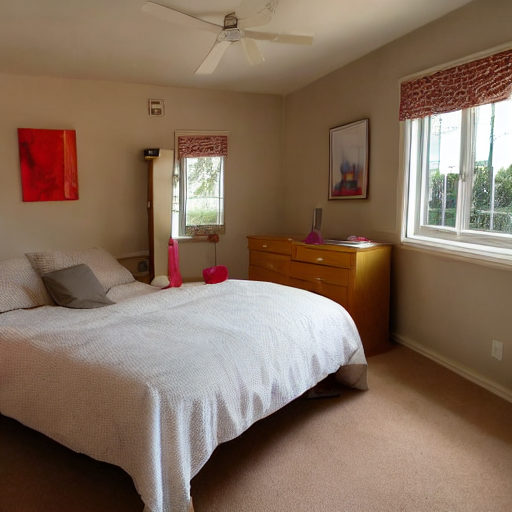}
        \caption{Fake (SD) - $0|1|0|1$}
    \end{subfigure}
    % \hspace{.1in}
    % \vspace{.1in}
    \begin{subfigure}[b]{0.21\textwidth}     
        \includegraphics[width=.8\linewidth]{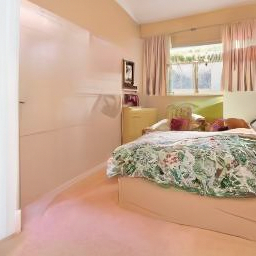}
        \caption{Fake (PNDM) - $0|0|0|1$}
    \end{subfigure}
%    \hspace{.1in}
%    \vspace{.1in}
     \begin{subfigure}[b]{0.21\textwidth}     
        \includegraphics[width=.8\linewidth]{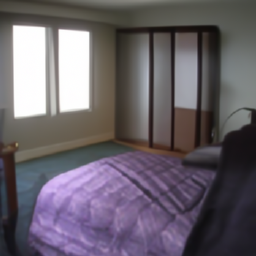}
         \caption{Fake (GLIDE) - $0|0|1|1$}
     \end{subfigure}  
    \caption{Each sub-figure is a random image from a testing set, labeled below. The 4-digit binary code shows results from ResNet, Xception, \acs{deit}, and \acs{blip}2 models, where '0' means real and '1' means fake. All generated images are considered fake.}
    \label{fig:Visual} \vspace{-4mm}
\end{figure*}

\begin{table*}[!ht]
\caption{Performance comparison on AntifakePrompt Dataset, \colorbox[HTML]{F2F2F2}{SSMs}, \colorbox[HTML]{E6F7FF}{CNNs}, \colorbox[HTML]{EAFFEA}{Attentions}, \colorbox[HTML]{F5F5DC}{VLMs}}
\label{tab:TrainedOnAntifakePrompt}
\begin{adjustbox}{width=\linewidth}

\begin{tabular}{@{}ccccccccc@{}}
\toprule
Methods & Training Set & \#params & MS COCO & Flickr & SD2 & SDXL & IF & DALLE-2 \\ \midrule
\rowcolor[HTML]{F2F2F2} Vim~\cite{zhu2024vision} & MS COCO vs. SD2+LaMa & 6.96M & 97.90 & 98.93 & 99.77 & 15.93 & 33.87 & 22.40 \\
\rowcolor[HTML]{F2F2F2} VSSD~\cite{shi2025vssd} & MS COCO vs. SD2+LaMa & 23.76M & 100.00 & 100.00 & 99.97 & 00.23 & 14.00 & 28.13 \\
\rowcolor[HTML]{F2F2F2} MambaVision~\cite{hatamizadeh2025mambavision} & MS COCO vs. SD2+LaMa & 31.15M & 99.43 & 99.80 & 99.83 & 03.50 & 04.93 & 13.50 \\
\rowcolor[HTML]{E6F7FF} ResNet~\cite{he2016deep} & MS COCO vs. SD2+LaMa & 23.51M & 99.77 & 99.73  & 99.93  & 05.03  & 20.67  & 02.03  \\
\rowcolor[HTML]{E6F7FF} Xception~\cite{chollet2017xception} & MS COCO vs. SD2+LaMa & 20.81M & 99.67  & 99.53  & 98.97  & 40.03 & 48.93  & 11.67  \\
\rowcolor[HTML]{EAFFEA} ViT~\cite{dosovitskiy2020image} & MS COCO vs. SD2+LaMa & 85.80M & 96.37  & 96.57  & 100.00  & 08.37  & 21.37  & 15.50  \\
\rowcolor[HTML]{EAFFEA} DeiT~\cite{touvron2021training} & MS COCO vs. SD2+LaMa & 5.52M & 99.20 & 99.40  & 99.60  & 02.57 & 12.77 & 09.63  \\
\rowcolor[HTML]{F5F5DC} AntifakePrompt~\cite{chang2023antifakeprompt} & MS COCO vs. SD2+LaMa & 7.91B & 91.83  & 84.53  & 98.57  & 99.17  & 93.73  & 98.13  \\
\rowcolor[HTML]{F5F5DC} Bi-LORA~\cite{keita2025bi} & MS COCO vs. SD2+LaMa & 3.75B & 80.40  & 72.90  & 99.47  & \bf99.37 & \bf95.03  & \bf99.67 \\ 
\bottomrule

\toprule
\multirow{2}{*}{Methods} & \multirow{2}{*}{Training Set} & \multirow{2}{*}{\#params} & \multirow{2}{*}{SGXL} & \multirow{2}{*}{ControlNet} & \multicolumn{2}{c}{Inpainting} & \multicolumn{2}{c}{Super Res.} \\ \cmidrule(l){6-9} 
 &  &  &  &  & LaMa & SD2 & LTE & SD2 \\ \cmidrule(r){1-5}
\rowcolor[HTML]{F2F2F2} Vim~\cite{zhu2024vision} & MS COCO vs. SD2+LaMa & 6.96M & 10.00 & 18.67 & 93.60 & 79.07 & 88.77 & 32.40 \\
\rowcolor[HTML]{F2F2F2} VSSD~\cite{shi2025vssd} & MS COCO vs. SD2+LaMa & 23.76M & 10.70 & 3.63 & 99.60 & 95.27 & 100.00 &  03.43\\
\rowcolor[HTML]{F2F2F2} MambaVision~\cite{hatamizadeh2025mambavision} & MS COCO vs. SD2+LaMa & 31.15M & 06.27 & 04.47 & 98.60 & 85.87 & 99.13 & 02.37\\

\rowcolor[HTML]{E6F7FF} ResNet~\cite{he2016deep} & MS COCO vs. SD2+LaMa & 23.51M & 01.57  & 04.40  & 99.90  & 99.73  & 95.67  & 99.87 \\
\rowcolor[HTML]{E6F7FF} Xception~\cite{chollet2017xception} & MS COCO vs. SD2+LaMa & 20.81M & 03.63  & 21.90   & 99.80  & 99.63  & 77.60  & 100.00  \\
\rowcolor[HTML]{EAFFEA} ViT~\cite{dosovitskiy2020image} & MS COCO vs. SD2+LaMa & 85.80M  & 04.33  & 13.20 & 99.77  & 99.67  & 99.57  & 99.90  \\
\rowcolor[HTML]{EAFFEA} DeiT~\cite{touvron2021training} & MS COCO vs. SD2+LaMa & 5.52M & 02.13  & 02.67 & 100.00  & 99.87  & 99.43 & 99.97 \\ 
\rowcolor[HTML]{F5F5DC} AntifakePrompt~\cite{chang2023antifakeprompt} & MS COCO vs. SD2+LaMa & 7.91B & \bf 99.73 & 94.77  & 59.27  & 89.00 & \bf100.00  &  99.93  \\ 
\rowcolor[HTML]{F5F5DC} Bi-LORA~\cite{keita2025bi} & MS COCO vs. SD2+LaMa & 3.75B & 98.57  & \bf98.90  & 80.17  & \bf94.20  & 99.67  &  \bf99.97 \\ 
\bottomrule

\toprule
\multirow{2}{*}{Methods} & \multirow{2}{*}{Training Set} & \multirow{2}{*}{\#params} & \multirow{2}{*}{\begin{tabular}[c]{@{}c@{}}Deeper-\\ Forensics\end{tabular}} & \multirow{2}{*}{Adver.} & \multicolumn{2}{c}{Attack} & \multicolumn{2}{c}{\multirow{2}{*}{Average}} \\ \cmidrule(lr){6-7}
 &  &  &  &  & Backdoor & Data Poisoning & \multicolumn{2}{c}{} \\ \cmidrule(r){1-5} \cmidrule(l){8-9} 
\rowcolor[HTML]{F2F2F2} Vim~\cite{zhu2024vision} & MS COCO vs. SD2+LaMa & 6.96M & 77.90 & 00.47 & 16.20 & 08.10 & \multicolumn{2}{c}{49.62} \\
\rowcolor[HTML]{F2F2F2} VSSD~\cite{shi2025vssd} & MS COCO vs. SD2+LaMa & 23.76M & 61.77 & 00.00 & 01.80 & 01.47 & \multicolumn{2}{c}{45.00} \\
\rowcolor[HTML]{F2F2F2} MambaVision~\cite{hatamizadeh2025mambavision} & MS COCO vs. SD2+LaMa & 31.15M & \bf93.53 & 01.67 &  07.00 & 03.03 & \multicolumn{2}{c}{45.18} \\
\rowcolor[HTML]{E6F7FF} ResNet~\cite{he2016deep} & MS COCO vs. SD2+LaMa & 23.51M & 01.63  & 98.13  & 98.53  & 95.63  & \multicolumn{2}{c}{63.89 } \\
\rowcolor[HTML]{E6F7FF} Xception~\cite{chollet2017xception} & MS COCO vs. SD2+LaMa & 20.81M & 79.50  & 92.80  & 99.07  & 94.73  & \multicolumn{2}{c}{72.97 } \\
\rowcolor[HTML]{EAFFEA} ViT~\cite{dosovitskiy2020image} & MS COCO vs. SD2+LaMa & 85.80M  & 97.10  & 100.00  & 99.80  & 99.90  & \multicolumn{2}{c}{71.96 } \\
\rowcolor[HTML]{EAFFEA} DeiT~\cite{touvron2021training} & MS COCO vs. SD2+LaMa & 5.52M & 03.07 & 100.000  & 100.00  & 99.90 & \multicolumn{2}{c}{64.39 } \\
\rowcolor[HTML]{F5F5DC} AntifakePrompt~\cite{chang2023antifakeprompt} & MS COCO vs. SD2+LaMa & 7.91B & \bf95.30  & \bf88.67  &  \bf91.60  & \bf83.57  & \multicolumn{2}{c}{\bf91.74 } \\
\rowcolor[HTML]{F5F5DC} Bi-LORA~\cite{keita2025bi} & MS COCO vs. SD2+LaMa & 3.75B & 87.60  & 88.50  & 89.73  & 74.30  & \multicolumn{2}{c}{91.15 } \\ 
\bottomrule
\end{tabular}

\end{adjustbox}
\end{table*}

\begin{figure*}[!ht]
    \centering
    \begin{subfigure}[b]{0.75\textwidth}
        \includegraphics[width=\linewidth]{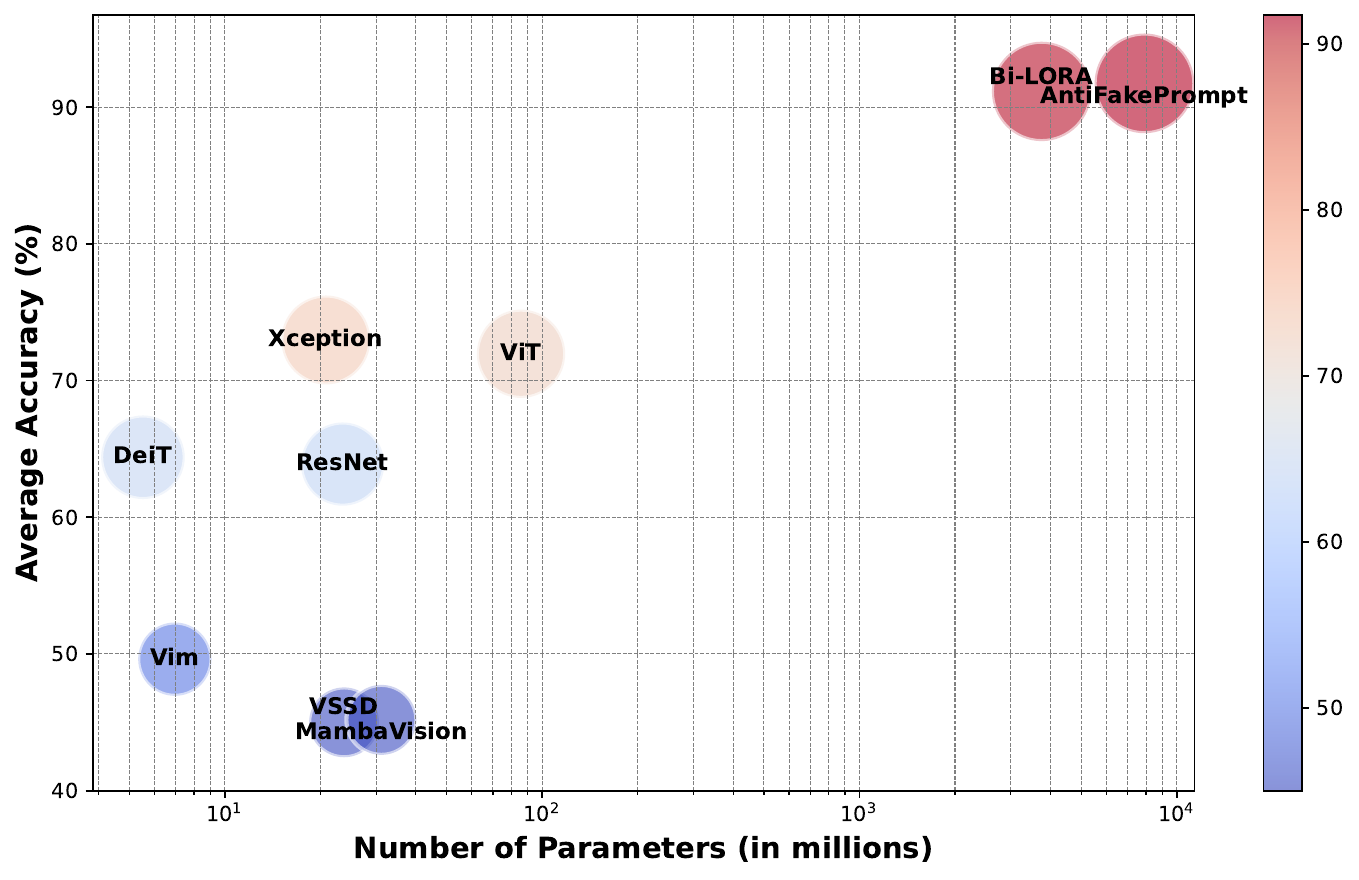}
        \caption{AntifakePrompt Dataset}
    \end{subfigure}
    \begin{subfigure}[b]{0.75\textwidth}
        \includegraphics[width=\linewidth]{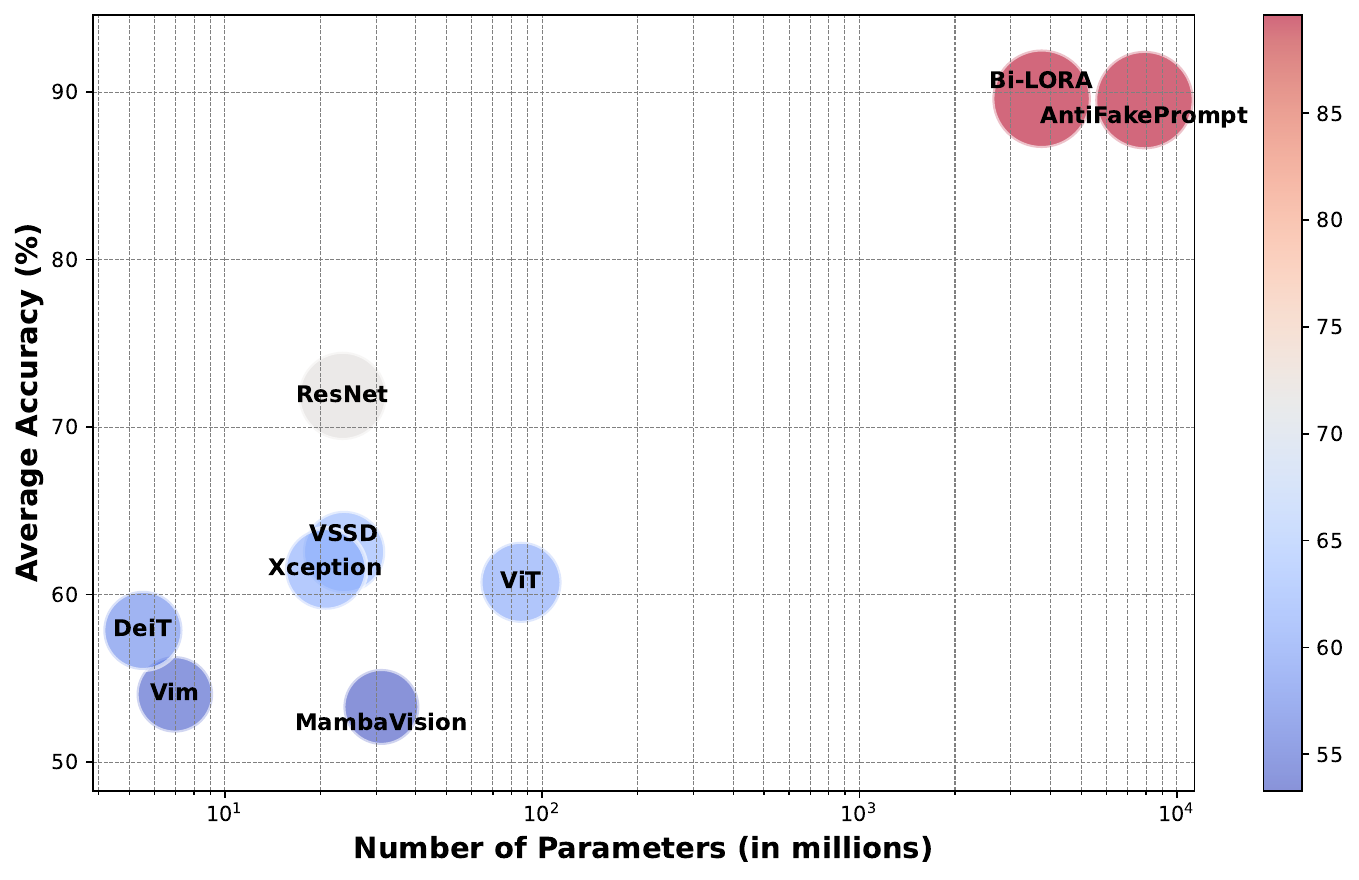}
        \caption{Bedroom Dataset}
    \end{subfigure}
    \caption{Performance comparison of multiple models on AntifakePrompt and Bedroom datasets. The x-axis shows model size (parameters), and the y-axis represents accuracy. Model families like SSMs, CNNs, attention-based models, and VLMs are compared to evaluate efficiency and effectiveness trade-offs.}
    \label{fig:modelPerformanceBedroomAntifakePrompt}
\end{figure*}

\begin{table*}[]
\caption{Accuracy performance comparison  lsun bedroom dataset, \colorbox[HTML]{F2F2F2}{SSMs}, \colorbox[HTML]{E6F7FF}{CNNs}, \colorbox[HTML]{EAFFEA}{Attentions}, \colorbox[HTML]{F5F5DC}{VLMs}.}
\label{tab:TrainedOnBedroom}
\begin{adjustbox}{width=0.7\linewidth}

\begin{tabular}{@{}cccccccc@{}}
\toprule
Model   & Training Set & \#params & REAL          & LDM           & ADM           & DDPM          & IDDPM         \\ \midrule
\rowcolor[HTML]{F2F2F2} Vim~\cite{zhu2024vision}  & LSun Bed vs. LDM          & 6.96M           &   99.97   &  96.93   & 00.93   &  02.02  & 01.89   \\ 

\rowcolor[HTML]{F2F2F2} VSSD~\cite{shi2025vssd} & LSun Bed vs. LDM          & 23.76M &    99.99   &  100.00   &   01.45  &  02.99   &  06.39\\ 

\rowcolor[HTML]{F2F2F2} MambaVision~\cite{hatamizadeh2025mambavision} & LSun Bed vs. LDM & 31.15M &  100.00    &  99.76  & 03.98   &  00.56   & 14.00   \\ 
\rowcolor[HTML]{E6F7FF} ResNet~\cite{he2016deep} & LSun Bed vs. LDM          & 23.51M           &  99.98   &  \bf100.00   &  63.65  &  00.85  & 82.74   \\ 
\rowcolor[HTML]{E6F7FF} Xception~\cite{chollet2017xception} & LSun Bed vs. LDM          & 20.81M          & 99.90  &  \bf100.00   & 04.87   & 08.60  & 10.30   \\
\rowcolor[HTML]{EAFFEA} ViT~\cite{dosovitskiy2020image} & LSun Bed vs. LDM          & 85.80M  & \bf100.00  & 99.79  & 04.66  & 13.08  & 13.98   \\ 
\rowcolor[HTML]{EAFFEA} DeiT~\cite{touvron2021training} & LSun Bed vs. LDM          & 5.52M          &  99.99 & 97.80 & 04.54  & 05.65  & 09.30  \\ 

\rowcolor[HTML]{F5F5DC} AntifakePrompt~\cite{chang2023antifakeprompt} & LSun Bed vs. LDM & 7.91B & 94.65  & 99.55 & \bf76.59 & \bf99.05  & \bf95.89  \\ 
\rowcolor[HTML]{F5F5DC} Bi-LORA~\cite{keita2025bi} & LSun Bed vs. LDM          & 3.75B          & 98.55 & 99.68 &  71.98 & 98.47 & 95.51 \\ 
\bottomrule

\toprule

Model   & Training Set & \#params & PNDM          & SD v1.4       & GLIDE         & \multicolumn{2}{c}{Average} \\ \midrule
\rowcolor[HTML]{F2F2F2} Vim~\cite{zhu2024vision}  & LSun Bed vs. LDM          & 6.96M           & 36.05  & 98.86  &  95.64   & \multicolumn{2}{c}{ 54.04  }  \\
\rowcolor[HTML]{F2F2F2} VSSD~\cite{shi2025vssd} & LSun Bed vs. LDM          & 23.76M           &   92.29  &  99.99   &  97.20  & \multicolumn{2}{c}{  62.54  }  \\
\rowcolor[HTML]{F2F2F2} MambaVision~\cite{hatamizadeh2025mambavision} & LSun Bed vs. LDM & 31.15M  & 42.17  & 96.50  & 69.35  & \multicolumn{2}{c}{ 53.29 }  \\
\rowcolor[HTML]{E6F7FF} ResNet~\cite{he2016deep} & LSun Bed vs. LDM          & 23.51M           & 28.89  & \bf99.97  & 98.84  & \multicolumn{2}{c}{ 71.86 }  \\
\rowcolor[HTML]{E6F7FF} Xception~\cite{chollet2017xception} & LSun Bed vs. LDM          & 20.81M           & 74.06  & 97.74  & 96.78  & \multicolumn{2}{c}{ 61.53 }    \\
\rowcolor[HTML]{EAFFEA} ViT~\cite{dosovitskiy2020image} & LSun Bed vs. LDM          & 85.80M           & 55.27 & 99.73 & \bf99.29 & \multicolumn{2}{c}{60.73}    \\
\rowcolor[HTML]{EAFFEA} DeiT~\cite{touvron2021training} & LSun Bed vs. LDM          & 5.52M           & 48.27  & 99.04  & 98.29  & \multicolumn{2}{c}{ 57.86 }    \\
\rowcolor[HTML]{F5F5DC} AntifakePrompt~\cite{chang2023antifakeprompt} & LSun Bed vs. LDM          & 7.91B & \bf99.93  & 53.83  &  96.66  & \multicolumn{2}{c}{ 89.52 }     \\
\rowcolor[HTML]{F5F5DC} Bi-LORA~\cite{keita2025bi} & LSun Bed vs. LDM          & 3.75B           & 99.88 & 56.98  & 95.71  & \multicolumn{2}{c}{ \bf89.60 }     \\ \bottomrule

\end{tabular}

\end{adjustbox}
\end{table*}

\begin{table*}[]
\caption{Cross-model Accuracy (Acc) Performance on the UniversalFakeDetect Dataset.}
\label{tab:TrainedOnUniversalFakeDetect}

\begin{adjustbox}{width=\linewidth}
\begin{tabular}{@{}lccccccccccccccccccccc@{}}
\toprule
\multicolumn{1}{c}{\multirow{2}{*}{Methods}} & \multirow{2}{*}{Ref} & \multicolumn{6}{c}{GAN} & \multirow{2}{*}{\begin{tabular}[c]{@{}c@{}}Deep\\ Fakes\end{tabular}} & \multicolumn{2}{c}{Low level} & \multicolumn{2}{c}{Perceptual loss} & \multirow{2}{*}{Guided} & \multicolumn{3}{c}{LDM} & \multicolumn{3}{c}{Glide} & \multirow{2}{*}{Dalle} & \multirow{2}{*}{mAcc} \\ \cmidrule(lr){3-8} \cmidrule(lr){10-13} \cmidrule(lr){15-20}
\multicolumn{1}{c}{} &  & \begin{tabular}[c]{@{}c@{}}Pro-\\ GAN\end{tabular} & \begin{tabular}[c]{@{}c@{}}Cycle-\\ GAN\end{tabular} & \begin{tabular}[c]{@{}c@{}}Big-\\ GAN\end{tabular} & \begin{tabular}[c]{@{}c@{}}Style-\\ GAN\end{tabular} & \begin{tabular}[c]{@{}c@{}}Gau-\\ GAN\end{tabular} & \begin{tabular}[c]{@{}c@{}}Star-\\ GAN\end{tabular} &  & SITD & SAN & CRN & IMLE &  & \begin{tabular}[c]{@{}c@{}}200\\ steps\end{tabular} & \begin{tabular}[c]{@{}c@{}}200\\ w/cfg\end{tabular} & \begin{tabular}[c]{@{}c@{}}100\\ steps\end{tabular} & \begin{tabular}[c]{@{}c@{}}100\\ 27\end{tabular} & \begin{tabular}[c]{@{}c@{}}50\\ 27\end{tabular} & \begin{tabular}[c]{@{}c@{}}100\\ 10\end{tabular} &  &  \\ \cmidrule(r){1-2} \cmidrule(lr){9-9} \cmidrule(lr){14-14} \cmidrule(l){21-22} 
CNN-Spot & CVPR2020 & 99.99 & 85.20 & 70.20 & 85.70 & 78.95 & 91.70 & 53.47 & 66.67 & 48.69 & 86.31 & 86.26 & 60.07 & 54.03 & 54.96 & 54.14 & 60.78 & 63.80 & 65.66 & 55.58 & 69.58 \\
Patchfor & ECCV2020 & 75.03 & 68.97 & 68.47 & 79.16 & 64.23 & 63.94 & 75.54 & 75.14 & 75.28 & 72.33 & 55.30 & 67.41 & 76.50 & 76.10 & 75.77 & 74.81 & 73.28 & 68.52 & 67.91 & 71.24 \\
Co-occurence & Elect. Imag. & 97.70 & 97.70 & 53.75 & 92.50 & 51.10 & 54.70 & 57.10 & 63.06 & 55.85 & 65.65 & 65.80 & 60.50 & 70.70 & 70.55 & 71.00 & 70.25 & 69.60 &  69.90 & 67.55 & 66.86 \\
Freq-spec & WIFS2019 & 49.90 & \bf99.90 & 50.50 & 49.90 & 50.30 & 99.70 & 50.10 & 50.00 & 48.00 & 50.60 & 50.10 & 50.90 & 50.40 & 50.40 & 50.30 & 51.70 & 51.40 & 50.40 & 50.00 & 55.45 \\
F3Net & ECCV2020 & 99.38 & 76.38 & 65.33 & 92.56 & 58.10 & \bf100.00 & 63.48 & 54.17 & 47.26 & 51.47  & 51.47  & 69.20 &  68.15 & 75.35 & 68.80  &  81.65 & 83.25 & 83.05  & 66.30 & 71.33  \\
UniFD & CVPR2023 & \bf100.00 & 98.50 & 94.50 & 82.00 & \bf99.50 & 97.00 & 66.60 & 63.00 & 57.50 & 59.50 & 72.00 & 70.03 & 94.19 & 73.76 & 94.36 & 79.07 & 79.85 & 78.14 & 86.78 & 81.38 \\
LGrad & CVPR2023 & 99.84 & 85.39 & 82.88 & 94.83 & 72.45 & 99.62 & 58.00 & 62.50 & 50.00 & 50.74 & 50.78 & 77.50 & 94.20 & 95.85 & 94.80 & 87.40 & 90.70 & 89.55 & 88.35 & 80.28 \\

\rowcolor[HTML]{F5F5DC} Bi-LORA & ICASSP2023 & 98.71 & 96.74 & 81.18 & 78.30 & 96.30 & 86.32 & 57.78 & 68.89 & 52.28 & 73.00 & 82.60 & 65.10 & 85.15 & 59.20 & 85.00 & 83.50 & 85.65 & 84.90 & 72.70 & 78.59 \\
\rowcolor[HTML]{F5F5DC} AntiFakePrompt & CVPR2023 & 99.26 & 96.82 & 87.88 & 80.00 & 98.13 & 83.57 & 60.20 & 70.56 & 53.70 & 79.21 & 79.01 & 73.75 & 89.55 & 64.10 & 89.80 & 93.55 & 93.90 & 92.95 & 80.10 & 82.42 \\
FreqNet & AAAI2024 & 97.90 & 95.84 & 90.45 & 97.55 & 90.24 & 93.41 & 97.40 & 88.92 & 59.04 & 71.92 & 67.35 & \bf86.70 & 84.55 & \bf99.58 & 65.56 & 85.69 & \bf97.40 & 88.15 & 59.06 & 85.09 \\
NPR & CVPR2024 & 99.84 & 95.00 & 87.55 & 96.23 & 86.57 & 99.75 & 76.89 & 66.94 & \bf98.63 & 50.00 & 50.00 & 84.55 & 97.65 & 98.00 & 98.20 & \bf96.25 & 97.15 & \bf97.35 & 87.15 & 87.56 \\
FatFormer & CVPR2024 & 99.89 & 99.32 & \bf99.50 & 97.15 & 99.41 & 99.75 & 93.23 & 81.11 & 68.04 & 69.45 & 69.45 & 76.00 & 98.60 & 94.90 & 98.65 & 94.35 & 94.65 & 94.20 & 98.75 & 90.86 \\
C2P-CLIP & AAAI2025 & 99.71 & 90.69 & 95.28 & \bf99.38 & 95.26 & 96.60 & 89.86 & \bf98.33 & 64.61 & 90.69 & 90.69 & 77.80 & 99.05 & 98.05 & 98.95 & 94.65 & 94.20 & 94.40 & \bf98.80 & 93.00 \\
C2P-CLIP & AAAI2025 & 99.98 & 97.31 & 99.12 & 96.44 & 99.17 & 99.60 & \bf93.77 & 95.56 & 64.38 & \bf93.29 & 93.29 & 69.10 & \bf99.25 & 97.25 & \bf99.30 & 95.25 & 95.25 & 96.10 & 98.55 & \bf93.79 \\ 

\rowcolor[HTML]{F2F2F2} Vim & - & 98.61 & 78.43 & 73.60 & 88.33 & 72.44 & 88.79 & 88.94 & 63.89 & 50.23 & 58.14 & 58.46 & 74.10 & 66.55 & 68.80 & 66.90 & 70.90 & 75.55 & 71.75 & 74.10 & 73.08 \\
\rowcolor[HTML]{F2F2F2} VSSD & ICCV2025 & 100.00 & 74.90 & 73.10 & 86.70 & 63.30 & 100.00 & 78.80 & 59.70 & 51.10 & 52.30 & 52.30 & 90.50 & 89.00 & 88.90 & 89.00 & 64.80 & 67.20 & 65.70 & 86.80 & 75.48 \\

\rowcolor[HTML]{F2F2F2} MambaVision & CVPR2025 & 99.62 & 75.74 & 73.28 & 87.01 & 63.38 & 100.00 & 79.02 & 58.61 & 50.91 & 54.27 & 54.30 & 83.35 & 67.25 & 66.95 & 68.60 & 59.55 & 61.45 & 60.70 & 69.55 & 70.19 \\

% & - &  &  &  &  &  &  &  &  &  &  &  &  &   &  &  &  &  &  &  &  \\
\bottomrule
\end{tabular}

\end{adjustbox}
\end{table*}

\begin{figure}[ht]
    \center
    \includegraphics[width=0.9\linewidth]{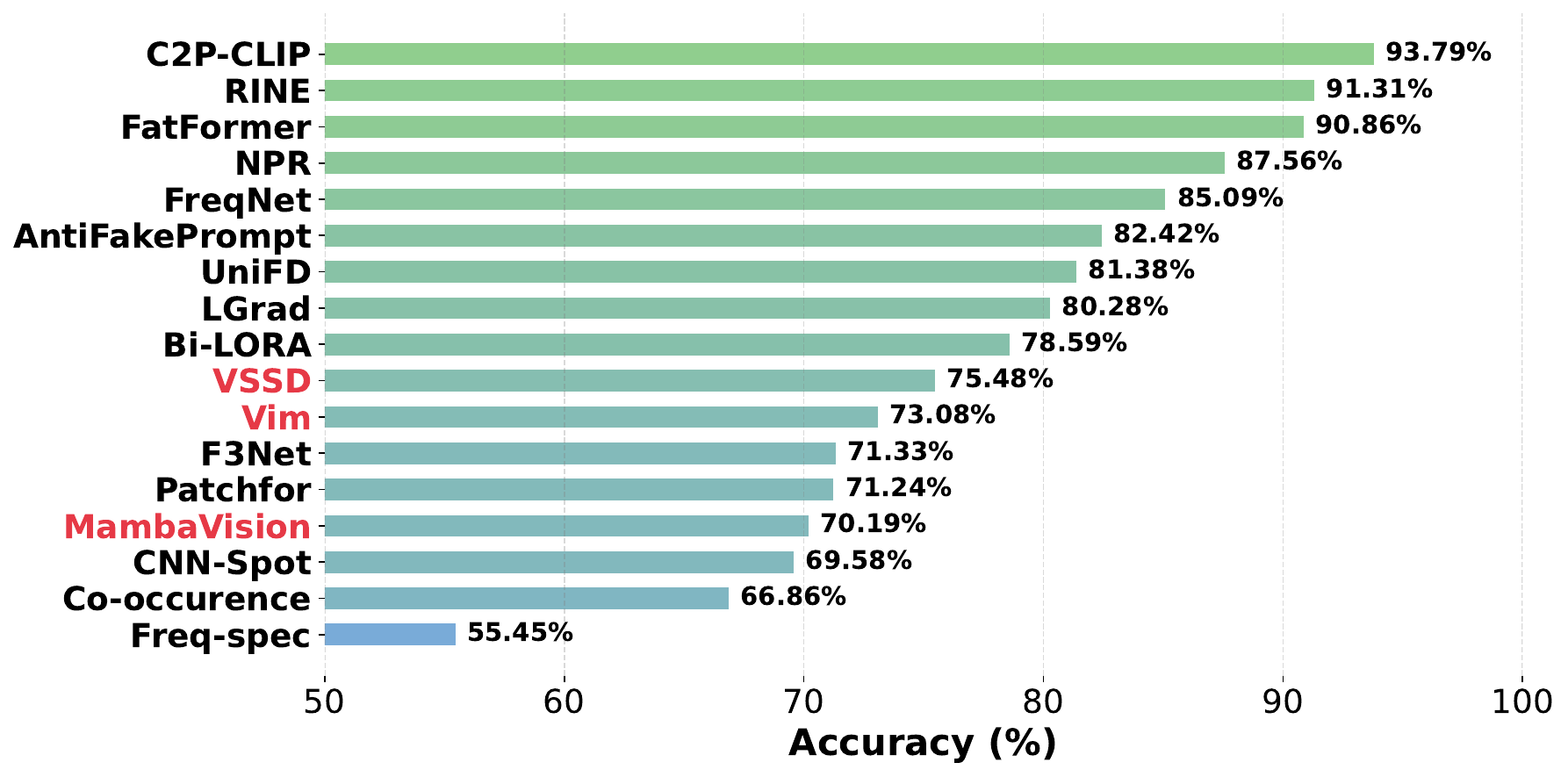}
    \caption{Performance comparison of models on the UniversalFakeDetect dataset, including frequency-based (Freq-spec, Co-occurrence), convolutional (CNN-Spot, F3Net), transformer-based (FatFormer), SSM-based (Vim, VSSD), hybrid (MambaVision, Bi-LORA, LGrad), and multimodal (AntifakePrompt, Bi-LORA, C2P-CLIP) architectures.}
\label{fig:modelPerformanceUniversalFakeDetect}
\end{figure}

\subsection{Discussion and Analysis}
This section investigates the efficacy of Vision Mamba models for detecting \ac{ai}-generated images. We benchmark Vision Mamba against various established architectures, encompassing \acp{cnn}, attention-based models (specifically, Transformers), and \acp{vlm}. This evaluation is conducted across three datasets: AntifakePrompt, LSUN Bedroom, and UniversalFakeDetect. It is important to note the structural differences between these datasets. While AntifakePrompt and LSUN Bedroom are organized so that each subset contains real or fake samples only, the UniversalFakeDetect dataset is structured differently, with each subset containing a mixture of real and fake samples. This distinction is essential for understanding the evaluation configuration and interpreting the results. Our analysis aims to elucidate the strengths and limitations of the Vision Mamba architecture, identify potential avenues for improvement, and contextualize its performance within the broader field of \ac{ai}-generated image detection.

\noindent Table~\ref{tab:TrainedOnAntifakePrompt} compares detector performance on the AntifakePrompt dataset after initial training on the "MS COCO vs. SD2+LaMa" training set. The results reveal significant performance variations based on model architecture and parameter count. AntifakePrompt and Bi-LORA, the largest models with billions of parameters, exhibit superior generalization across most subsets, achieving high accuracy on LTE, SGXL, DeeperForensics, and ControlNet. AntifakePrompt is particularly effective on SGXL, LTE, and attack scenario subsets, demonstrating high accuracy and resilience even under adversarial conditions. In contrast, while effective on certain subsets, SSM-based models like Vim, VSSD, and MambaVision struggle with generalization, particularly on SDXL, SGXL, DALLE-2, and Adver. Attention-based models, such as ViT and DeiT, deliver competitive results across most subsets and achieve outstanding performance on adversarial attack, backdoor, and data poisoning scenarios, highlighting the robustness of their architecture despite their relatively smaller size. CNN-based models, including ResNet and Xception, provide balanced performance across all datasets but show occasional weaknesses, notably on SGXL. In conclusion, larger models with sophisticated architectures, such as AntifakePrompt and Bi-LORA, outperform others in accuracy and robustness, whereas lightweight models exhibit limitations in their generalization capabilities.

\noindent Table~\ref{tab:TrainedOnBedroom} presents a comprehensive overview of detector performance on the LSUN Bedroom test set, evaluating various models, including text-to-image generators (LDM, SDv1.4, GLIDE) and unconditional diffusion models (ADM, DDPM, IDDPM, PNDM). The results reveal varying degrees of generalization ability across these model types. Notably, larger models with sophisticated architectures, such as AntifakePrompt and Bi-LORA, demonstrate superior generalization compared to smaller models like Vim, ViT, DeiT, ResNet, MambaVision, and VSSD. These larger models show strong robustness to text-to-image and unconditional diffusion-based generated images, maintaining high accuracy even in complex scenarios. SSM-based models (Vim, MambaVision, and VSSD) exhibit consistent detection accuracy on real images and the LDM subset but struggle to generalize to unconditional diffusion models. Vim, the smallest model, achieves near-perfect results on real images (99.97\%) and LDM (96.93\%) but performs poorly on ADM (00.93\%) and IDDPM (01.89\%). MambaVision, with a medium parameter count (31.15M), shows slightly improved generalization to the unseen PNDM subset (42.17\%). VSSD demonstrates strong performance on PNDM (92.29\%), surpassing both Vim and MambaVision, but struggles on ADM, DDPM, and IDDPM. \ac{cnn}-based models, ResNet and Xception, perform well on certain subsets like ADM, IDDPM, and PNDM but less effectively on others such as DDPM. Attention-based models ViT and DeiT leverage their enhanced feature extraction capabilities, with ViT achieving good results on several datasets, including PNDM (55.27\%) and GLIDE (99.29\%). Despite its smaller size, DeiT performs well on all subsets except ADM, DDPM, and IDDPM.

\noindent Figure~\ref{fig:modelPerformanceBedroomAntifakePrompt} illustrates the average performance of various models on the LSUN Bedroom (right subfigure) and AntifakePrompt (left subfigure) datasets, highlighting the trade-off between model size (complexity) and accuracy in AI-generated image detection. The x-axis represents the number of parameters (in millions), while the y-axis denotes the average accuracy (as a percentage). As shown in Tables~\ref{tab:TrainedOnAntifakePrompt} and \ref{tab:TrainedOnBedroom}, Vision Mamba models, despite their smaller parameter count, exhibit significant generalization challenges, resulting in poor performance on test subsets from unseen generative models. Conversely, the vision-language models, Bi-LORA and AntifakePrompt, with their larger parameter counts, demonstrate strong generalization across different test subsets from unseen generative models. Furthermore, CNN and attention-based models achieve moderate performance, with average accuracy generally at or above 60\%.

\noindent Figure~\ref{fig:modelPerformanceUniversalFakeDetect} presents a bar chart that illustrates the average performance of state-of-the-art methods on the UniversalFakeDetect benchmark, as detailed in Table~\ref{tab:TrainedOnUniversalFakeDetect}. This visualization provides a global overview of model capabilities in \ac{ai}-generated image detection, highlighting the strengths and limitations of different architectures. The results demonstrate an apparent performance disparity, with Transformer-based and \acp{vlm} significantly outperforming \ac{cnn} and Vision Mamba (SSM-based) models. Top-performing models like C2P-CLIP V1, C2P-CLIP V2, FatFormer, and NPR achieve accuracies approaching or exceeding 90\%, suggesting a strong ability to discern subtle differences between real and AI-generated images. This can be attributed to their architectural advantages: Transformer-based models capture long-range dependencies and global contextual information through attention mechanisms, which are crucial for detecting subtle, high-level patterns that distinguish real from synthetic content. Similarly, \acp{vlm} leverage their ability to integrate visual and textual information gained through large-scale multimodal pretraining to understand better and discern the subtle cues that differentiate authentic from \ac{ai}-generated images.

\noindent In contrast, Vision Mamba models exhibit considerably lower accuracy, around 70\%, reflecting the inherent limitations of state-space models in this domain. These models struggle to match the accuracy of Transformer and \ac{vlm} counterparts, indicating potential architectural deficiencies. While effective in long-sequence modeling tasks, state-space models demonstrate a reduced capacity to capture the spatial dependencies and hierarchical features necessary for effective \ac{ai}-generated image detection, which demands recognition of subtle and complex artifacts. This suggests that, despite their potential, significant architectural modifications or complementary mechanisms may be required for state-space models to compete effectively in this domain.\\

\begin{figure*}
    \centering
    \begin{subfigure}[b]{0.24\linewidth}
        \includegraphics[width=\linewidth]{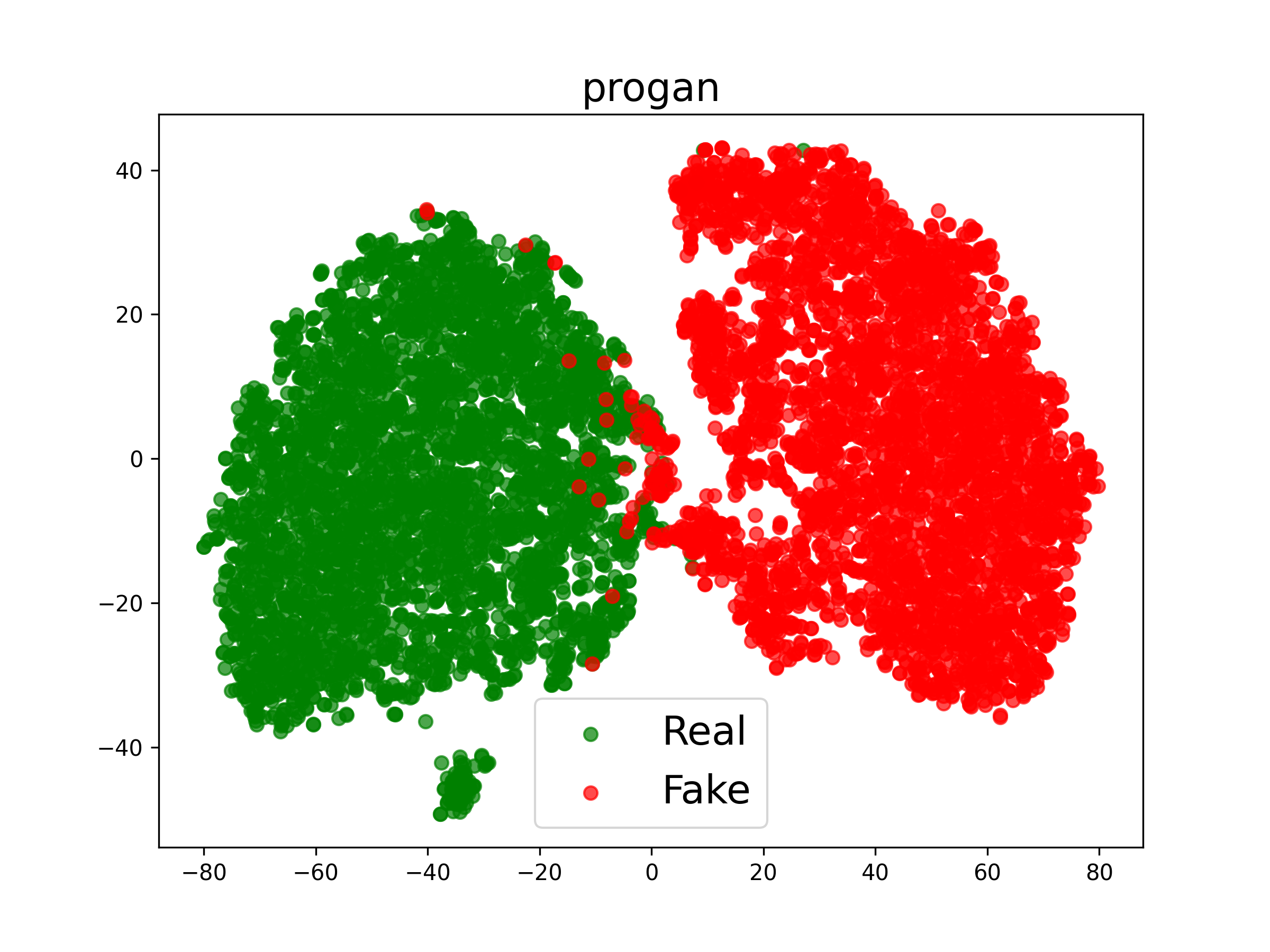}
        %\caption*{progan \hspace{8mm} }
    \end{subfigure}
%    \hspace{.1in}
   \vspace{.001in}
    \begin{subfigure}[b]{0.24\linewidth}
        \includegraphics[width=\linewidth]{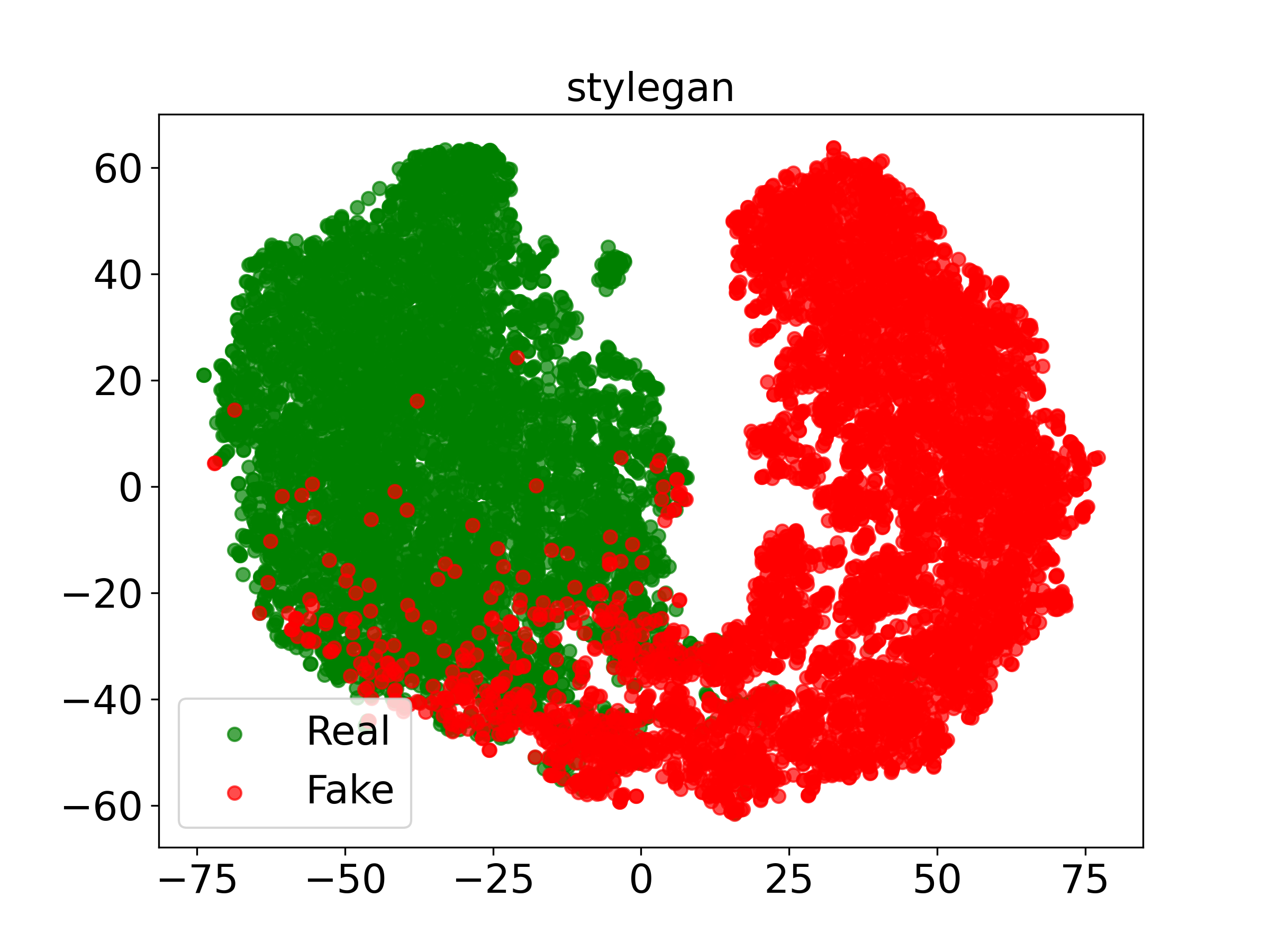}
        %\caption*{stylegan \hspace{15mm}}
    \end{subfigure}
   % \hspace{.1in}
   \vspace{.001in}
    \begin{subfigure}[b]{0.24\linewidth}
        \includegraphics[width=\linewidth]{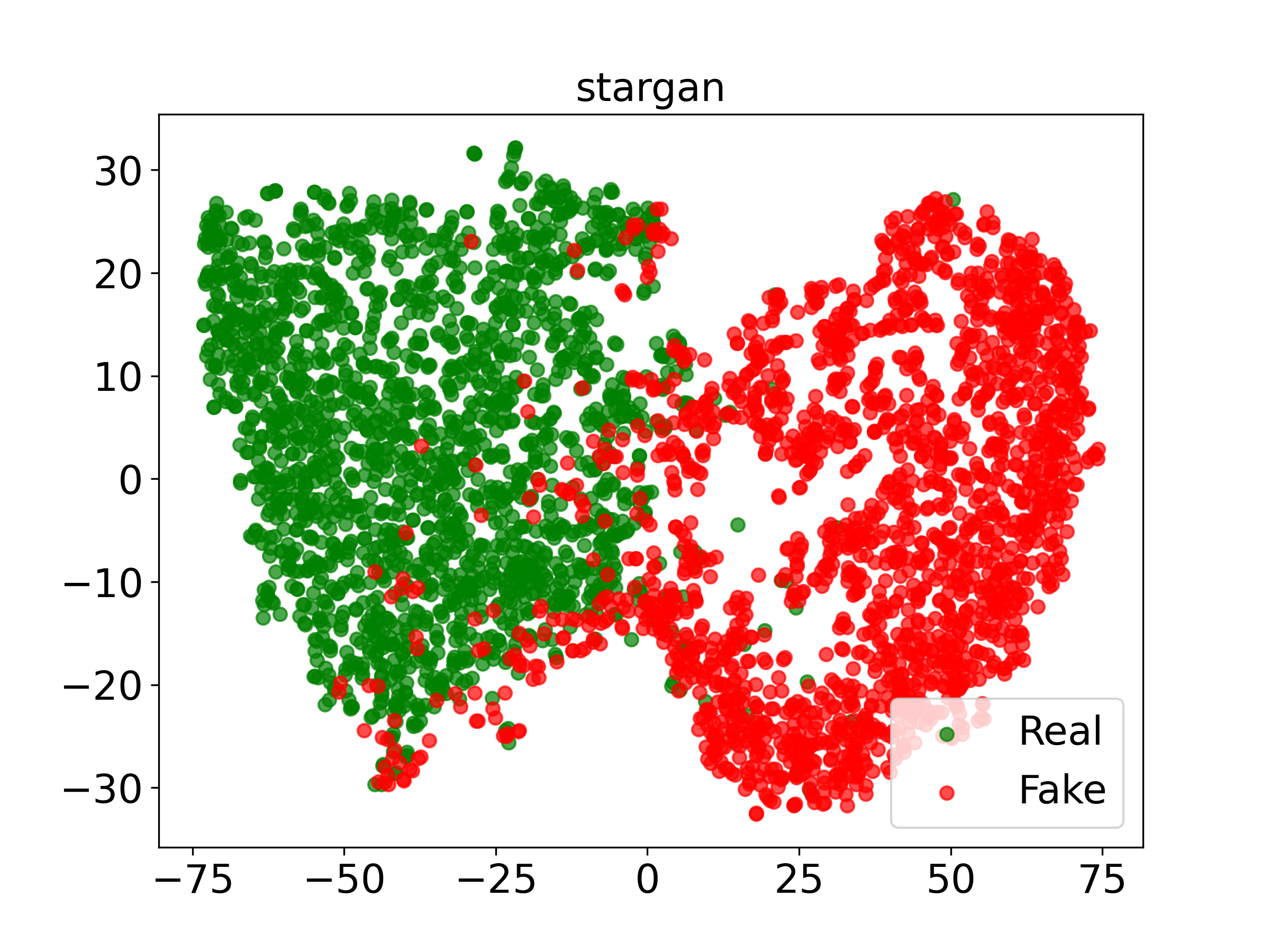}
        %\caption*{stargan \hspace{5mm}}
    \end{subfigure}
   % \hspace{.1in}
   \vspace{.001in}
    \begin{subfigure}[b]{0.24\linewidth}
        \includegraphics[width=\linewidth]{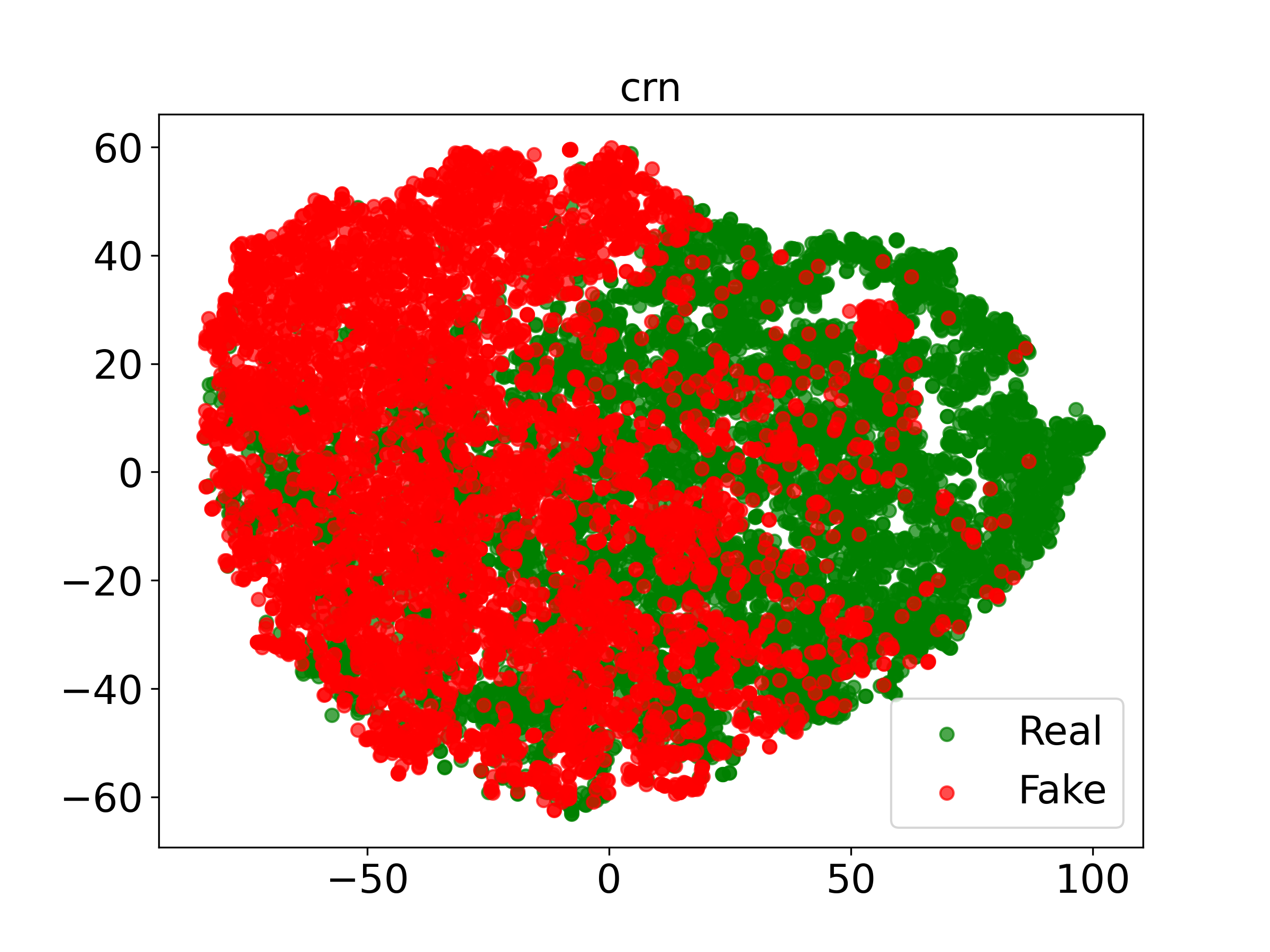}
        %\caption*{crn\hspace{5mm}}
    \end{subfigure}
    \begin{subfigure}[b]{0.24\linewidth}
\includegraphics[width=\linewidth]{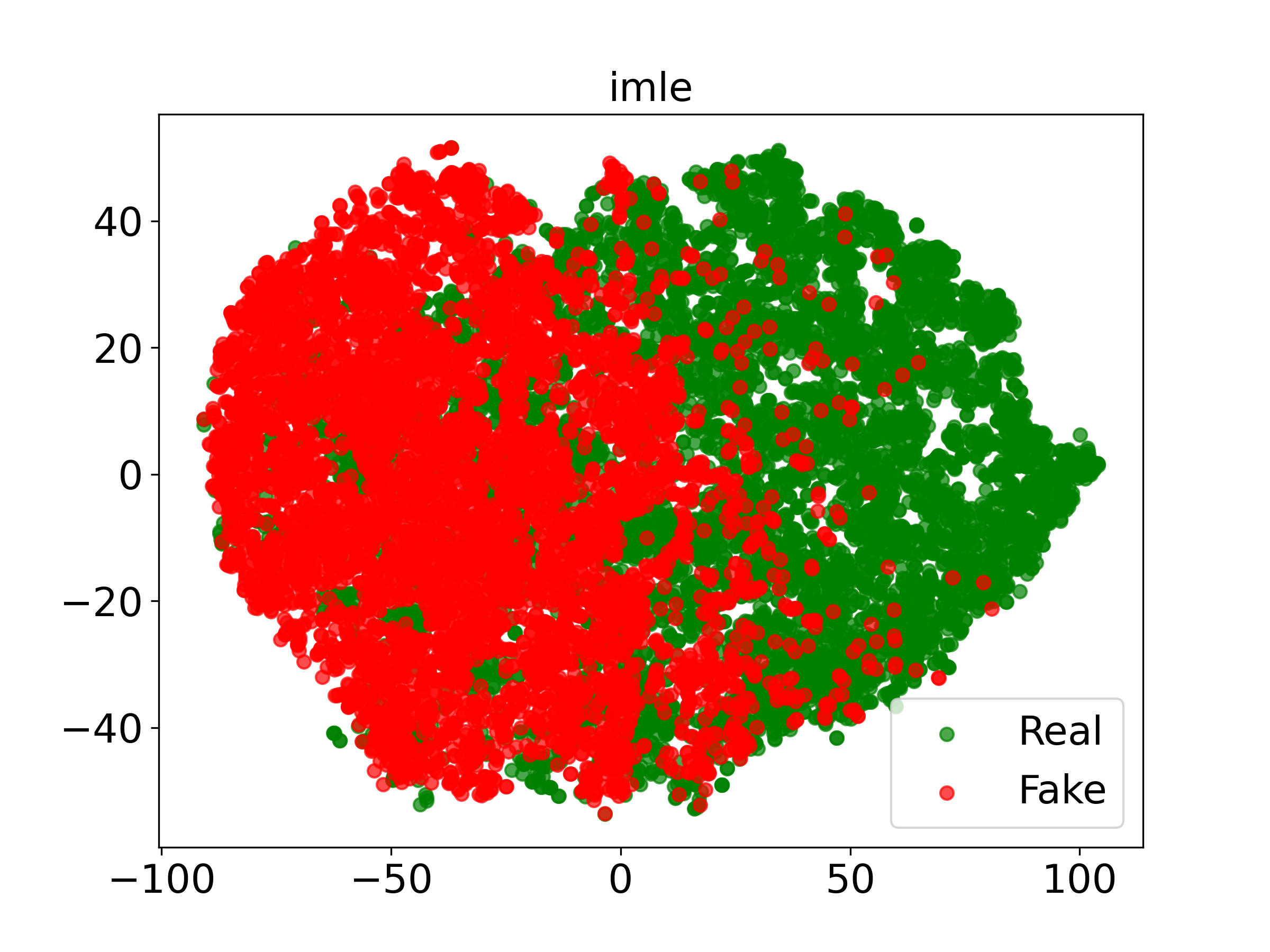}
        %\caption*{imle}
    \end{subfigure}
  %  \hspace{.1in}
   \vspace{.001in}
    \begin{subfigure}[b]{0.24\linewidth}
        \includegraphics[width=\linewidth]{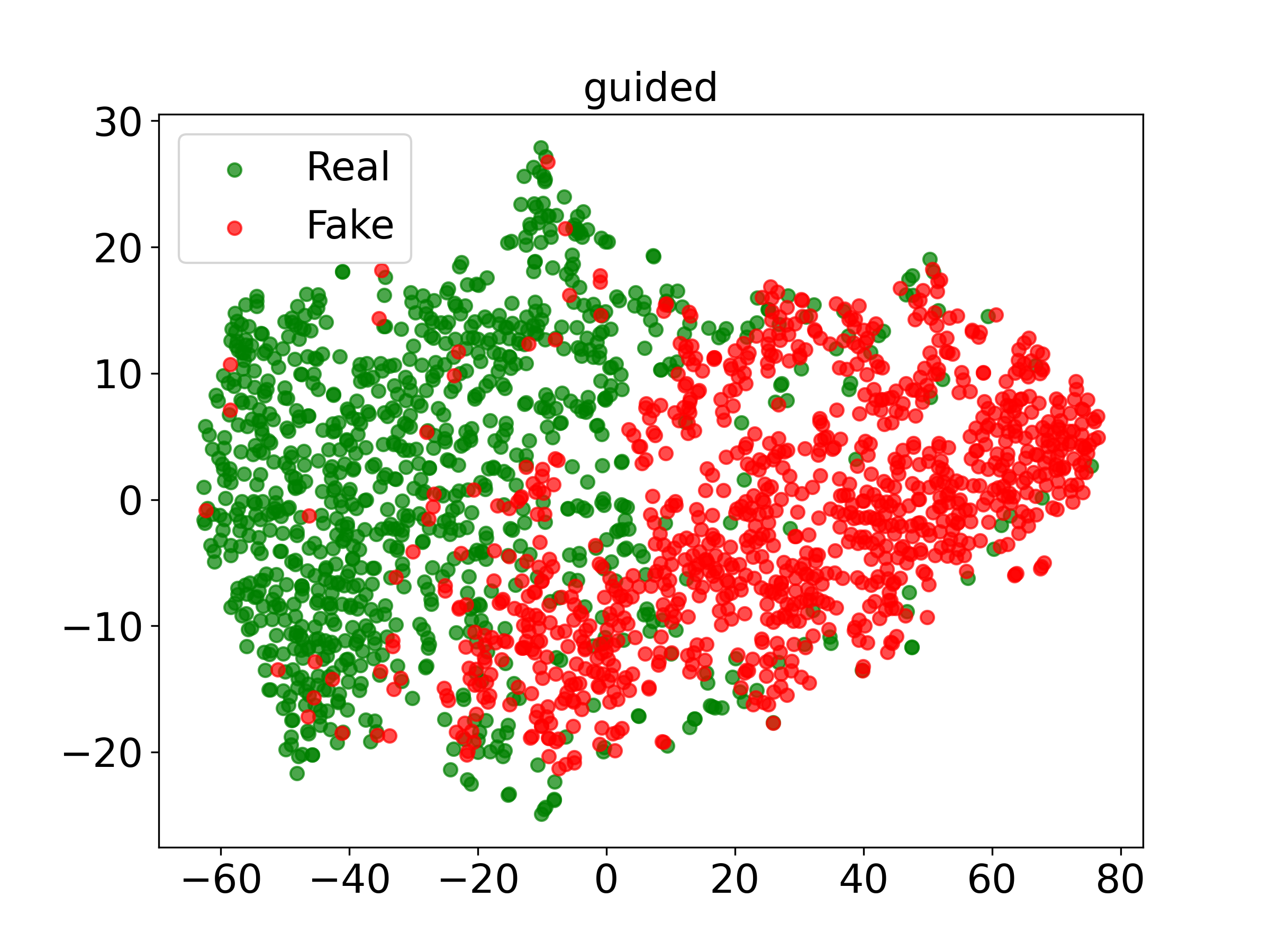}
        %\caption*{guided}
    \end{subfigure}
    % \hspace{.1in}
    \vspace{.001in}
    \begin{subfigure}[b]{0.24\linewidth}     \includegraphics[width=\linewidth]{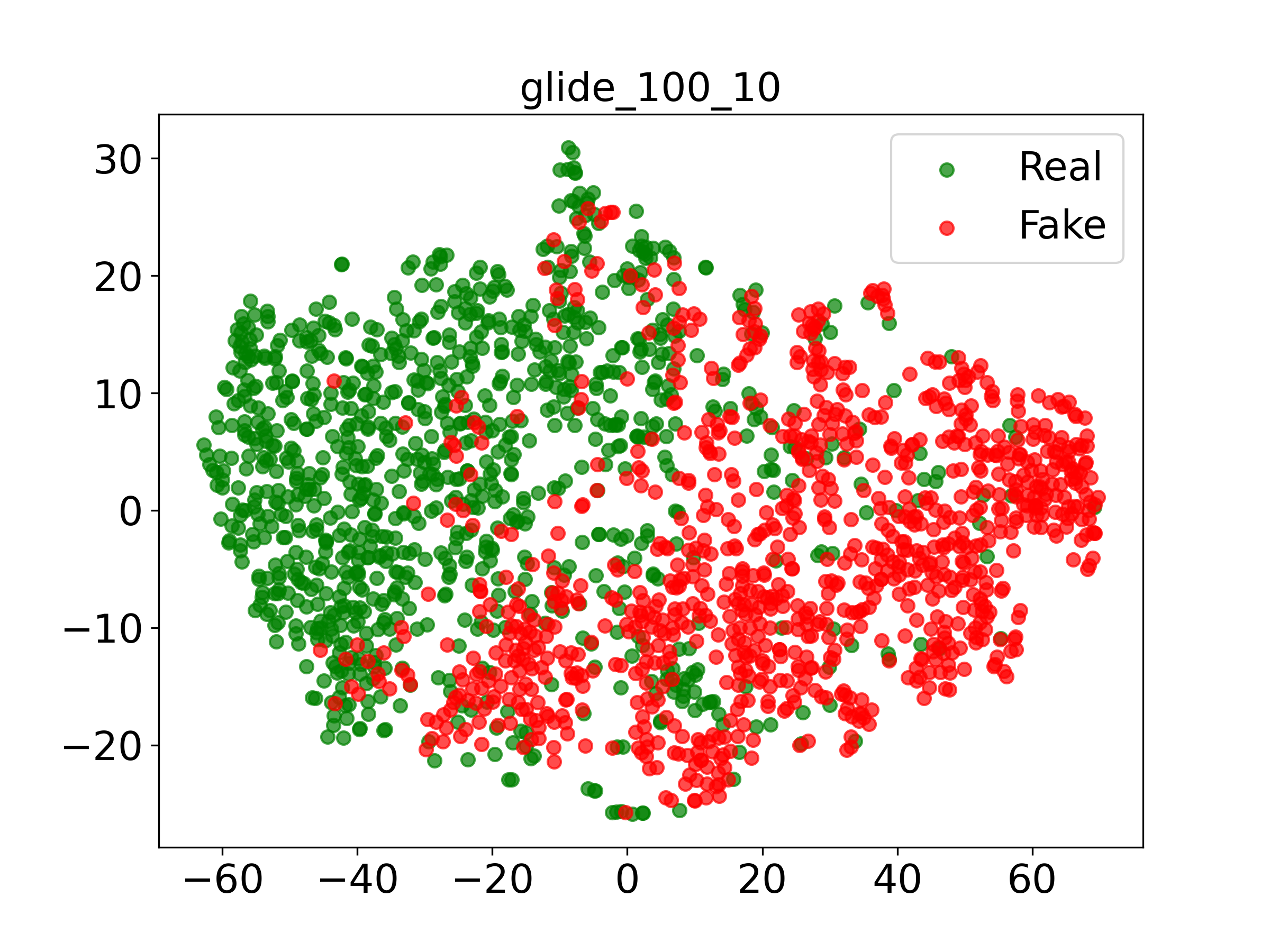}
        %\caption*{glide\_100\_10}
    \end{subfigure}
%    \hspace{.1in}
   \vspace{.001in}
     \begin{subfigure}[b]{0.24\linewidth}     \includegraphics[width=\linewidth]{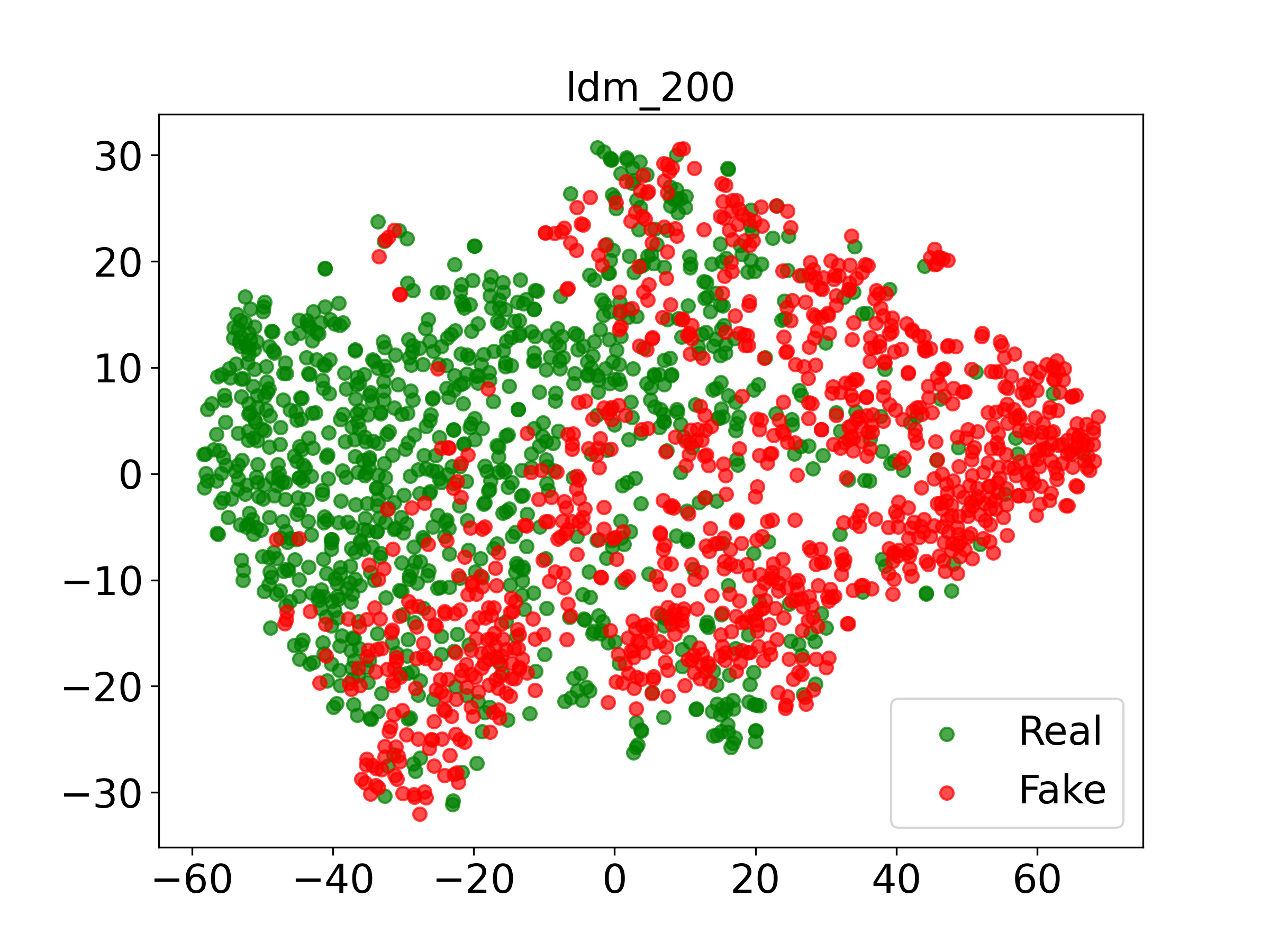}
         %\caption*{ldm\_200}
     \end{subfigure}  
    \caption{t-SNE Visualization of Feature Distributions of Vim Model. The scatter plots illustrate the t-SNE embeddings of features extracted from real (green) and generated (red) images across various generative models, showing how well the features separate real from fake images.}
    \label{fig:tsneVIM} 
\end{figure*}

\begin{figure*}
    \centering
    \begin{subfigure}[b]{0.24\linewidth}
        \includegraphics[width=\linewidth]{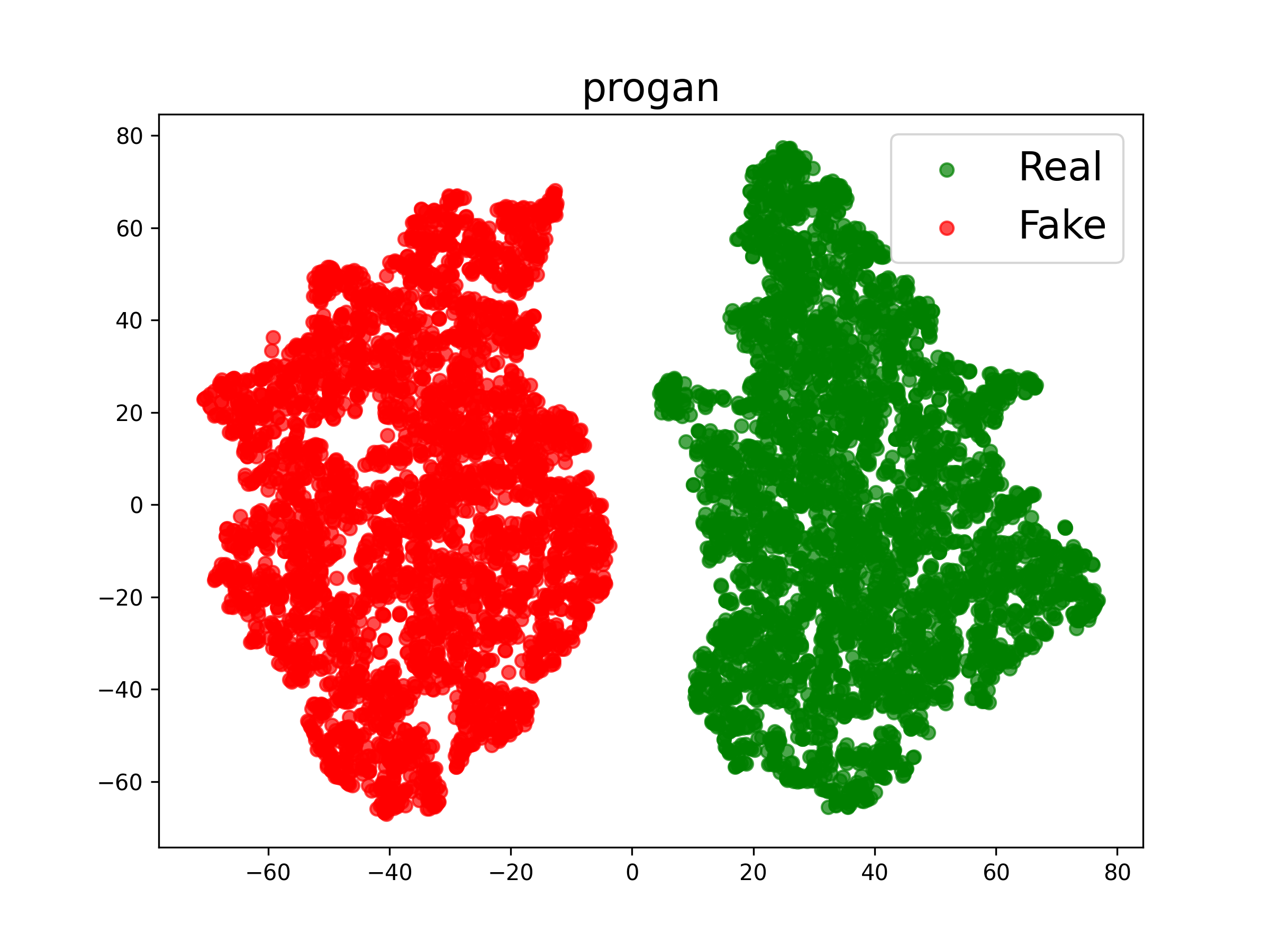}
        %\caption*{progan \hspace{8mm} }
    \end{subfigure}
%    \hspace{.1in}
   \vspace{.001in}
    \begin{subfigure}[b]{0.24\linewidth}
        \includegraphics[width=\linewidth]{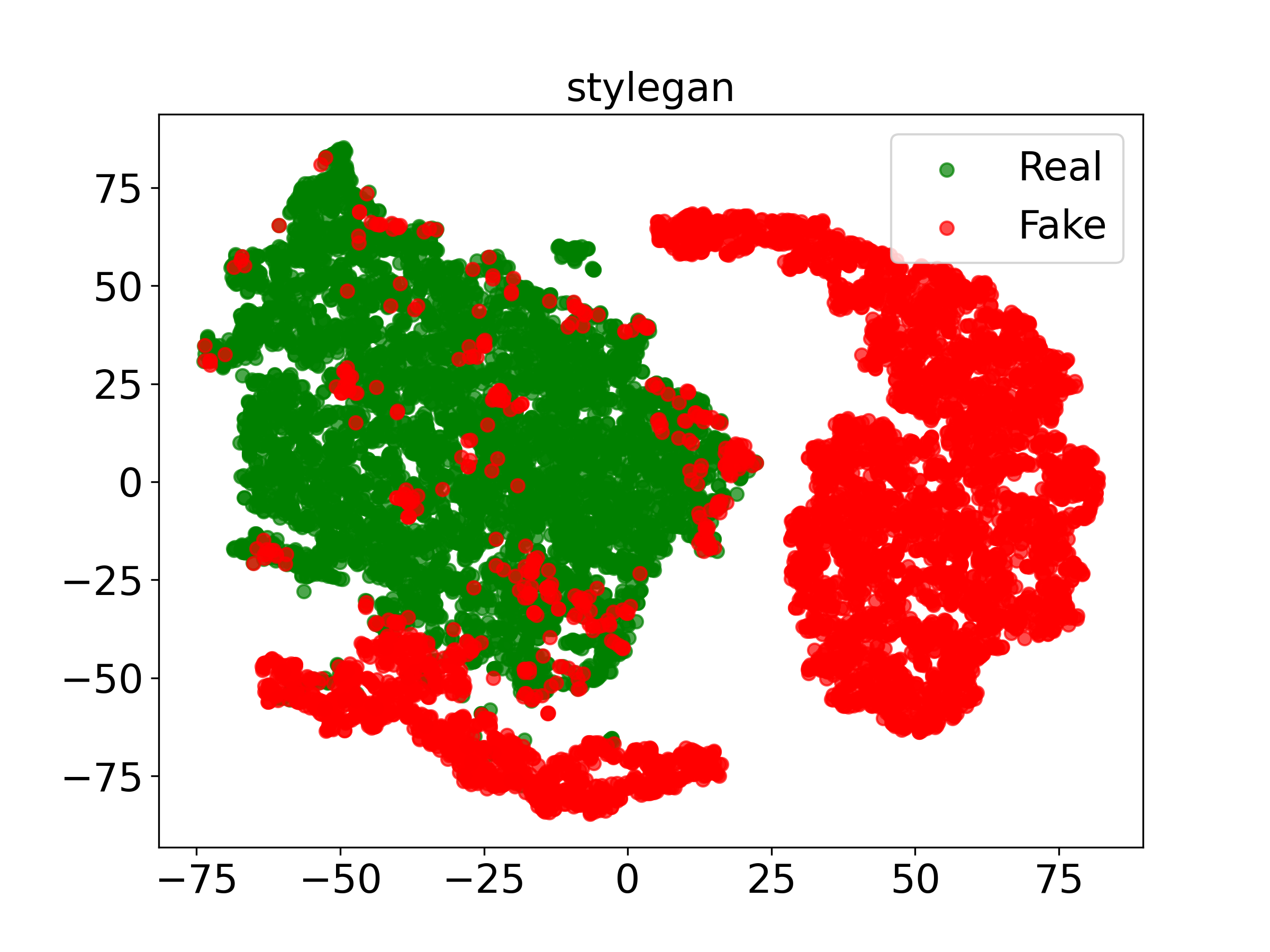}
        %\caption*{stylegan \hspace{15mm}}
    \end{subfigure}
   % \hspace{.1in}
   \vspace{.001in}
    \begin{subfigure}[b]{0.24\linewidth}
        \includegraphics[width=\linewidth]{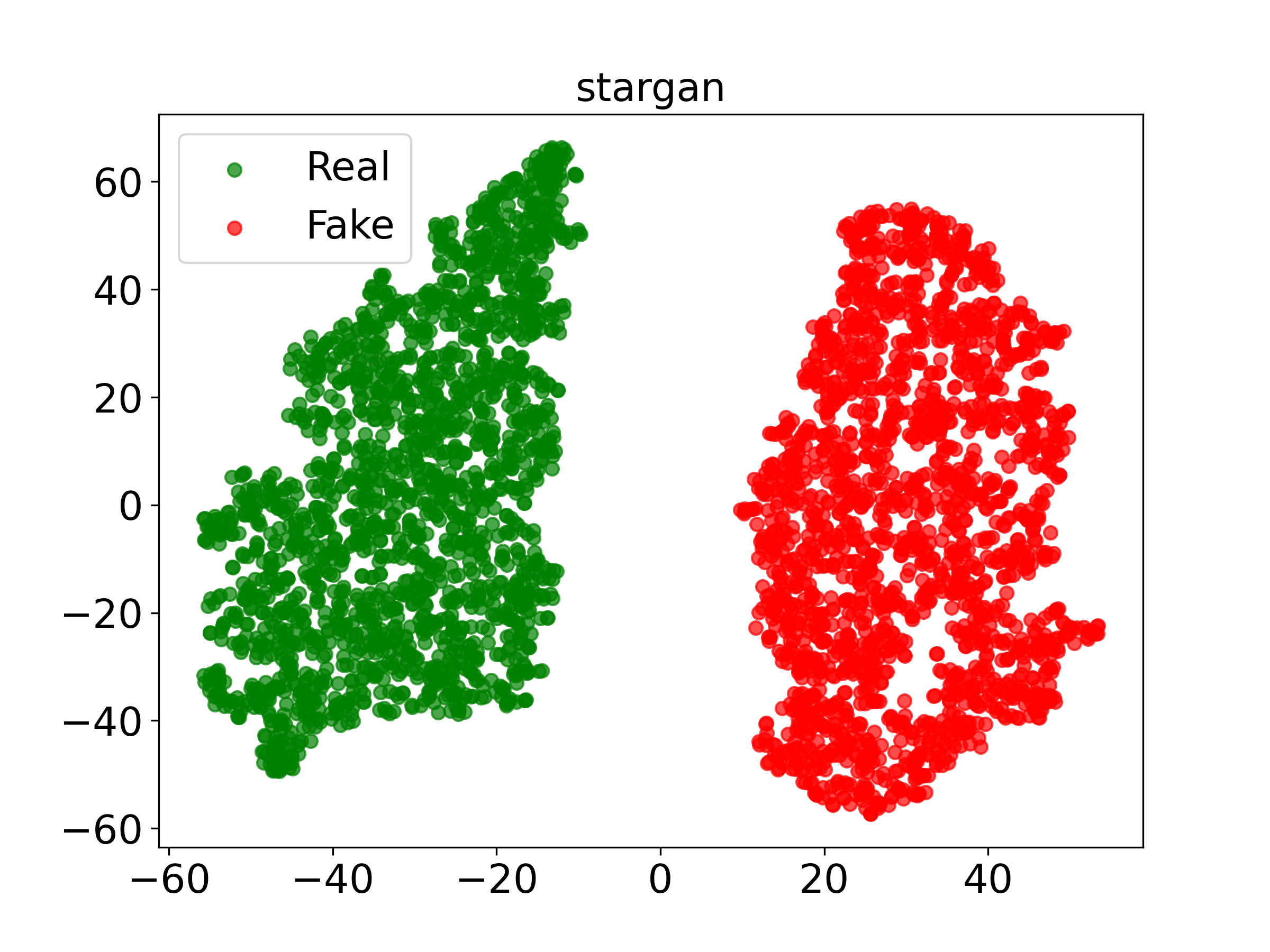}
        %\caption*{stargan \hspace{5mm}}
    \end{subfigure}
   % \hspace{.1in}
   \vspace{.001in}
    \begin{subfigure}[b]{0.24\linewidth}
        \includegraphics[width=\linewidth]{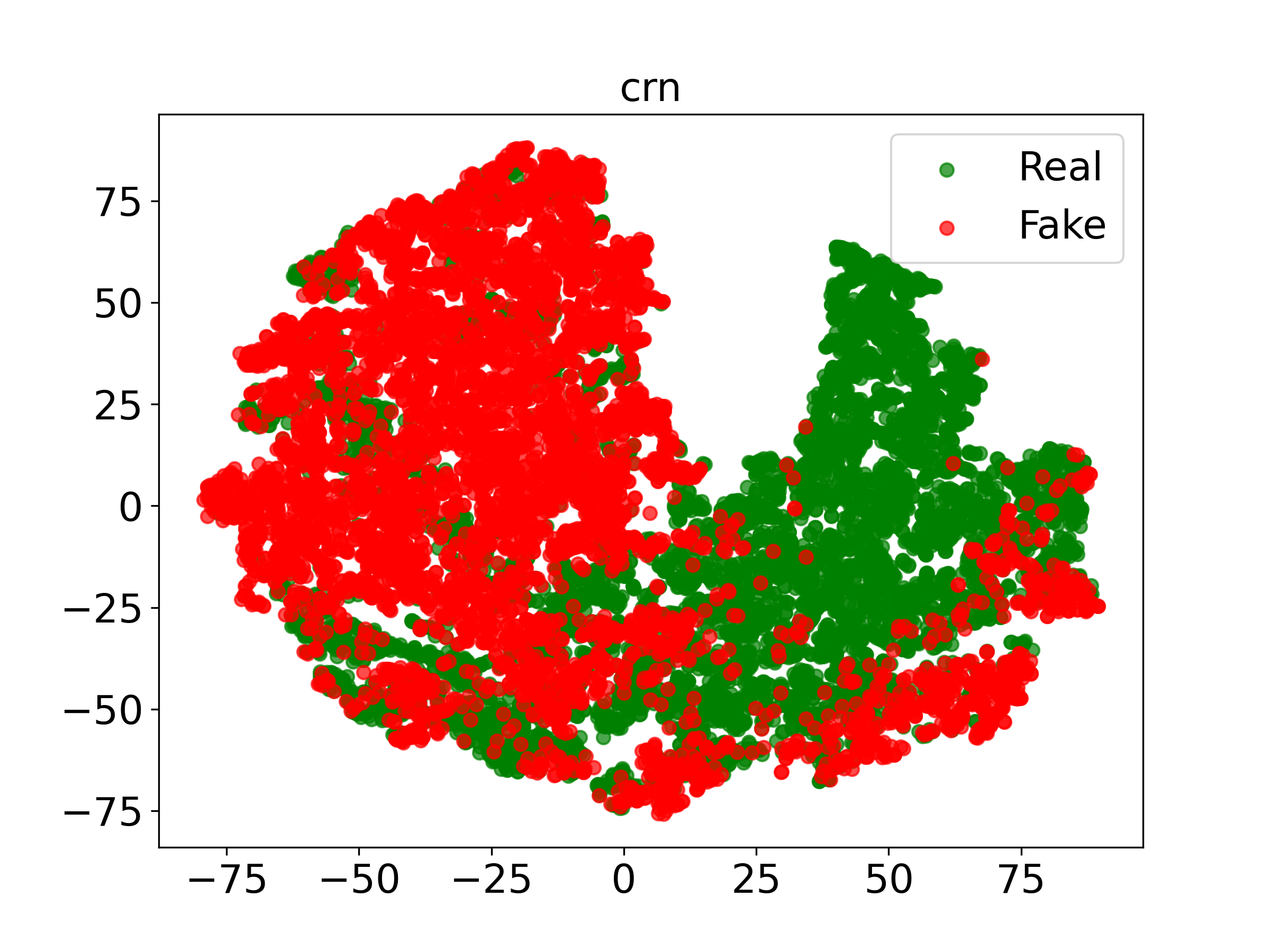}
        %\caption*{crn\hspace{5mm}}
    \end{subfigure}
    \begin{subfigure}[b]{0.24\linewidth}
\includegraphics[width=\linewidth]{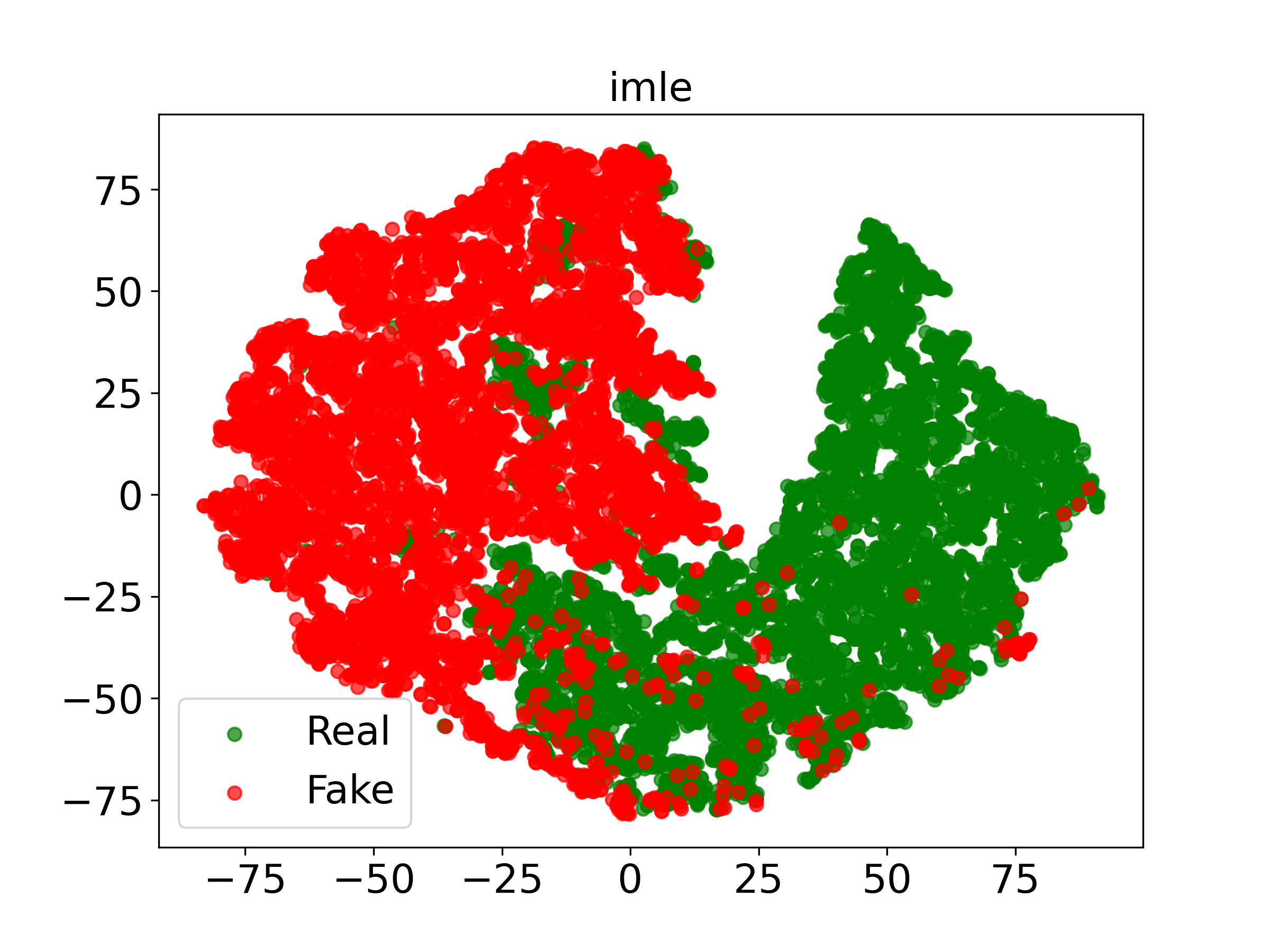}
        %\caption*{imle}
    \end{subfigure}
  %  \hspace{.1in}
   \vspace{.001in}
    \begin{subfigure}[b]{0.24\linewidth}
        \includegraphics[width=\linewidth]{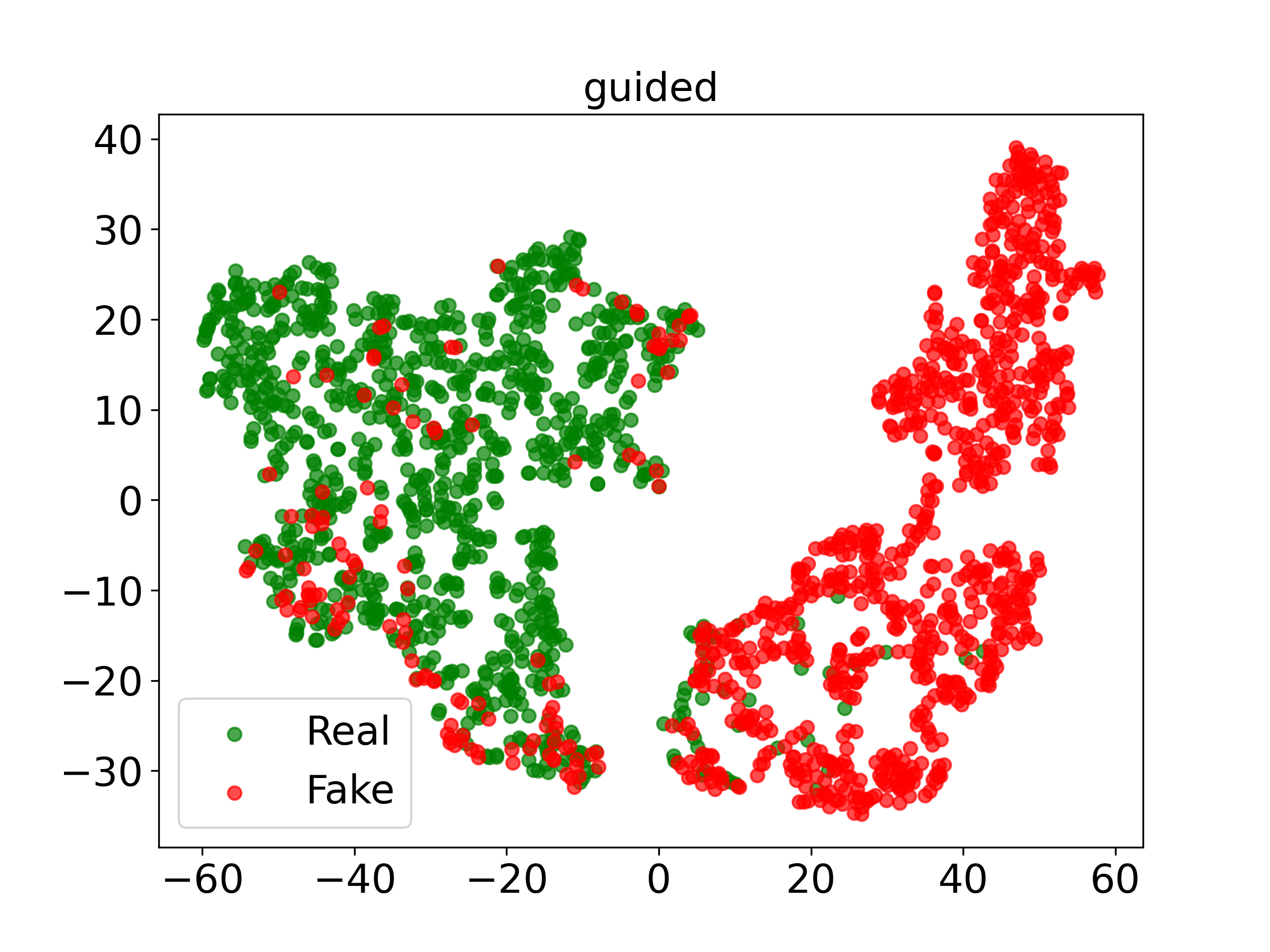}
        %\caption*{guided}
    \end{subfigure}
    % \hspace{.1in}
    \vspace{.001in}
    \begin{subfigure}[b]{0.24\linewidth}     \includegraphics[width=\linewidth]{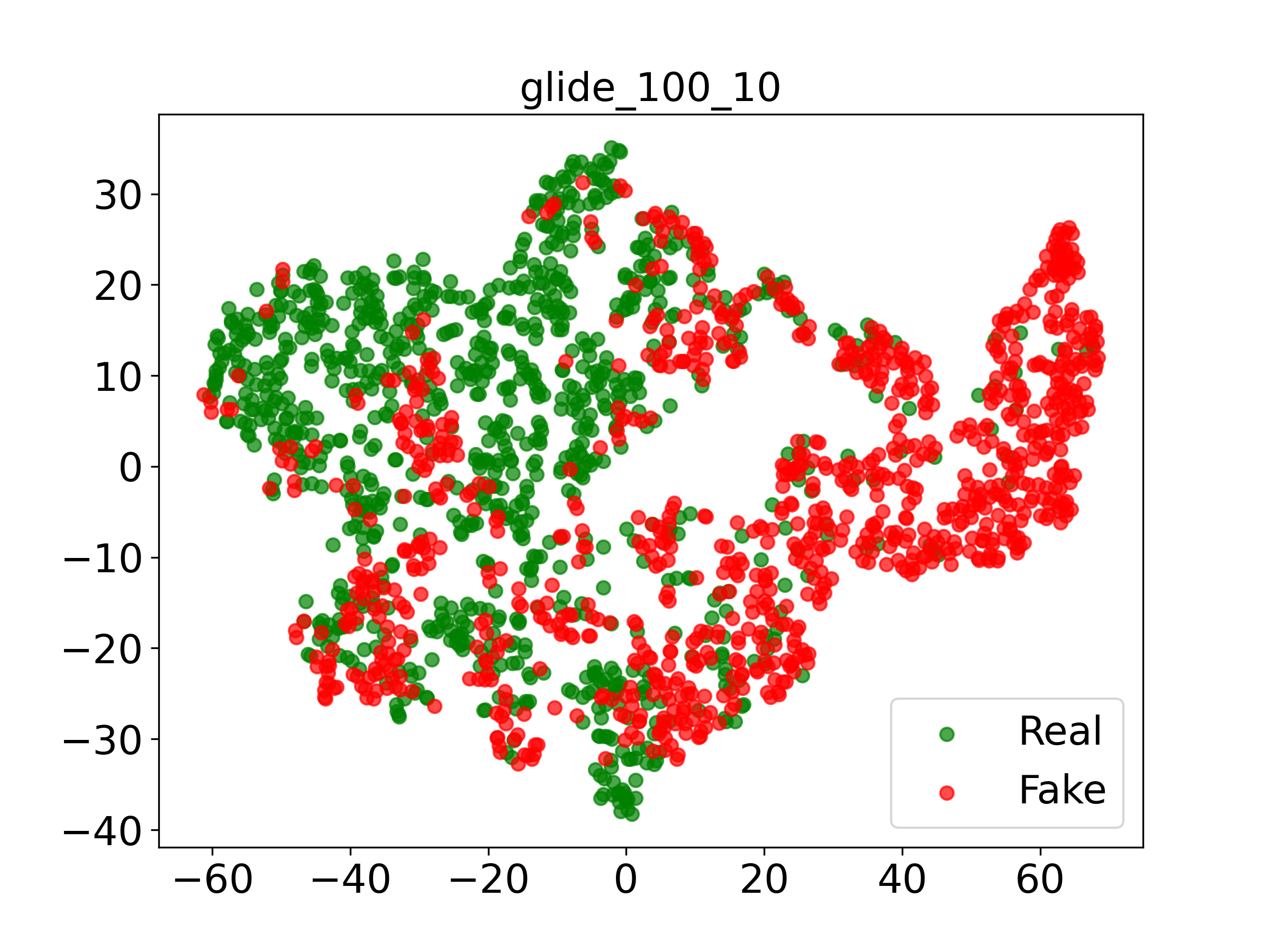}
        %\caption*{glide\_100\_10}
    \end{subfigure}
%    \hspace{.1in}
   \vspace{.001in}
     \begin{subfigure}[b]{0.24\linewidth}     \includegraphics[width=\linewidth]{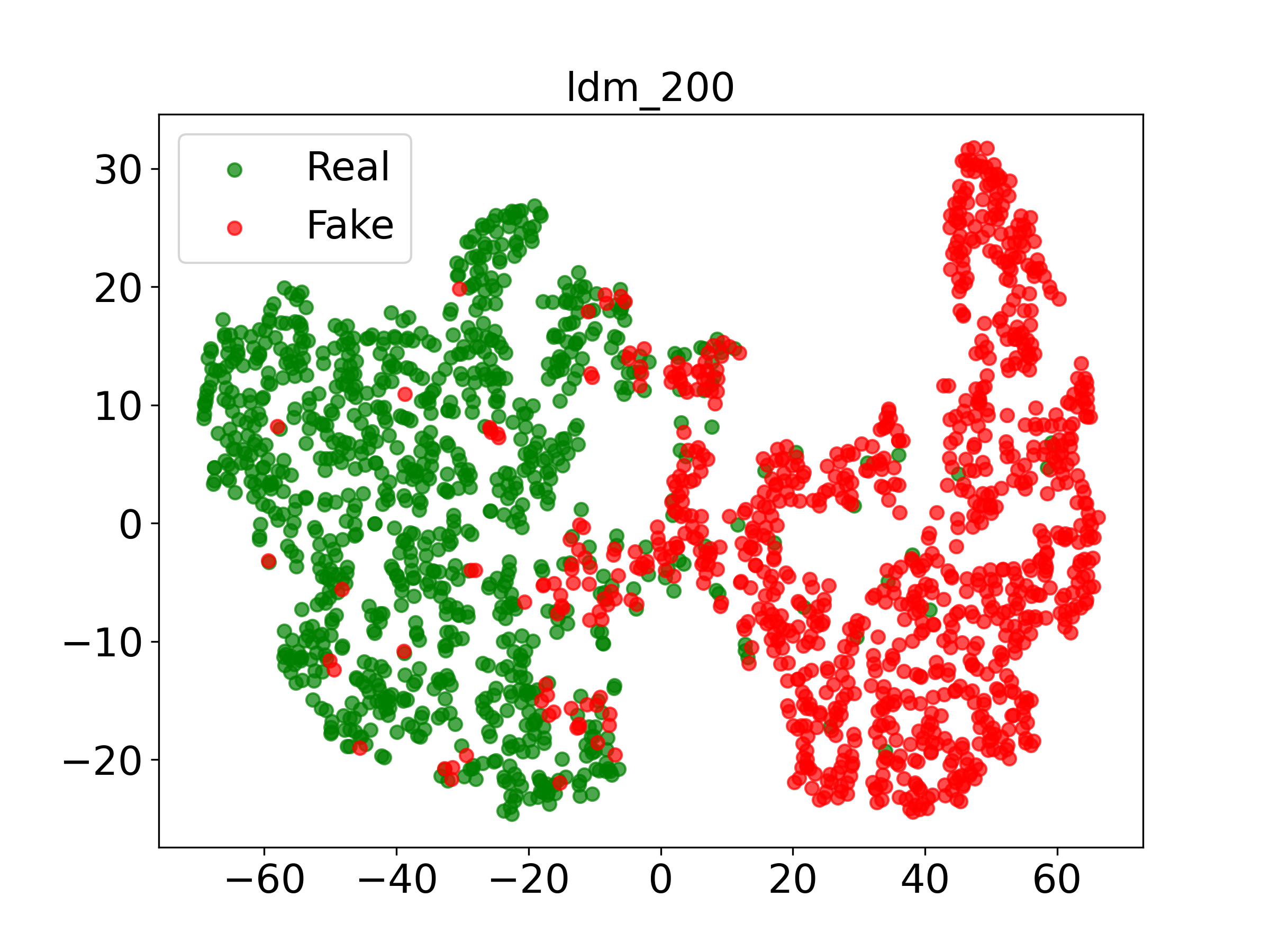}
         %\caption*{ldm\_200}
     \end{subfigure}  
    \caption{t-SNE Visualization of Feature Distributions of VSSD Model. The scatter plots illustrate the t-SNE embeddings of features extracted from real (green) and generated (red) images across various generative models, showing how well the features separate real from fake images.}
    \label{fig:tsneVSSD} 
\end{figure*}

\begin{figure*}
    \centering
    \begin{subfigure}[b]{0.24\linewidth}
        \includegraphics[width=\linewidth]{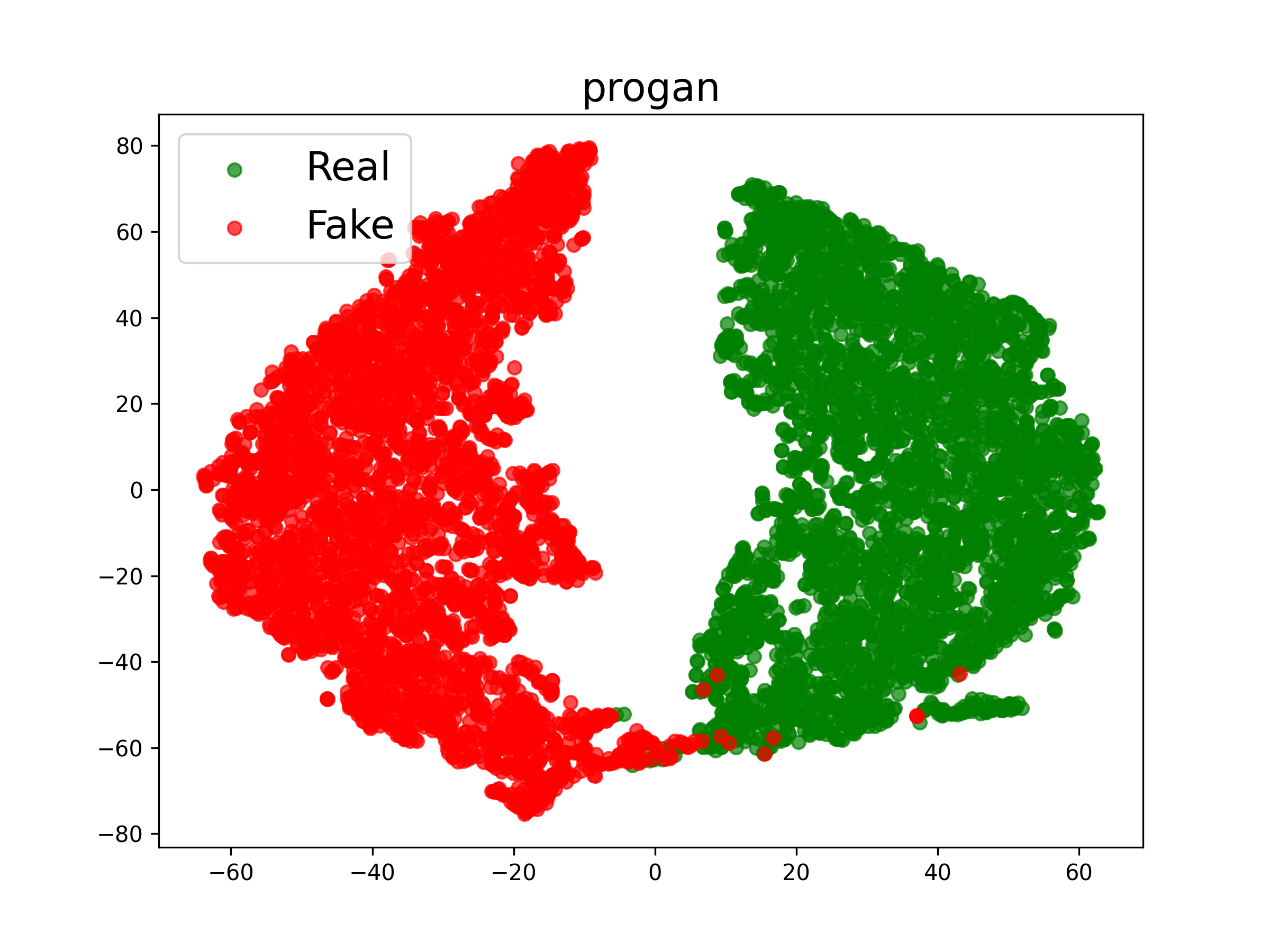}
        %\caption*{progan \hspace{8mm} }
    \end{subfigure}
%    \hspace{.1in}
   \vspace{.001in}
    \begin{subfigure}[b]{0.24\linewidth}
        \includegraphics[width=\linewidth]{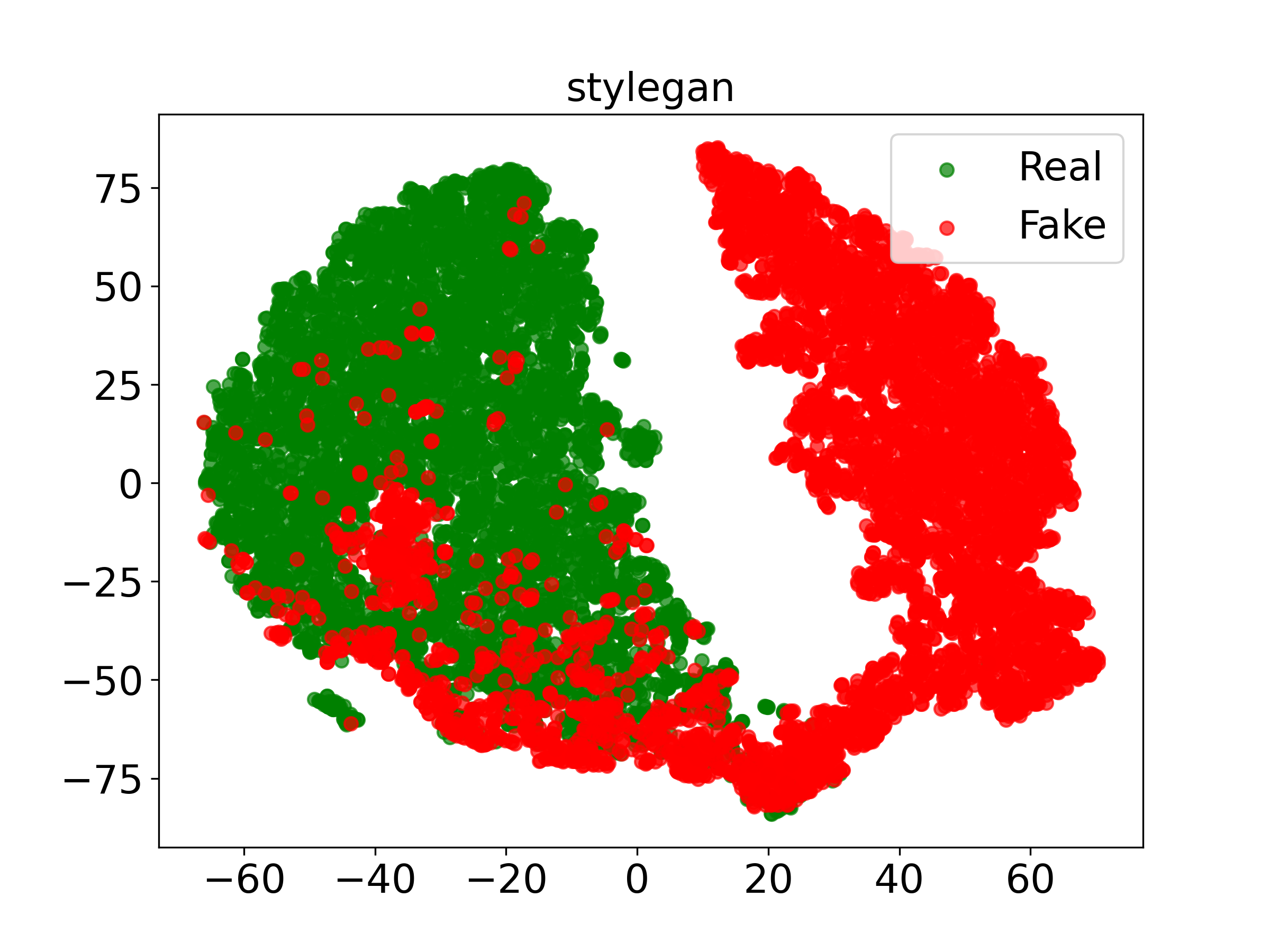}
        %\caption*{stylegan \hspace{15mm}}
    \end{subfigure}
   % \hspace{.1in}
   \vspace{.001in}
    \begin{subfigure}[b]{0.24\linewidth}
        \includegraphics[width=\linewidth]{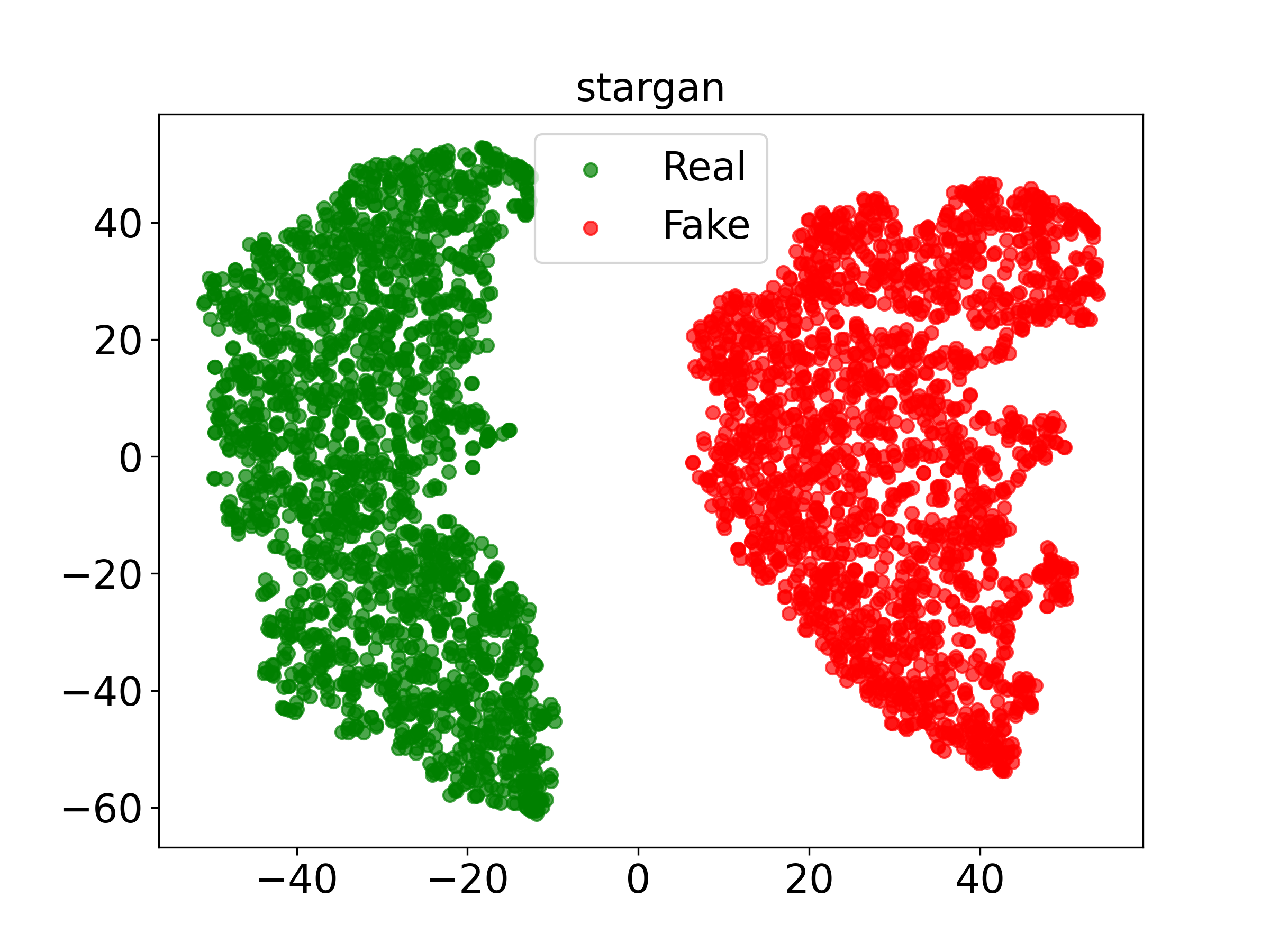}
        %\caption*{stargan \hspace{5mm}}
    \end{subfigure}
   % \hspace{.1in}
   \vspace{.001in}
    \begin{subfigure}[b]{0.24\linewidth}
        \includegraphics[width=\linewidth]{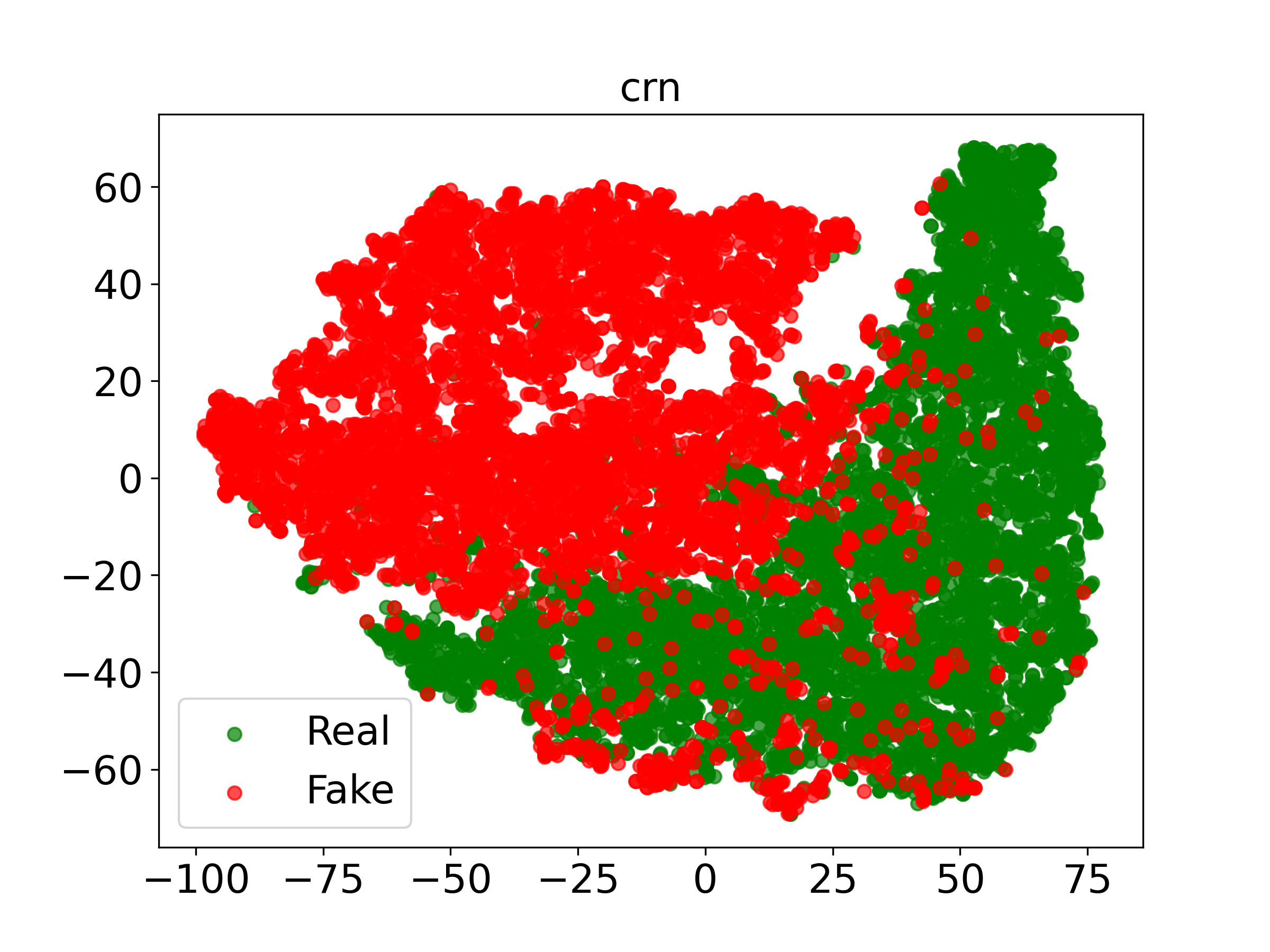}
        %\caption*{crn\hspace{5mm}}
    \end{subfigure}
    \begin{subfigure}[b]{0.24\linewidth}
\includegraphics[width=\linewidth]{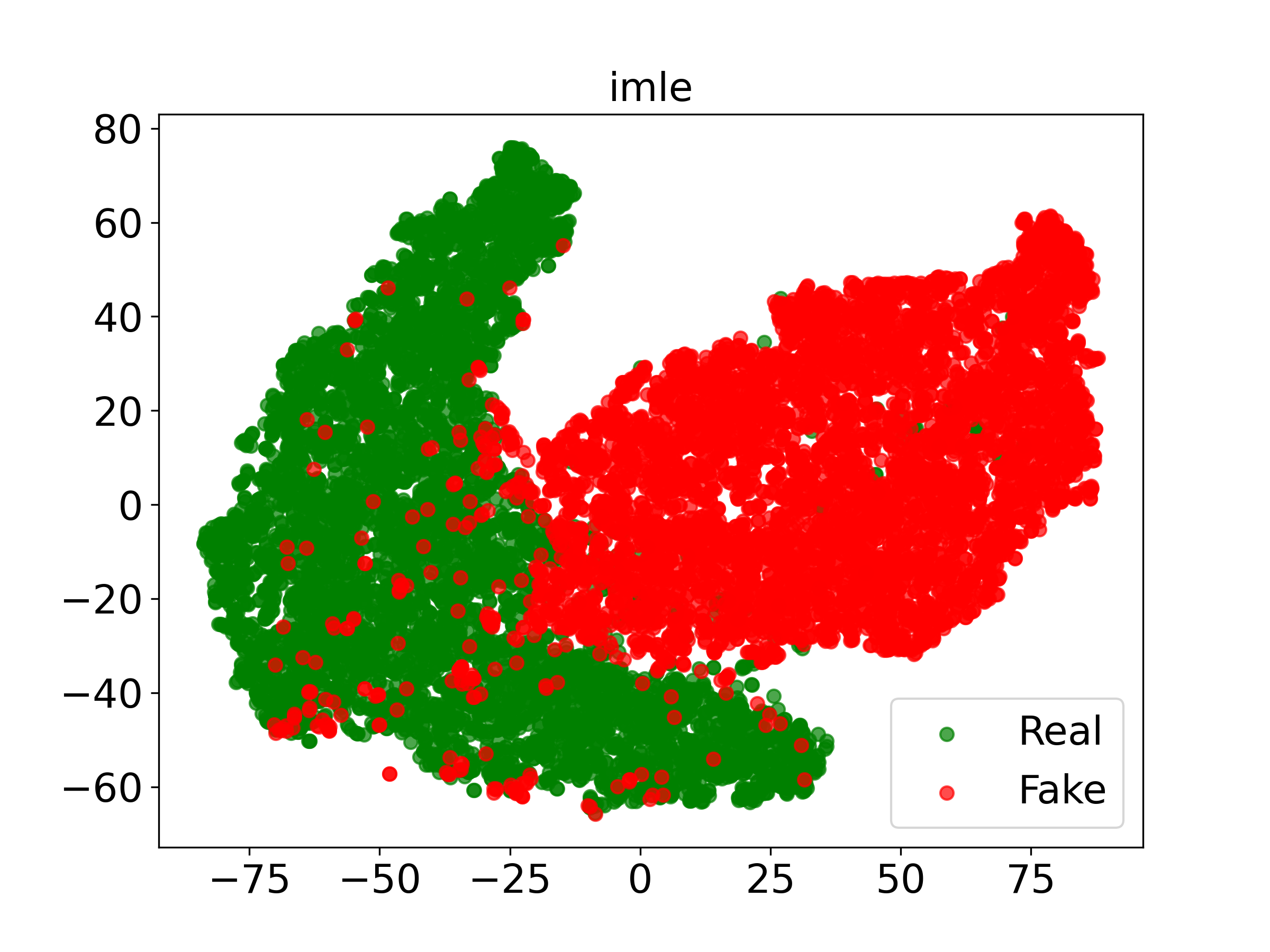}
        %\caption*{imle}
    \end{subfigure}
  %  \hspace{.1in}
   \vspace{.001in}
    \begin{subfigure}[b]{0.24\linewidth}
        \includegraphics[width=\linewidth]{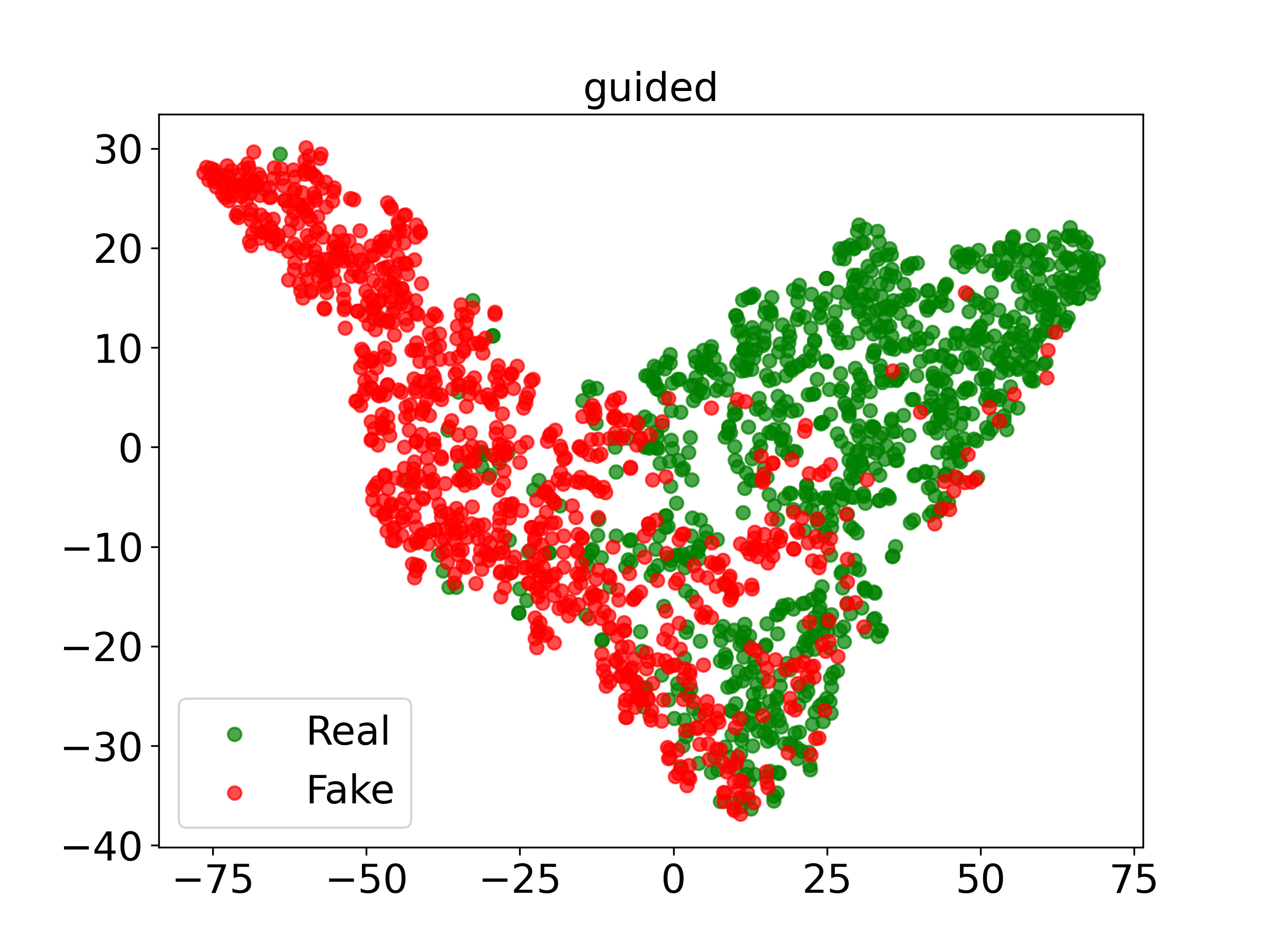}
        %\caption*{guided}
    \end{subfigure}
    % \hspace{.1in}
    \vspace{.001in}
    \begin{subfigure}[b]{0.24\linewidth}     \includegraphics[width=\linewidth]{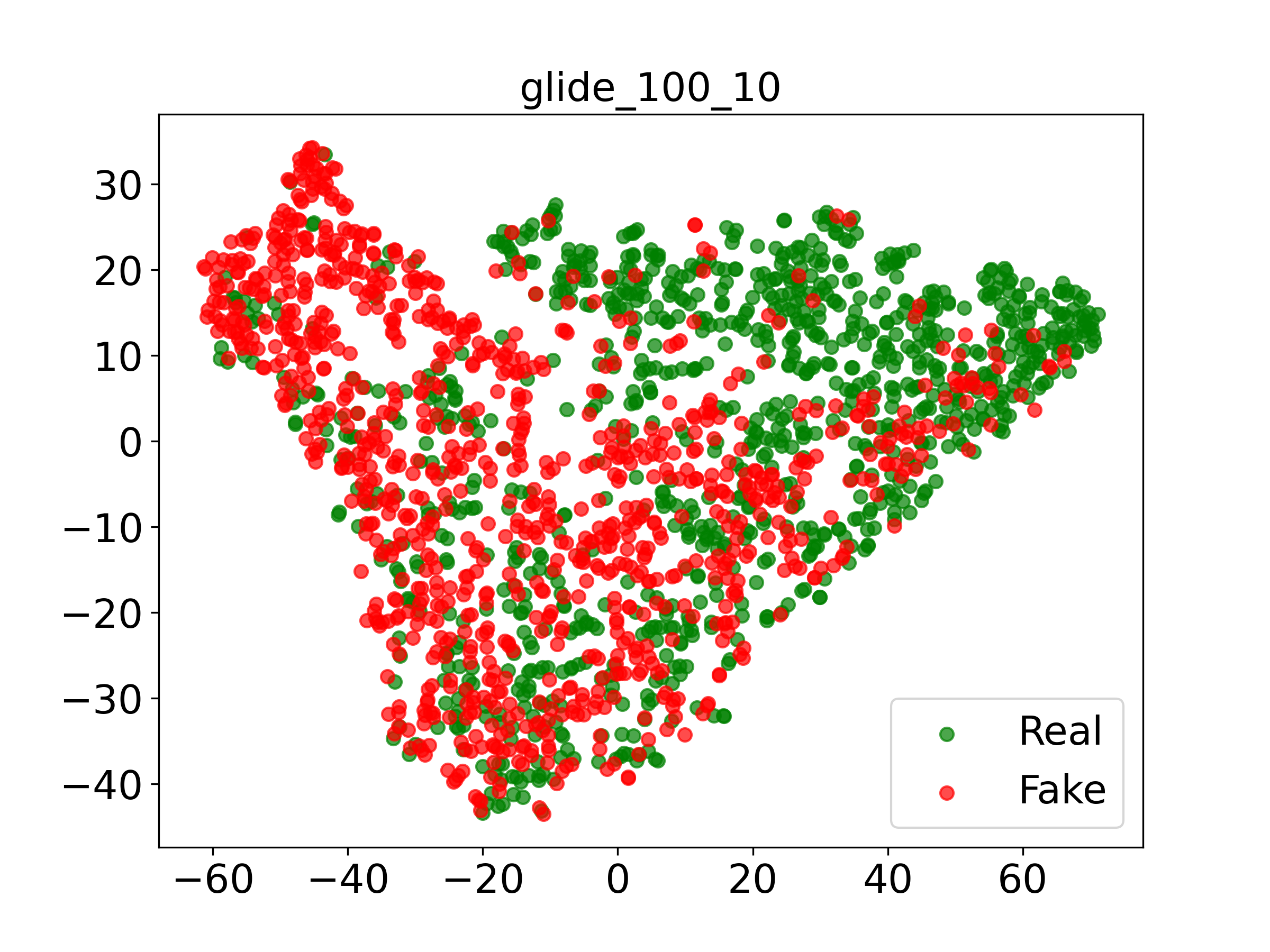}
        %\caption*{glide\_100\_10}
    \end{subfigure}
%    \hspace{.1in}
   \vspace{.001in}
     \begin{subfigure}[b]{0.24\linewidth}     \includegraphics[width=\linewidth]{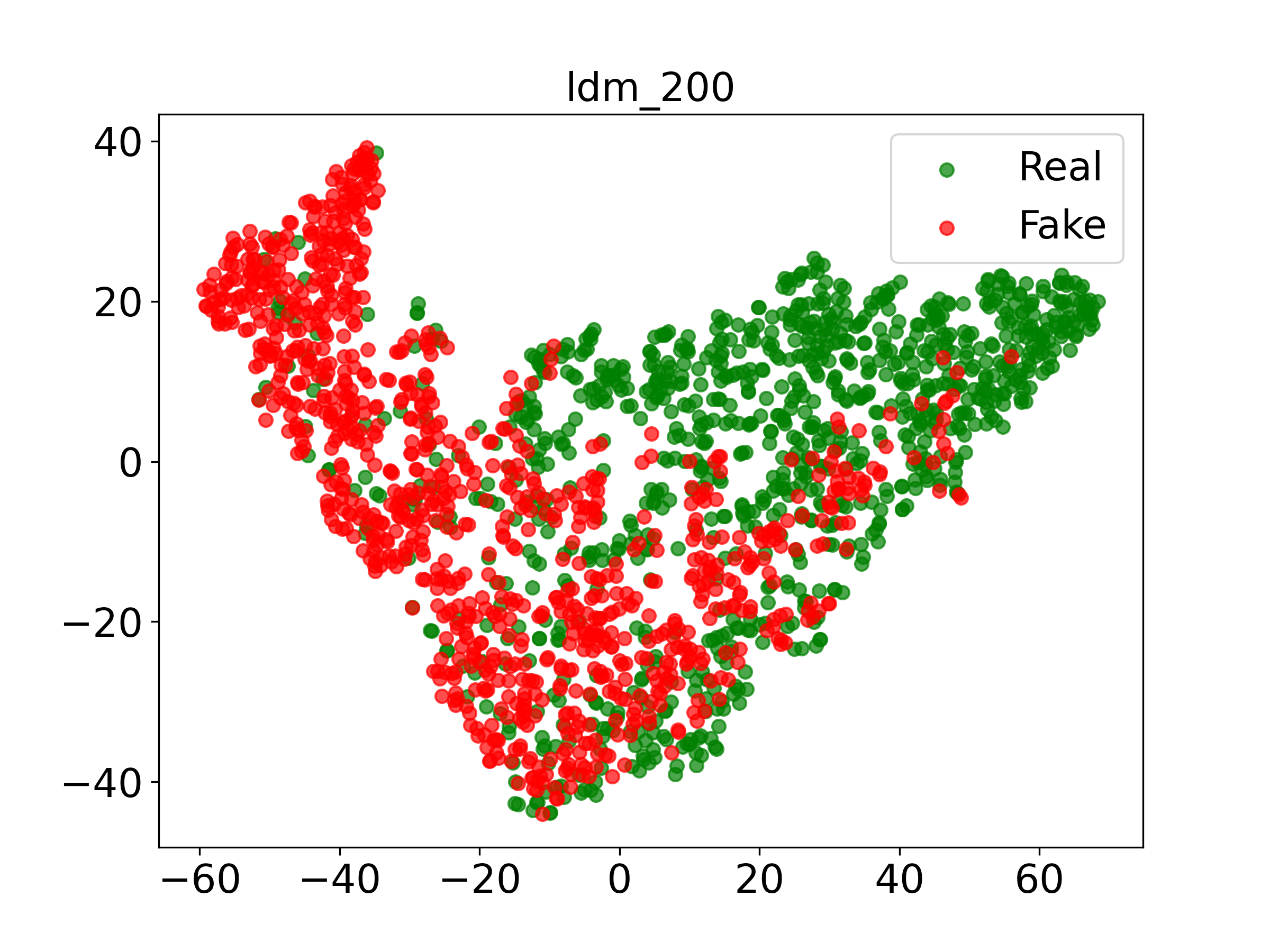}
         %\caption*{ldm\_200}
     \end{subfigure}  
    \caption{t-SNE Visualization of Feature Distributions of MambaVision Model. The scatter plots illustrate the t-SNE embeddings of features extracted from real (green) and generated (red) images across various generative models, showing how well the features separate real from fake images.}
    \label{fig:tsneMambaVision} 
\end{figure*}

\noindent {\textbf{Feature Visualization.}} To further analyze the effectiveness of vision mamba models, we use t-SNE visualization to illustrate the feature space of the detectors on UniversalFakeDetect, as shown in Figures~\ref{fig:tsneVIM},~\ref{fig:tsneVSSD}, and~\ref{fig:tsneMambaVision}. Each detector is trained on 4-class progan training set of UniversalFakeDetect dataset. We used the entire samples of each testing subset for visualization. Features are extracted from the layer before the classification layer.

\noindent Figures~\ref{fig:tsneVIM},~\ref{fig:tsneVSSD}, and~\ref{fig:tsneMambaVision} show that images features from different classes have distinct distributions, each marked by a unique color, green for real images and red for fake images. It can be seen that the detectors can separate the two classes well for certain testing subsets (i.e; progan, stargan), due to the wide margins between the distributions. However, there is also a limit for other test subsets with a margin unclear. This can be attributed either to the fact that these generative models produce images with features more similar to real ones or to the fact that the detectors have difficulty in capturing generalizable features to unseen data distributions.

\subsection{Vision Mamba's challenges and limitations}
The relatively poor performance of Vision Mamba models in detecting high-fidelity \acs{ai}-generated images raises significant concerns regarding their applicability in real-world scenarios. Their accuracy drops sharply when confronted with images from nearly all \acs{ai}-image generators tested, including SDXL, SD2+SR, ADM, and DDPM. This suggests a fundamental difficulty in adapting to new and complex image distributions. This limitation may stem from the underlying \acs{ssm} architecture, which appears unable to effectively capture the fine-grained hierarchical features necessary to differentiate between real and synthetic images. Furthermore, the inability of Vision Mamba models to effectively handle adversarial attacks, as evidenced by their poor accuracy on the adversarial, backdoor, and data poisoning datasets in Table~\ref{tab:TrainedOnAntifakePrompt}, underscores another significant limitation. These models demonstrate substantial vulnerability to such manipulations, which is particularly concerning given the increasing sophistication of adversarial techniques designed to evade detection systems. The investigation into Vision Mamba's effectiveness in detecting \acs{ai}-generated images reveals a dichotomy between its potential and limitations. Our experiments indicate that Vision Mamba performs well when trained and tested on images generated by the same model, demonstrating a capacity to capture specific characteristics inherent to that generator. However, its performance significantly deteriorates when presented with images produced by different generators, highlighting a critical shortcoming in its generalization ability. \\

% In vision transformer-based models, the attention matrix explicitly captures the dependencies between each pair of visual tokens, allowing predictions at any location to have a global receptive field. However, in Mamba-based models the S6 blocks make predictions for each token based on the hidden state $h_{t-1}$ and the current input $x_t$. The hidden state $h_{t-1}$ incorporates specifically information from previously analyzed tokens and ignores information from other tokens. Consequently, the vanilla Mamba is a causal system in which predictions are based solely on current and previous inputs. Although causality is a natural fit for autoregressive models in sequential vision tasks~\cite{jiang2024survey}, it is at odds with the nature of spatial images, resulting in a fundamental mismatch between Mamba-based models and common visual tasks, such as AI-generated image detection.

\noindent Vision transformer-based models capture the dependencies between all visual tokens through the attention matrix, thus providing a global receptive field. In contrast, Mamba-based models use $S6$ blocks, where each prediction relies solely on the previous hidden state $h_{t-1}$ and the current input $x_t$. This makes it a causal model, well suited to sequential tasks, but unsuitable for spatial images, such as \acs{ai}-generated image detection, where it is essential to capture global relationships. To overcome these limitations, some approaches introduce multiple scanning techniques to expand the receptive field. However, these methods have two major drawbacks:

\begin{enumerate}
    \item Spatial inconsistency: Tokens are scanned in a specific order, introducing bias and failing to preserve spatial relationships correctly.
    \item Unnecessary redundancy: Adding multiple scans increases the information captured, but creates unnecessary overhead, affecting model efficiency, and can also hinder effective cue extraction for \acs{ai}-generated image detection task.
\end{enumerate}

\noindent Thus, the major challenges is to design an approach that enables Mamba to consistently capture spatial relationships in visual data, without introducing these drawbacks.\\

\noindent Despite their computational efficiency, Vision Mamba models generally perform less well than Transformers when it comes to image feature extraction. This discrepancy can be explained by several architectural factors. Firstly, SSMs rely on sequential scanning, which makes it more difficult to model rich 2D spatial dependencies compared with the fully parallel token interactions made possible by self-attention. Secondly, the effective global receptive field is more limited, as information propagates directionally through selective scanning rather than through direct long-range pairwise connections. In addition, Vision Mamba models are sensitive to the choice of patch order and scanning direction, which can lead to inconsistencies in the representation of spatial structure. Finally, SSMs have difficulty in capturing fine local variations, such as subtle textures, noise irregularities or small artifacts, which are essential for distinguishing synthetic images. Together, these factors explain the performance gap observed between Vision Mamba and Transformer-based approaches.

\subsection{Future directions for improving Vision Mamba models}

Although Vision Mamba models show promise, our experiments reveal that they currently exhibit limited robustness and poor generalization to unseen generative models. To address these shortcomings, several research directions can be explored to strengthen their capacity for modeling complex visual artifacts and improve their performance in AI-generated image detection.

\begin{enumerate}
  \item  \textbf{Hybrid SSM-Attention architectures:}
Incorporating lightweight self-attention blocks into SSM-based models could help compensate for their weaker ability to capture long-range spatial dependencies, enabling richer global interactions between features.

  \item \textbf{Improved 2D scanning strategies:}
Current SSM scanning mechanisms often rely on fixed raster or serpentine patterns. Exploring other 2D scanning methods or integrating **multiscale scanning** could improve the model's ability to represent the complex spatial structures present in synthetic manipulations.

  \item \textbf{Frequency domain components:}
Synthetic images often contain high-frequency artifacts. The addition of modules operating in the frequency domain, such as DCT- or FFT-based feature extractors, can improve sensitivity to these artifacts and enhance overall detection performance.

  \item \textbf{Lightweight pre-training on diverse generative models:}
SSM-based models could benefit from small-scale, diverse pretraining on images generated by multiple diffusion and GAN pipelines. This could improve their robustness and generalization across unseen generative sources.

\end{enumerate}

\section{Conclusion}
\label{sec:conclusion}
% In this paper, we have investigated the performance of Vision Mamba models in detecting \ac{ai}-generated images, comparing them to methods based on \acp{cnn}, attention mechanisms (Transformers), and \acp{vlm}. Our experiments revealed that Vision Mamba models exhibit difficulty generalizing across diverse data distributions, particularly when confronted with images generated by models unseen during training. The results underscore the superior generalization ability of \acp{vlm} over existing baselines and \acs{sota} methods, highlighting a potential architectural limitation of Vision Mamba models and, despite their smaller parameter count and fast inference times, making them suitable for real-time applications.
In this paper, we conducted a comprehensive evaluation of Vision Mamba models for \acs{ai}-generated image detection and compared their performance with \acp{cnn}, Transformer-based methods, and \acp{vlm}. Our experiments demonstrate that while Vision Mamba models achieve competitive accuracy when trained and tested on images generated by the same model, their performance deteriorates substantially when evaluated on images from unseen generators. This lack of generalization is consistently reflected in our quantitative results, where Vision Mamba models show large drops across nearly all cross-generator evaluation settings and adversarial robustness benchmarks.

\noindent The observed performance variation relative to prior work can be attributed to fundamental architectural limitations of Vision Mamba. Unlike Transformers, which model global relationships through full self-attention, Vision Mamba relies on sequential state-space propagation, restricting its ability to capture long-range spatial dependencies and fine-grained visual cues. As shown in our experiments, these architectural constraints make Vision Mamba particularly sensitive to distribution shifts introduced by different generative models, leading to reduced generalization and robustness. In contrast, \acp{vlm} exhibit stronger cross-distribution performance due to their large-scale pretraining and richer visual-linguistic representations.

\noindent Overall, our findings highlight both the potential and the limitations of Vision Mamba for synthetic image detection. Although its low parameter count and fast inference make it appealing for real-time applications, significant architectural enhancements are required to improve its generalization and robustness. We have outlined several promising research directions, including hybrid SSM–attention designs, improved 2D scanning mechanisms, frequency-domain feature modeling, and diverse pretraining strategies, which may help bridge the performance gap observed in this study.\\

\noindent {\bf Acknowledgments:} This work has been partially funded by the project PCI2022-134990-2 (MARTINI) of the CHISTERA IV Cofund 2021 program. Abdenour Hadid is funded by TotalEnergies collaboration agreement with Sorbonne University Abu Dhabi.

% Use the style of numbering in square brackets.
% If nothing is used, default style will be taken.
%\begin{enumerate}[a)]
%\item 
%\item 
%\item 
%\end{enumerate}  

%\begin{itemize}
%\item 
%\item 
%\item 
%\end{itemize}  

%\begin{description}
%\item[]
%\item[] 
%\item[] 
%\end{description}  

% \clearpage %%Remove this from your manuscript

\printcredits

\bibliographystyle{cas-model2-names}

\bibliography{cas-refs}



\end{document}